\documentclass[journal]{IEEEtran}
\pdfoutput=1
\usepackage{amsmath,amsfonts}
\usepackage{mathtools}
\usepackage{algorithmic}
\usepackage{algorithm}
\usepackage{array}
\usepackage[caption=false,font=normalsize,labelfont=sf,textfont=sf]{subfig}
\usepackage{textcomp}
\usepackage{stfloats}
\usepackage{url}
\usepackage{verbatim}
\usepackage{graphicx}
\usepackage{cite}
\usepackage{booktabs}
\usepackage{makecell}
\usepackage{multicol}
\usepackage{multirow}
\usepackage{longtable}
\usepackage{threeparttable}
\usepackage{graphicx}
\usepackage{subcaption}
\usepackage{booktabs}
\usepackage{amsmath}
\usepackage{amssymb}
\usepackage{cite}
\usepackage{algorithm, algorithmic}
\usepackage{color}
\usepackage{amssymb}
\usepackage{pifont}
\usepackage[intoc]{nomencl}
\makenomenclature

\usepackage[table,xcdraw]{xcolor}
\usepackage{amssymb}
\definecolor{rblue}{rgb}{0,0.5,1}
\definecolor{awesome}{rgb}{1.0, 0.13, 0.32}
\definecolor{hollywoodcerise}{rgb}{0.96, 0.0, 0.63}
\definecolor{lasallegreen}{rgb}{0.03, 0.47, 0.19}
\definecolor{hanpurple}{rgb}{0.32, 0.09, 0.98}
\definecolor{green(pigment)}{rgb}{0.0, 0.65, 0.31}

\usepackage[pagebackref=false,breaklinks=true,colorlinks,bookmarks=false]{hyperref}
\hypersetup{colorlinks=true,linkcolor={red},citecolor={hanpurple},urlcolor={magenta}}

\begin{document}

\title{CT-UIO: Continuous-Time UWB-Inertial-Odometer Localization Using Non-Uniform B-spline with Fewer Anchors}

\author{Jian Sun, Wei Sun, Genwei Zhang, Kailun Yang, Song Li, Xiangqi Meng, Na Deng, and Chongbin Tan%
\thanks{This work is supported by the National Natural Science Foundation of China (U22A2059 and No. 62473139); Project of State Key Laboratory of Advanced Design and Manufacturing Technology for Vehicle; Open Foundation of Engineering Research Center of Multi-Mode Control Technology and application for Intelligent System, Ministry of Education; the Hunan Provincial Research and Development Project (Grant No. 2025QK3019); Open Research Fund of the Hunan Province Key Laboratory of Electric Power Robotics, Changsha University of Science \& Technology, China (Grant No. 2025ZKPT061). 
(\textit{Corresponding authors: Wei Sun; Genwei Zhang; Kailun Yang.})
}
\thanks{J. Sun, K. Yang, S. Li, X. Meng, N. Deng, and C. Tan  are with the National Engineering Research Center of Robot Visual Perception and Control Technology, 
Hunan University, Changsha 410012, China (e-mail: sunjian1993@hnu.edu.cn).}
\thanks{J. Sun is also with the School of Artificial Intelligence,  Changsha University of Science and Technology, Changsha 410004.}
\thanks{W. Sun is with the School of Artificial Intelligence and Robotics, Hunan University, Changsha 410012, China (e-mail: wei\_sun@hnu.edu.cn).}
\thanks{K. Yang is also with the School of Artificial Intelligence and Robotics, Hunan University, Changsha 410012, China (e-mail: kailun.yang@hnu.edu.cn).}
\thanks{G. Zhang is with the State Key Laboratory of Chemistry for NBC Hazards Protection, Beijing 102205, China (e-mail: zhanggenwei@sklnbcpc.cn).}
}

\markboth{IEEE Transactions on Mobile Computing, December~2025}%
{Sun \MakeLowercase{\textit{et al.}}: CT-UIO}

\maketitle

\begin{abstract}
Ultra-wideband (UWB) based positioning with fewer anchors has attracted significant research interest in recent years, especially under energy-constrained conditions. However, most existing methods rely on discrete-time representations and smoothness priors to infer a robot's motion states, which often struggle with ensuring multi-sensor data synchronization. In this article, {we present a continuous-time UWB-Inertial-Odometer localization system (CT-UIO), utilizing a non-uniform B-spline framework with fewer anchors.} 
Unlike traditional uniform B-spline-based continuous-time methods, we introduce an adaptive knot-span adjustment strategy for non-uniform continuous-time trajectory representation. This is accomplished by adjusting control points dynamically based on movement speed. To enable efficient fusion of {inertial measurement unit (IMU)} and odometer data,  we propose an improved extended Kalman filter (EKF) with innovation-based adaptive estimation to provide short-term accurate motion prior. Furthermore, to address the challenge of achieving a fully observable UWB localization system under few-anchor conditions, the virtual anchor (VA) generation method based on multiple hypotheses is proposed. At the backend, we propose an adaptive sliding window strategy for global trajectory estimation. {Comprehensive experiments are conducted on three self-collected datasets with different UWB anchor numbers and motion modes.} {The result shows that the proposed CT-UIO achieves $0.403m$, $0.150m$, and $0.189m$ localization accuracy in corridor, exhibition hall, and office environments, yielding $17.2\%$, $26.1\%$, and $15.2\%$ improvements compared with competing state-of-the-art UIO systems, respectively.} The codebase and datasets of this work will be open-sourced at~\url{https://github.com/JasonSun623/CT-UIO}.
\end{abstract}


\nomenclature[A,17]{$U,I,O,W$}{UWB, IMU, odometer and world frame}
\nomenclature[A,19]{$\hat{\textbf{x}}_{t_k}$,$\hat{v}_{t_k}$}{Estimated robot's position and velocity at time $t_k$}
\nomenclature[A,12]{$\textbf{p}_n $}{Position of the  $n$-th UWB anchor}
\nomenclature[A,14]{$r_{t_{k+1}}$}{UWB ranging measurement at  time $t_{k+1}$}
\nomenclature[B,06]{$\omega_{t_k},\tilde{\omega_{t_{k}}}$}{Measured and true value of the angular velocity at time $t_k$}
\nomenclature[A,03]{$a_{t_k},\tilde{a}_{t_{k}}$}{Measured and true value of the linear acceleration at time $t_k$}
\nomenclature[A,15]{ $\prescript{I}{W}{\textbf{R}}$}{Rotation matrix from frame $W$   to frame $I$}
\nomenclature[B,02]{$\varepsilon_{g},n_{g}$,$\varepsilon_{a},n_{a}$}{Bias and zero-mean Gaussian noises of the accelerometer and gyroscope}
\nomenclature[B,07]{$^o\omega,^ov,^o\theta,^o\textbf{d}$}{Angular velocity, linear velocity, derived  yaw angle and 2D relative motion position from the odometer
measurements}
\nomenclature[A,20]{$Z(k),\tilde{Z}(k),H(k)$}{Observation measurement vector, innovation matrix and Jacobian matrix of the pose at time $k$}
\nomenclature[A,09]{$M$}{Adaptive size of the observation statistics}
\nomenclature[A,07]{$\textbf{F},\textbf{J}$}{Fisher information Matrix and Jacobian matrix of the ranging measurements}
\nomenclature[A,06]{$\text{Exp}(\cdot)$}{Function that converts quaternion to rotation matrice}
\nomenclature[A,05]{$det(\cdot)$}{Determinant of a matrix} 
\nomenclature[A,13]{$\hat{\textbf{p}}_{n}^{j}$}{The $j$-th VA’s position derived from the $n$-th UWB anchor} 
\nomenclature[A,04]{$\textbf{cp}(t),\textbf{R}(t)$}{The $i$-th  control point of the translation part and the $i$-th  control point of the rotation part}  
\nomenclature[B,03]{$\tilde{\boldsymbol{\Lambda}}^{(k+1)}(i)$}{Non-constant cumulative basis matrix which is defined by its order $k$}
\nomenclature[A,18]{$\textbf{u}(t)$}{ Normalized time vector}
\nomenclature[A,10]{$n(\space,\space)$}{The number of samples within a sliding window}
\nomenclature[A,11]{$n_{cp}$ }{The number of control points}
\nomenclature[A,16]{$\bigtriangleup t_i$}{The $i$-th knot span}
\nomenclature[A,07]{$\textbf{e}_{t}^{r}, \textbf{e}_{t}^\text{IMU}, \mathbf{e}_{t}^\text{odom}$}{Range, IMU and odometer measurement residual at time $t$}
\nomenclature[A,02]{$\boldsymbol{\textbf{a}(t),\omega}(t), \textbf{v}(t)$}{linear acceleration, Angular velocity,  and linear velocity at time $t$ obtained from the derivatives
of the continuous-time B-splines}
\nomenclature[B,04]{$\xi(\cdot), \psi(\cdot)$}{Pre-integration of IMU and odometer measurements}
\nomenclature[A,08]{$\Delta L$}{Adaptive sliding window length}
\nomenclature[B,01]{$\alpha_{cp}$}{Control point density}
\nomenclature[B05]{$\sum_{k}^r,\sum_{k}^\text{IMU},\sum_{k}^\text{odom},\sum_{k}^\text{Prior}$}{Covariance matrix associated with UWB ranging, IMU, odometer measurements, and prior factor from marginalization}

\printnomenclature

\section{Introduction}
\IEEEPARstart{L}{ocation-based} services~\cite{9184260,10538051,9705111,10679703,Diff9D} such as trajectory prediction, object tracking, and automatic picking operations, have become an integral part of robotics research. 
Global navigation satellite systems (GNSS) can achieve meter-level accuracy in outdoor clear-sky environments. However, reliable localization of GNSS is impractical indoors due to the attenuation of GNSS signals. 
To achieve robust indoor positioning services in GPS-denied environments, Ultra-wideband (UWB) technology is considered as a promising alternative solution benefiting from its stability, low cost, and scalability in large-scale deployment~\cite{10400775,10704747}. 
Based on UWB signal ranging, most of the current range-based localization schemes require at least three anchors to determine a moving target’s 2D position, which is also referred to as multilateration~\cite{2013.0815,PARK2024103234}. 
{Compared to the range-based localization schemes, the fingerprint estimation (FPE) could provide relatively high localization accuracy under multipath and non-line-of-sight (NLOS) scenarios~\cite{8806041}. 
In recent years, deep-learning-based UWB localization has emerged as a rapidly evolving field. 
Poulose and Han~\cite{app10186290} proposed a long short-term memory (LSTM) based model to mitigate positioning degradation caused by UWB ranging errors and NLOS conditions, which achieved improved performance with a mean localization error of $7cm$.}
However, in energy-constrained UWB networks or deployment scenarios with spatial limitations, such as narrow corridors or tunnels, only one or two anchors may be available. 
These limitations render traditional multilateration, fingerprinting, and even deep-learning-based methods ineffective in such environments.

In such situations, range-based localization with few anchors is challenging and has become an attractive research direction. 
Existing methods~\cite{10268655,tong2022single,9829196,9440892} typically require special anchors’ antenna arrays or integrate additional sensors, such as {inertial measurement unit} (IMU), wheel odometers, and cameras, to provide supplementary velocity or angle measurements. 
IMU and odometers are environment-independent and provide instantaneous response speeds, enabling the construction of relative position constraints between consecutive UWB ranging measurements.
Conventional sensor fusion schemes~\cite{10268655,9440987,Xiong2022} such as extended Kalman filter (EKF) and particle filter (PF), typically rely on smoothness priors to infer a robot’s motion states. 
However, these methods are based on discrete-time (DT) trajectory estimation, which may fail to adequately represent smoothness. 
Interpolation schemes used to infer robot motion between discrete states may be inaccurate. Furthermore, multi-modal sensor setups often involve asynchronous measurements from different sensors, which causes difficulty in fusing data at the same time instants during the estimation process. 

Within recent years, continuous-time (CT) representation has been increasingly applied to multi-sensor calibration, motion planning, and target tracking~\cite{10311080,lang2023coco,cioffi2022continuous,nguyen2024eigen}. 
The key advantage of the continuous-time formulation lies in its ability to generate smooth trajectories with continuous data streams. 
In particular, B-spline defines temporal polynomials to represent trajectories by a set of control points, which enables pose querying at any time instant with locality. 
Most existing methods~\cite{li2023continuous,10045587,9636676} adopt the uniform B-spline-based strategy for continuous-time trajectory smooth updates, where control points are spaced uniformly and typically predetermined. 
However, these methods often rely on the assumption of a constant velocity model, which is unsuitable for many real-world applications. 
Since the robot's velocity is constantly changing, smaller control point spacing may result in overfitting and increased computational complexity, while larger spacing reduces overall accuracy. Benefiting from dynamic control point distribution, non-uniform B-splines are becoming an attractive alternative. 
Related research works by Lang~\textit{et al.}~\cite{lang2023coco} and Dubé~\textit{et al.}~\cite{dube2016non} illustrate that non-uniform knot sampling strategies can accommodate trajectory complexity more effectively. 
These methods have also proved that the non-uniform sampling strategy can be superior to the uniform strategy in terms of trajectory estimation accuracy and time performance.

However, research on non-uniform B-spline methods has primarily focused on {light detection and ranging (LiDAR)}~\cite{lang2023coco} and cameras~\cite{9320417} applications for {simultaneous localization and mapping (SLAM)}, it has not been applied to few-anchor UWB-based positioning. Moreover, existing non-uniform strategies for determining appropriate knot spans often rely solely on velocity estimates from IMU measurements, which are prone to noise, resulting in inaccurate spacing interval divisions.

In this work, we propose {a continuous-time UWB-Inertial-Odometer system (CT-UIO)} using non-uniform B-splines with few anchors.
CT-UIO dynamically adjusts control point spacing based on changes in motion trajectories. In general, although continuous-time methods have been extensively studied, most existing works rely on uniform B-splines to achieve continuous-time representations and estimation. 
{To the best of the authors’ knowledge, this article is the first to utilize a non-uniform continuous-time representation for UWB/IMU/odometer fusion localization, which can achieve efficient pose estimation in few-anchor scenes or with time-varying motion states. }

The contributions of this article are summarized as follows.
\begin{itemize}
    \item{{We propose the first continuous-time UWB-Inertial-Odometer localization system (CT-UIO) that employs a non-uniform B-spline trajectory representation. This enables seamless fusion of time-unsynchronized UWB, IMU, and odometer measurements, achieving  both high positioning accuracy and real-time performance  in few-anchor, time-varying motion conditions.}}
 
    \item{{We develop an improved EKF-based IMU/odometer fusion model with innovation-based adaptive estimation, providing accurate short-term motion priors.  These priors are then used to dynamically adjust knot spans in the trajectory optimization, increasing control points only when motion speed changes significantly. This adaptive knot span adjustment strategy allows accurate modeling of rapid motion changes and improves overall pose estimation accuracy.}}

    \item{{We propose a virtual anchor (VA) generation method that combines UWB ranging with motion priors from the improved EKF-based IMU/odometer fusion model, ensuring a fully observable UWB localization system even when fewer physical UWB anchors are available. Additionally, we design a multiple-hypothesis-based VA deployment scheme to avoid collinearity issues and enhance positioning robustness.}}

    \item{The CT-UIO system is evaluated on several real-world datasets, including comparison with state-of-the-art methods. Extensive experimental results demonstrate the system's superiority, particularly in handling fast motion. To benefit the community, the relevant datasets and source codes are made publicly available at \url{https://github.com/JasonSun623/CT-UIO}.}
\end{itemize}

{The remainder of this article is organized as follows. 
In Sec.~\ref{related_works}, the related works for few-anchor UWB localization and continuous-time representation are discussed. {In Sec.~\ref{Motivation}, the motivation is presented.}
In Sec.~\ref{overview}, the proposed CT-UIO is introduced in detail. 
To verify the performance of the proposed methods, extensive experiments are conducted in Sec.~\ref{experiments}. 
Finally, in Sec.~\ref{conclusion}, we draw the conclusion and discuss future research directions. 

\section{Related Works}
\label{related_works}
\subsection{Few-Anchor UWB Localization}
Few-anchor UWB localization has recently attracted increasing attention and has been widely studied due to its low complexity and cost-effective deployment~\cite{tong2022single,cao2020accurate,qin2024single}. 
These research works focus exclusively
on fusion schemes that integrate UWB measurements with other motion sensors. Mainstream works primarily employ filter-based, optimization-based, and two-stage methods to incorporate additional velocity observations. 
In filter-based solutions, Cao~\textit{et al.}~\cite{cao2020accurate} employed  EKF to combine a single UWB anchor with a nine-DOF IMU for velocity and orientation estimation. Qin~\textit{et al.}~\cite{qin2024single} designed a set membership filter (SMF) method based on constrained zonotopes for single-beacon localization problems. 
Gao~\textit{et al.}~\cite{penggang2022novel} introduced UWB ranging, non-holonomic constraints from the carrier, and trajectory constraints as observations of the EKF system. The proposed method can realize positioning in corridor-like areas. However, it relies on non-holonomic and corridor-specific trajectory constraints, which may not be applicable in unknown environmental structures. 
In optimization-based solutions, {Dong~\textit{et al.}~\cite{9829196} proposed a velocity-constrained Moving Horizon Estimation (VMHE). 
By formulating the trajectory optimization problem through incorporating indirectly derived velocity constraints and the geometric evolution equation. Experimental results demonstrated that the proposed method achieves robust trajectory estimation in long-duration flights using only a single UWB beacon.}  
Li~\textit{et al.}~\cite{li2021computationally} proposed a Moving Horizon Estimation (MHE) algorithm, which utilizes historical measurement data from a single UWB anchor and IMU during a time horizon. Additionally, a Gradient Aware Levenberg-Marquardt (GALM) algorithm was further proposed to solve the optimization problem with a calculation cost. 
For two-stage methods, Yang~\textit{et al.}~\cite{yang2023novel} proposed an improved positioning method that leverages Bayesian optimization to provide the estimated IMU data based on an error model. 
A filter then updates the position at the ranging time using the prior position error. 
Zhou~\textit{et al.}~\cite{zhou2024optimization} proposed a two-stage optimization method that could yield accurate solutions with the single-anchor and odometer.
In the first stage, they construct a factor graph that incorporates odometry positions and UWB measurements to optimize state estimation. In the second stage, an adaptive trust region algorithm is performed to refine the location estimate and maintain robustness with inequality constraints. 
{Cheng~\textit{et al.}~\cite{WOS:001173317800046} proposed the loosely coupled fusion framework utilizing generalized probability data association (GPDA). This method effectively associates UWB measurements with INS-derived state estimates, reducing the impact of NLOS error.}
While these methods approximate the time offsets between state estimation and sensor measurements, the hypothesis of time synchronization may not always hold in practical scenarios. 
Unlike the existing few-anchor UWB localization methods, our method
formulates the trajectory in continuous-time with B-spline, which allows asynchronous, high-frequency sensor measurements to be aligned at any time instants along the trajectory, rather than relying on linear interpolation of discrete-time poses only at measurement times.

\subsection{Continuous-Time Representation Using B-spline}
Continuous-time (CT) representation is a popular and natural choice for formulating trajectories and ensuring smoothness. 
It can estimate the robot's pose as a continuous function of time without the need to introduce additional states at every measurement time. The most common continuous-time-based model is based on splines and Gaussian processes~\cite{cioffi2022continuous}. 
In this article, we focus on B-spline-based trajectory representation. Current literature has paid increasing attention to applying B-splines for continuous-time state estimation in multi-sensor fusion systems, including the LiDAR-Inertial system, visual-inertial system, ultra-wideband-inertial, and so on.
Nguyen~\textit{et al.}~\cite{nguyen2024eigen} developed a real-time continuous-time LiDAR-inertial odometry (SLICT2), which achieved efficient optimization with few iterations using a simple solver.
Lu~\textit{et al.}~\cite{lu2023event} proposed an event-based visual-inertial velometer that incrementally incorporates measurements from a stereo event camera and IMU. 
Li~\textit{et al.}~\cite{li2023continuous} proposed a spline-based approach (SFUISE) for continuous-time Ultra-wideband-Inertial sensor fusion, which addressed the limitations of discrete-time sensor fusion schemes in asynchronous multi-sensor fusion and online calibration. 
These methods generally employ uniform knot B-splines for trajectory modeling, relying on assumptions of zero velocity or constant speed. However, these assumptions often fail to capture the dynamic nature of real-world motion. Non-uniform B-splines offer a more flexible distribution of control points, enabling a different density of control points based on the smoothness of the trajectory segment. 
Ovrén~\textit{et al.}~\cite{ovren2018spline} introduced an energy proportionality index in spline fitting to optimize the selection of knot spacing. 
By leveraging the different frequency response characteristics of spline basis functions and specific energy values, appropriate knot spacing can be automatically selected. 
Lang~\textit{et al.}~\cite{lang2023coco} tightly coupled the measurements from LiDAR, IMU, and cameras using non-uniform B-spline curves. 
They adaptively adjusted the number of control points based on IMU observations to detect different motion patterns, thereby improving adaptability to complex environments and motion patterns. However, relying solely on IMU data leads to velocity estimate divergence, which impacts the distribution of control points and ultimately reduces positioning accuracy. 
In contrast, the proposed method integrates an IMU/Odometer fusion model to provide short-term accurate velocity estimates, enabling the dynamic adjustment of control point density for more precise trajectory modeling. 

\begin{figure*}[!t]
\centering
\subfloat[\label{1A}]{
\includegraphics[width=\linewidth,trim=0.cm 6.9cm 0cm 2.8cm,clip]{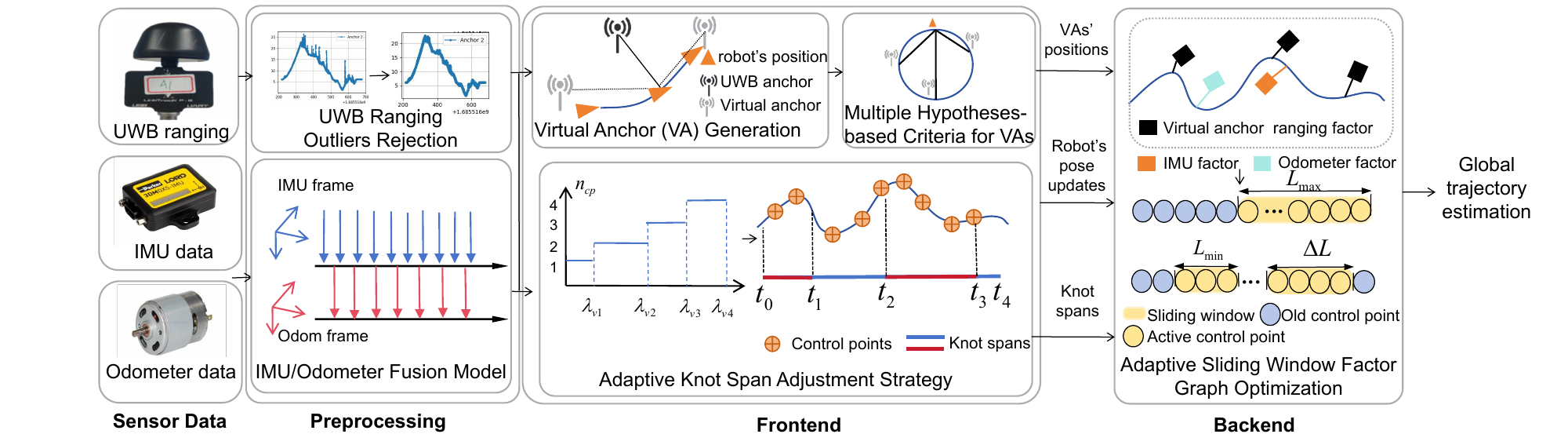}}
\\
\subfloat[\label{1B}]{
		\includegraphics[width=0.9\linewidth,trim=0.cm 5cm 0cm 2.8cm,clip]{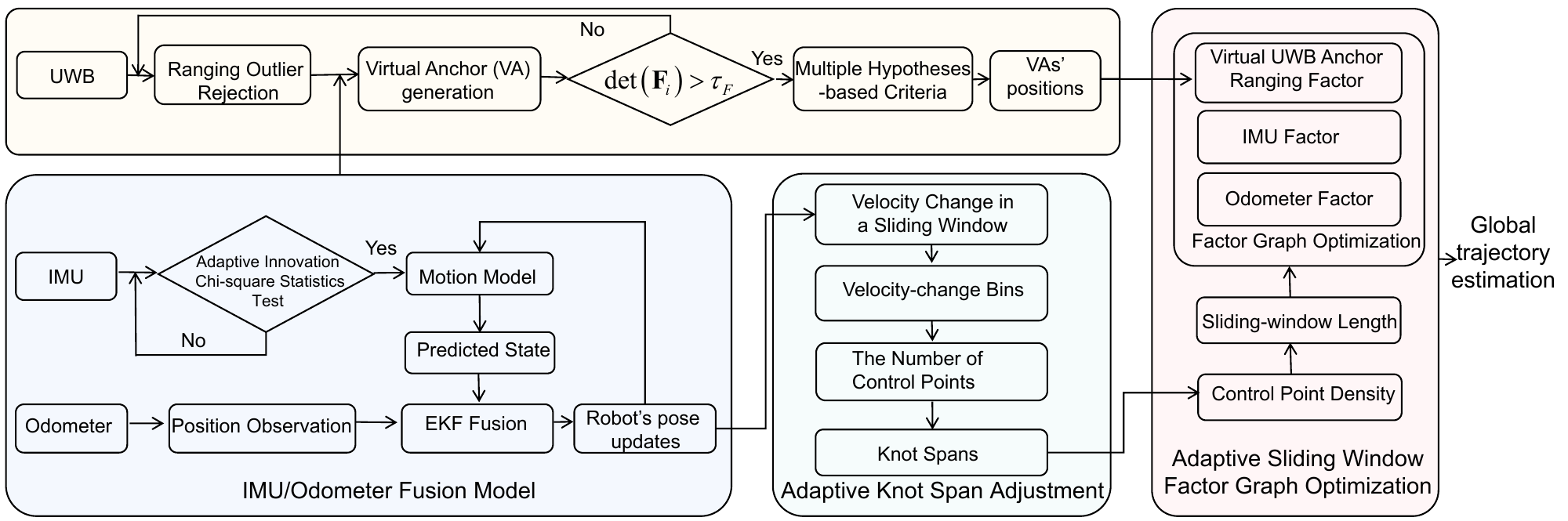} }
\caption{{The framework of CT-UIO: In the preprocessing stage, UWB ranging data is detected, and outliers are removed. Simultaneously,  incoming IMU and odometer data are fused using an improved EKF to provide short-term accurate motion priors. The frontend performs incremental pose updates of the robot using the IMU/odometer fusion model, and the resulting poses are combined with UWB ranging data to determine VAs' positions. Furthermore, leveraging the motion estimates from the IMU/odometer fusion model, the adaptive knot span adjustment strategy non-uniformly places the control points. The backend conducts factor graph optimization
 with an adaptive sliding window to achieve global trajectory estimation.}}
\label{fig_0}
\end{figure*}

\section{{Motivation}}
\label{Motivation}
{
Despite the progress in few-anchor UWB localization and continuous-time trajectory estimation,  these methods still suffer from several challenges related to sensor synchronization, adaptability, and observability under few-anchor conditions. The primary challenges can be summarized as follows.}

{\textbf{Observability degradation and asynchronous fusion:}  When only one or two UWB anchors are available, as commonly encountered in corridors or energy-constrained deployments. Traditional multilateration or range-based optimization methods become under-constrained due to the lack of sufficient observations. Besides, discrete-time fusion frameworks estimate motion states only at discrete sampling instants. This formulation depends on interpolation and smoothness priors to approximate trajectories, leading to motion distortion and drift when fusing asynchronous UWB, IMU, and odometer data.}

{\textbf{Limitations of uniform continuous-time representations:}  Most existing CT methods employ uniform B-spline representations with fixed knot intervals, assuming approximately constant motion. However, such uniform spacing cannot adapt to time-varying motion, causing overfitting and increased computational cost during slow motion, and loss of accuracy during rapid motion.}

{Motivated by the above discussion, we propose the Continuous-Time UWB-Inertial-Odometer (CT-UIO) framework that introduces an adaptive knot-span adjustment strategy for non-uniform B-spline trajectory modeling and employs an improved EKF-based IMU/odometer fusion model to generate virtual anchors (VAs), thereby ensuring full observability under few-anchor conditions.}

\section{System Overview}
\label{overview}
{Fig.~\ref{fig_0} shows the overview of our CT-UIO. 
In the beginning, we assume that the UWB-Inertial-Odometer system is calibrated, and the IMU bias and odometer scale are initialized. 
In the preprocessing, we apply inequality constraints to remove UWB-ranging outliers and build the IMU/odometer fusion model (Sec.~\ref{sec:preprocessing}). 
To make the few-anchor UWB localization system fully observable, we combine the UWB ranging measurements with short-term position estimates from the IMU/odometer fusion model to generate VAs (Sec.~\ref{sec:VA}). 
Furthermore, we present multiple hypotheses-based criteria to {avoid collinearity} of VAs (Sec ~\ref{sec:Criteria}).
Next, we propose a non-uniform continuous-time state estimation method, using the adaptive knot span adjustment strategy to tightly fuse the high-frequency measurements from UWB ranging, IMU, and odometer (Sec.~\ref{sec:Adaptive_Knot}).  
The backend conducts a UWB-Inertial-Odometer factor graph with an adaptive sliding window to perform joint optimization to obtain optimal estimations (Sec.~\ref{sec:optimization}). }

For the UWB/IMU/Odometer (UIO) integrated system, we use ${U}$ to denote the UWB frame, {$I$} for the IMU frame, {$O$} for the wheel odometer frame, {$W$} for the world coordinate frame.
\subsection{Preprocessing}
\label{sec:preprocessing}
\subsubsection{UWB Ranging Outlier Rejection}
\label{sec:outlier_rejection}
UWB ranging outliers occur regularly under Non-line-of-sight (NLOS) and multipath conditions~\cite{10458971}. 
If these outliers are not treated properly during data processing, the performance of UWB-based localization can be degraded due to the fusion of inaccurate ranging measurements. 
Therefore, we filter the UWB-ranging outliers using an inequality constraint. The ranging measurement $r_{t_{k+1}}$ at the $t_{k+1}$ time should satisfy:
\begin{equation}
  \left |{\left \| \hat{\textbf{x}}_{t_k}-\textbf{p}_n \right \| }_2-r_{t_{k+1}}\right | <\gamma \hat{v}_{t_k}\left ( t_{k+1}- t_k\right ),   
\end{equation}
 where $\hat{\textbf{x}}_{t_k}$ and $\hat{v}_{t_k}$ are the estimated robot's position and velocity derived from the IMU/odometer fusion model, respectively. $\textbf{p}_n $ is the position of the UWB anchor $n$.  $\gamma$ is the outlier rejection gating threshold for the individual ranging measurement. 
\subsubsection{IMU/Odometer Fusion Model}
IMU and odometer offer fast response speeds and are environment-independent. For ground robot navigation, it appears to be straightforward to fuse IMU data with wheel odometer measurements to provide motion priors over short time periods~\cite{Bai_Lai_Lyu_Cen_Wang_Sun_2022}. 

The raw IMU measurements include the angular velocity \begin{math}\omega_{t_k}\end{math} and local linear acceleration \begin{math}a_{t_k}\end{math}:
\begin{equation}
^I\omega_{t_{k}}=\tilde{^I\omega_{t_{k}}}+\varepsilon_{g}+n_{g}
\end{equation}
\begin{equation}
    ^Ia_{t_{k}} = \tilde{^Ia_{t_{k}}} + \varepsilon_{a} + 
\prescript{I}{W}{\textbf{R}} g^W + n_{a},
\end{equation}
where \begin{math}\tilde{\omega}_{t_{k}}\end{math} and \begin{math} \tilde{a}_{t_{k}} \end{math} are true values of angular velocity and linear acceleration at time $t_k$, respectively.
\begin{math} \varepsilon_{g}\end{math} and \begin{math} \varepsilon_{a}\end{math} denote the accelerometer’s bias and the gyroscope’s bias in the IMU body frame. 
$\prescript{I}{W}{\textbf{R}}$ denotes the rotation matrix from frame $\left \{ W \right \} $ to $\left \{ I \right \} $, \begin{math}g^W\end{math} is the gravity vector in the world frame, \begin{math}n_g\end{math} and \begin{math}n_a\end{math} are zero-mean Gaussian noises.

We then integrate the wheel odometer measurements between IMU times [\begin{math}t_{k-1}\end{math}, \begin{math}t_k\end{math}] and 2D relative motion as follows:
\begin{equation}
z_o=:\begin{bmatrix}
 ^o\theta\\^o\textbf{d} 
\end{bmatrix}=\begin{bmatrix}
-\int\limits_{t\in [t_{k-1},t_k] }{^o\omega_{t_k}dt} 
 \\
\int\limits_{t\in [t_{k-1},t_k] }{^ov_{t_k}\text{cos}^o\theta dt} 
 \\
-\int\limits_{t\in [t_{k-1},t_k] }{^ov_{t_k}\text{sin}^o\theta dt} 
\end{bmatrix},
\label{2dIMU}
\end{equation}
where \begin{math} ^o\theta \end{math} and \begin{math}^o\textbf{d}\end{math} are the yaw angle and 2D relative motion position in odometer frame from \begin{math} t_{k-1} \end{math} to \begin{math} t_k\end{math}. \begin{math} ^o\omega \end{math} and \begin{math} ^ov \end{math} are 2D angular and linear velocities with zero-mean white Gaussian noises in the odometer frame. 
The relative motion position can also be derived from the IMU pose:
\begin{equation}
\tilde{z}_o=:\begin{bmatrix}
 _{I}^{o}\theta\\_{I}^{o}\textbf{d}

\end{bmatrix}=\begin{bmatrix}
\textit{Log}\left( \textbf{e}_{3}^\text{T}{^oR_{t_k}}\right)
 \\
{^oR_{t_k}}(_{I}^{o}R{^I\textbf{d}})
\end{bmatrix}.
\end{equation}
{$\textit{Log}(\cdot)$ is the matrix logarithm function, \begin{math}\textbf{e}_3\end{math} is the unit bias vector.}
The attitude update is formulated as
\begin{equation}
{^oR_{t_k}}={^oR_{t_{k-1}}}\otimes R_{\bigtriangleup t_k }^{b}.
\end{equation}
\begin{math}{^oR_{t_k}}\end{math} represents the rotation matrices transformed from \begin{math} ^o\theta_{t_k}\end{math}.
{We construct the observation measurement vector composed of the poses of the IMU and the odometer as}
\begin{equation}
Z(k)=z_o-\tilde{z}_o=H(k)X(k)+\eta(k), 
\end{equation}
where \begin{math} \eta(k)  \sim (0,\sigma ^2) \end{math} means the measurement noise with zero mean and a standard Gaussian distribution.
The Jacobian matrix \begin{math}H(k)\end{math} can be indicated by 
\begin{equation}  
H(k)=\frac{Z(k)}{\partial X(k)}. 
\end{equation}
The innovation \begin{math}\tilde{Z}(k)\end{math}  can be calculated as 
\begin{equation}
\tilde{Z}(k)=Z(k)-\bar{Z}_k=H(k)X(k)+\eta(k)-H(k)\bar{X}(k),
\end{equation}
where \begin{math} \tilde{Z}(k)\sim N(0,\textbf{e}_k) \end{math}, the correlation covariance matrix of the innovation is defined as
\begin{equation}
\textbf{e}_k=E(\tilde{Z}(k)(\tilde{Z}(k))^\text{T}),
\end{equation} 
\begin{equation}
e(k)=(\tilde{Z}(k))^\text{T}(E(k))^{-1}\tilde{Z}(k).
\end{equation}
During the above EKF fusion process, abnormal IMU measurement noise can directly impact the filtering robustness and observation accuracy. 
To improve the accuracy of model parameters and observation matrix, an adaptive innovation chi-square statistics test is designed to assess the reliability of IMU measurements.
\begin{equation}
E(k)=\frac{1}{M}\sum_{i=k-M+1}^{k}\tilde{Z}(i)^\text{T}e(i)^{-1}\tilde{Z}(i),  
\end{equation}
\begin{equation}
E(k)\sim{\chi^2},
\end{equation}
\begin{equation}
\begin{aligned}
    M=\left\{\begin{matrix}
1,e(k)\ge \lambda_\text{max}, 
 \\
\xi,e(k)\le \lambda_\text{min},
 \\
\xi \times \mu^{\frac{e(k)-\lambda_\text{min}}{\alpha } },\lambda_\text{min}< e(k)<\lambda_\text{max}. 
\end{matrix}\right.
\end{aligned}
\end{equation}
Here, $M$ is the adaptive size of the observation statistics, which controls the trade-off between estimation accuracy and computational complexity. \begin{math}\mu\end{math} is the convergence rate. {$\alpha$ is a constant associated with the frequency characteristics of the IMU measurement noise. 
$\xi$ is the maximum allowable size of $M$. $\lambda_\text{min}$ and $\lambda_\text{max}$ are lower and upper thresholds for the acceptable range of  $e(k)$.}
There is the rejection domain of  \begin{math}E(k)\end{math}:
\begin{equation}
\left\{\begin{matrix}
E(k)>thr& \text{Reject}
 \\
E(k)\le thr& \text{Accept}
\end{matrix}\right..
\end{equation}
When \begin{math}E(k)\end{math} exceeds the threshold \begin{math}thr\end{math}, the abnormal fault in  IMU measurements is detected by the chi-square test statistic, and we reject this IMU measurement and do not perform the EKF update.

\subsection{{Frontend}}

\subsubsection{{Virtual  Anchor (VA) Generation}}
\label{sec:VA}
Since directly fusing range measurements can lead to unobservability when only one or two anchors are available~\cite{10268655}, we utilize virtual anchors (VAs) to construct a fully observable UWB-based localization system.
Inspired by~\cite{9119679,9520065}, we propose a VA generation method based on multiple hypotheses. 
This method introduces location reference points that are virtually placed in the UWB network,  providing additional constraints for the robot's global localization, as shown in Fig.~\ref{virtual}. 
The VA is established by a location estimator from the IMU/odometer fusion model. In this article, considering a mobile robot equipped with a UWB tag that receives a signal from a UWB anchor, the corresponding VAs can be generated by range measurements and the estimated position using the IMU/odometer fusion model. The range residual is established as
\begin{equation}
\textbf{e}_{j}^{r}=r_j-\left \| \textbf{x}_j-\textbf{p}_n \right \|_2, \label{range res}
\end{equation}
where \begin{math}r_j\end{math} is range measurement received at time  \begin{math}j\end{math}, $\textbf{x}_j$ denotes the location of the UWB tag in the world frame at time \begin{math}j\end{math}, \begin{math}\textbf{p}_n\end{math} is the position of UWB anchor \begin{math}n\end{math}.

\begin{figure}[!htbp]
\centering
\includegraphics[width=0.8\linewidth,trim=2.5cm 5cm 13cm 2cm,clip]{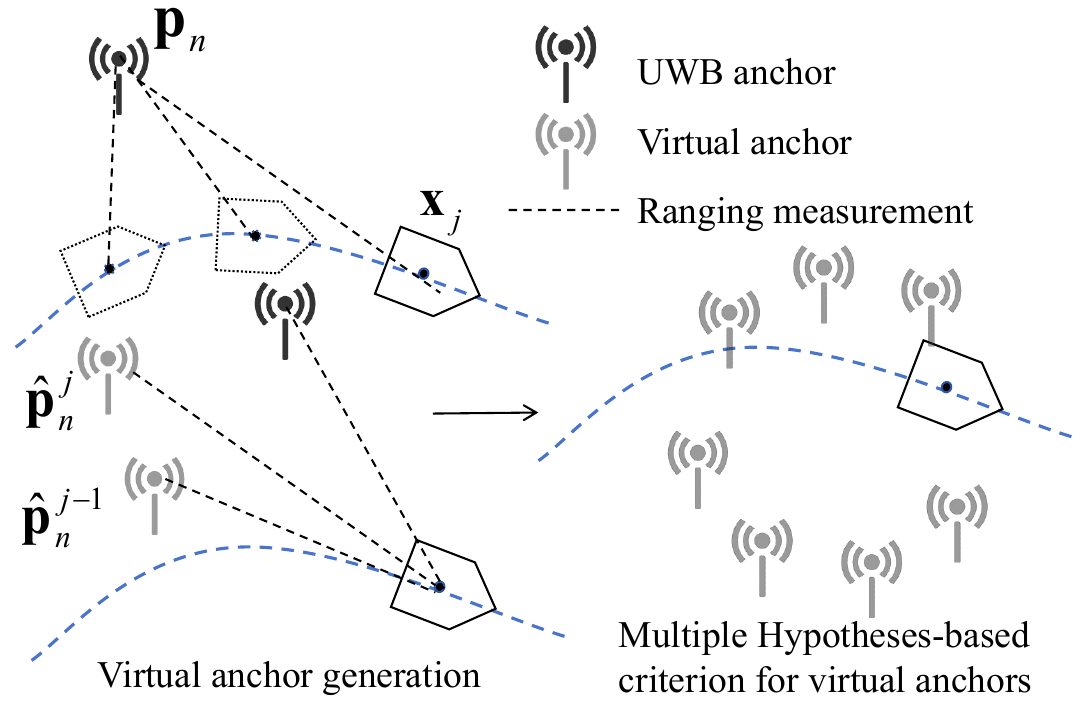}
\caption{Illustrations of the VA generation process.
}
\label{virtual}
\end{figure}

We can obtain a model of the IMU/odometer system and derive the relative position transform between \begin{math}\textbf{x}_i\end{math} and \begin{math}\textbf{x}_j\end{math} over a short time \begin{math}i\end{math} to \begin{math}j\end{math}
\begin{equation}
\textbf{x}_j=\textbf{x}_i+\text{Exp}(\textbf{q}^\omega)\triangle\textbf{d}_{ij}.
\end{equation}
Since \begin{math}\textbf{q}^\omega\end{math} and \begin{math}\triangle\textbf{d}_{ij}\end{math} can be derived from Eq.~\eqref{2dIMU}, the robot’s pose and {the derived relative} position can be expressed as the VA’s position.
Eq.~\eqref{range res} can be further reformulated as follows: 
\begin{equation}
\begin{split}
\textbf{e}_{j}^{r} &= r_j - \left \| \textbf{x}_i - (\textbf{p}_n - \text{Exp}(\textbf{q}^\omega) \triangle \textbf{d}_{ij}) \right \|_2 \\
&= r_j - \left \| \textbf{x}_i - \hat{\textbf{p}}_{n}^{j} \right \|_2,
\end{split}
\end{equation}
where \begin{math}\hat{\textbf{p}}_{n}^{j}\end{math} is regarded as the position of the VA. 
Based on this, the VA is generated through the ranging-gathering process that involves a set of robot waypoints. 
However, the random waypoints are often ill-posed in practice, resulting in suboptimal solutions. 
{Selecting an appropriate information metric to detect potential sensor measurement degeneration and determining the optimal waypoint placement are critical for enhancing localization accuracy.}
The Fisher information Matrix (FIM)~\cite{6882246} is considered to evaluate the performance of VA's position estimation, and it is defined as:
\begin{equation}\textbf{F}=\frac{1}{\sigma_{r}^{2}}\textbf{J}^\text{T}\textbf{J},\end{equation}
\begin{equation}\textbf{J}=\frac{\partial\textbf{e}_{j}^{r}}{\partial \hat{\textbf{p}}_{n}^{j} },\end{equation}
 where \begin{math}\sigma_{r}\end{math} is the standard deviation of the ranging measurements. \begin{math}
     \textbf{J}
 \end{math} represents the Jacobian matrix of the ranging measurement model with respect to the anchor’s position \begin{math}\hat{\textbf{p}}_{n}^{j}\end{math}. The determinant of the FIM \begin{math}det(\textbf{F})\end{math} contains all the ranging measurements received during time \begin{math}i\end{math} to \begin{math}j\end{math}. 
 We can set a threshold \begin{math}\tau_F\end{math} and generate a VA when \begin{equation}
    det(\textbf{F})>\tau_F.
\end{equation}  
By maximizing \begin{math}det(\textbf{F})\end{math}, the lower bound of the VA’s position estimation error can be directly determined, and the optimal VA placement can be decided. {The VA generation process continues until three non-collinear VAs are obtained.}

\subsubsection{Multiple Hypotheses-based Criteria for VAs} 
\label{sec:Criteria}
We present multiple hypotheses-based criteria to {avoid collinearity} of generated VAs.
A set of VAs is derived from the IMU/odometer fusion model. 
If all the VAs' 2D positions constitute a straight line, it becomes difficult to obtain a unique solution for the target’s position without additional constraints. To
mitigate this problem, we use the determinant to ensure that the generated VAs are non-collinear. 
Assuming the locations of three VAs are
\begin{math}
\hat{\textbf{p}}_{n}^{1}=(\hat{x}_{n}^{1},\hat{y}_{n}^{1}),
\hat{\textbf{p}}_{n}^{2}=(\hat{x}_{n}^{2},\hat{y}_{n}^{2}),
\hat{\textbf{p}}_{n}^{3}=(\hat{x}_{n}^{3},\hat{y}_{n}^{3})
\end{math}.
The determinant $D$ is defined as:
\begin{equation}
D=\begin{vmatrix}
 \hat{x}_{n}^{1} & \hat{y}_{n}^{1} & 1\\
  \hat{x}_{n}^{2} & \hat{y}_{n}^{2} & 1\\
  \hat{x}_{n}^{3} & \hat{y}_{n}^{3} & 1
\end{vmatrix}.
\end{equation}
If \begin{math}
D=0\end{math}, the three VAs are collinear. We use a sample circle with multiple VAs' position hypotheses to avoid collinearity.
Thus, assuming total \begin{math}h\end{math} different hypotheses  for position \begin{math}\hat{\textbf{p}}_{n}^{j}\end{math}, the VA's position will be written in the following form: \begin{equation}\textbf{b}_{n}^{j}=\textbf{H}_{j}^{k}\hat{\textbf{p}}_{n}^{j},\end{equation}
 \begin{equation}
 \textbf{H}_{j}^{k}=\begin{bmatrix}
 r_j & cos(2\pi(k-1)/h) & 0\\
  0&r_jsin(2\pi(k-1))/h)  & 0\\
  0& 0 &1
\end{bmatrix},k=1,...,h.
\end{equation}
The resultant VAs' positions are obtained by minimizing the following expression: 
\begin{equation}\text{arg min} \sum_{j=1}^{h}(r_j-\left \| \textbf{x}_i-\textbf{b}_{n}^{j} \right \| ).
\end{equation}

\subsubsection{{Adaptive Knot Span Adjustment Strategy}}

\label{sec:Adaptive_Knot}

In this article, we use the split non-uniform B-spline representation to
separately represent the translation part $\textbf{cp}(t)$ in \begin{math}\mathbb{R}^3\end{math} and the rotation part $\textbf{R}(t)$ in the Lie group \begin{math}SO(3)\end{math}   with two splines. 
The sampled translation vector in trajectory of the order $k$ B-spline at time \begin{math}t\in[t_k,t_{k+1}]\end{math}, can be denoted as:
\begin{equation}
\textbf{cp}(t)=\textbf{cp}_{i}+\sum_{j=1}^{k}\lambda_j(t)\cdot \textbf{d}_{j}^{i},
\end{equation}
with
\begin{equation}
\boldsymbol{\lambda}(t) =\tilde{\boldsymbol{\Lambda}}^{(k+1)}(i)\textbf{u}(t),  
\end{equation}
where \begin{math}\textbf{cp}_{i}\end{math} denotes the $i$-th  control point, \begin{math}\textbf{cp}_{i}\in \mathbb{R}^3 \end{math}, \begin{math}t\in T =\left \{ t_0,t_1,...,t_n \right \} \end{math} is the arbitrarily independent knot time of B-splines. The coefficients \begin{math}{\lambda}_j(t)\end{math} represents the $j$th column of \begin{math}\boldsymbol{\lambda}(t)\end{math}. \begin{math}\tilde{\boldsymbol{\Lambda}}^{(k+1)}(i)\end{math} is a non-constant cumulative basis matrix which is defined by its order $k$. 
\begin{math} \textbf{u}(t)=\begin{bmatrix}
1& u(t) & ... & u^k(t)
\end{bmatrix}^\text{T} \end{math} is the normalized time vector:
\begin{equation}
u(t)=\frac{t-t_i}{t_{i+1}-t_i}. 
\end{equation}
\begin{math}
    \textbf{d}_{j}^{i}
\end{math} is the difference distance vector,
\begin{equation}
\textbf{d}_{j}^{i}=\textbf{cp}_{i+j}-\textbf{cp}_{i+j-1}. 
\end{equation}
Analogically, the orientation of the trajectory in SO(3) in Lie groups can be expressed as follows:
\begin{equation}
\begin{aligned}
\textbf{R}(t)=\textbf{R}_{i}\cdot \prod_{j=1}^{k}\text{Exp}(\lambda_j(t)\cdot \boldsymbol{\Psi }_{j}^{i}), \\
\boldsymbol{\Psi }_{j}^{i}=\text{Log}(\textbf{R}_{i+j-1}^{-1}\textbf{R}_{i+j}),
\end{aligned}
\label{liegroup}
\end{equation}
where \begin{math}\textbf{R}_{i}\end{math} denotes the $i$-th rotation control points, \begin{math}\textbf{R}_{i}\in SO(3) \end{math}. \begin{math}\text{Exp}(\cdot)\end{math} is the exponential mapping that maps the Lie algebra to the Lie group, and \begin{math}\text{Log}(\cdot)\end{math} is the inverse operation, $\boldsymbol{\Psi }_{j}^{i}\in \mathbb{R}^3$.

To balance the overall accuracy and convergence speed, we exploit a cumulative cubic B-spline function ($k{=}3$) in this work, and the order of the spline is set to $4$.

For uniform-spline-based methods, control points are temporally uniformly distributed, and knot spans between them are equal to construct the B-spline.
However, this uniform placement does not guarantee optimal trajectory interpolation performance, especially when encoding time-varying states. 
We argue that it is reasonable to reduce redundant control points for steady movement. 
In addition, when the robot's motion has significant velocity fluctuations over short time intervals, a finer granularity of knots is required to capture rapid changes, which requires inserting additional control points. To solve this problem, we propose an adaptive knot span adjustment strategy that dynamically adjusts the distribution of control points to flexibly accommodate varying motion speeds, as shown in Fig.~\ref{cp}. 

\begin{figure} [!htb]
	\centering
	\subfloat[\label{1a}]{
		\includegraphics[width=0.8\linewidth,trim=1cm 0.3cm 2.0cm 1.5cm,clip]{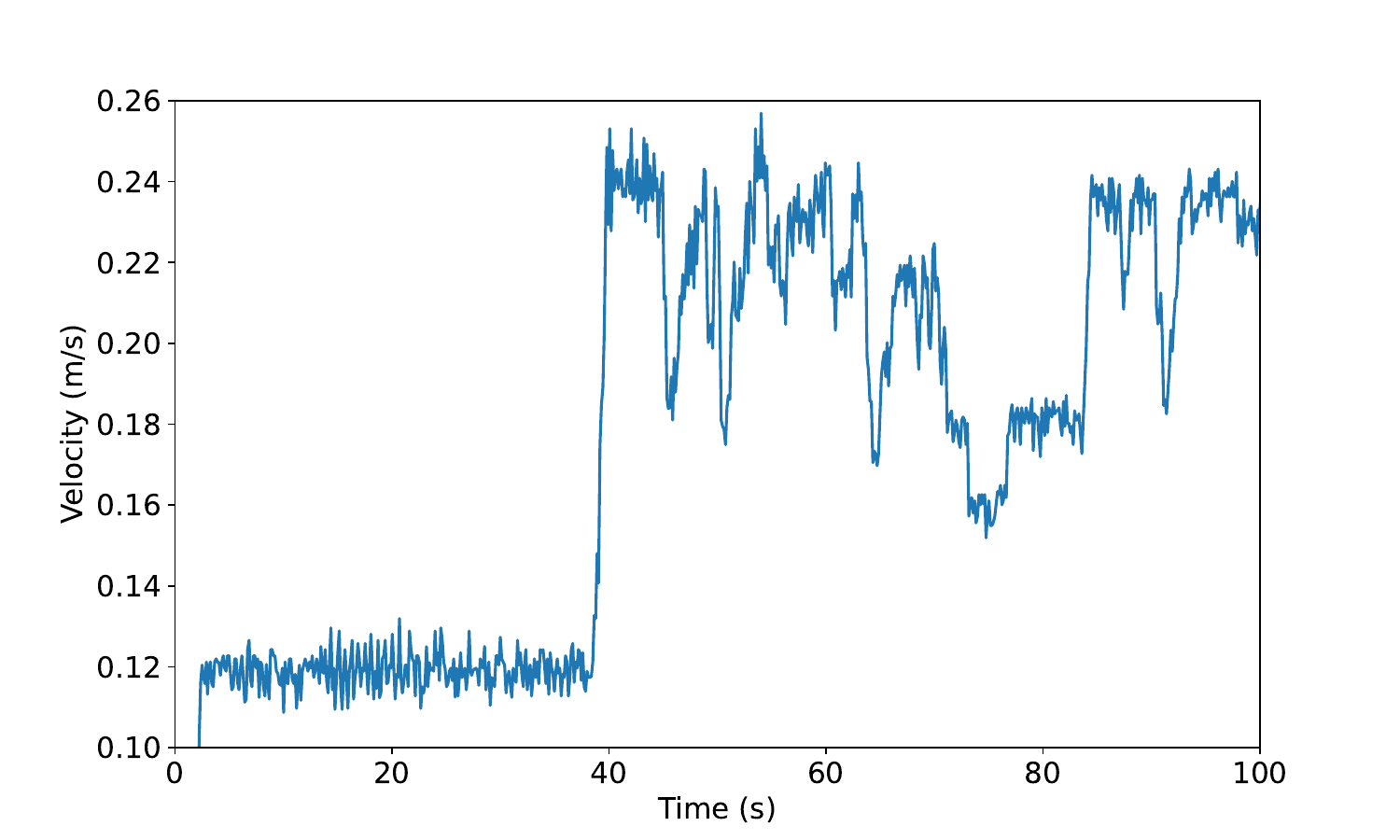}}
	\\
	\subfloat[\label{1b}]{
		\includegraphics[width=0.8\linewidth,trim=1cm 0.3cm 2.0cm 1.5cm,clip]{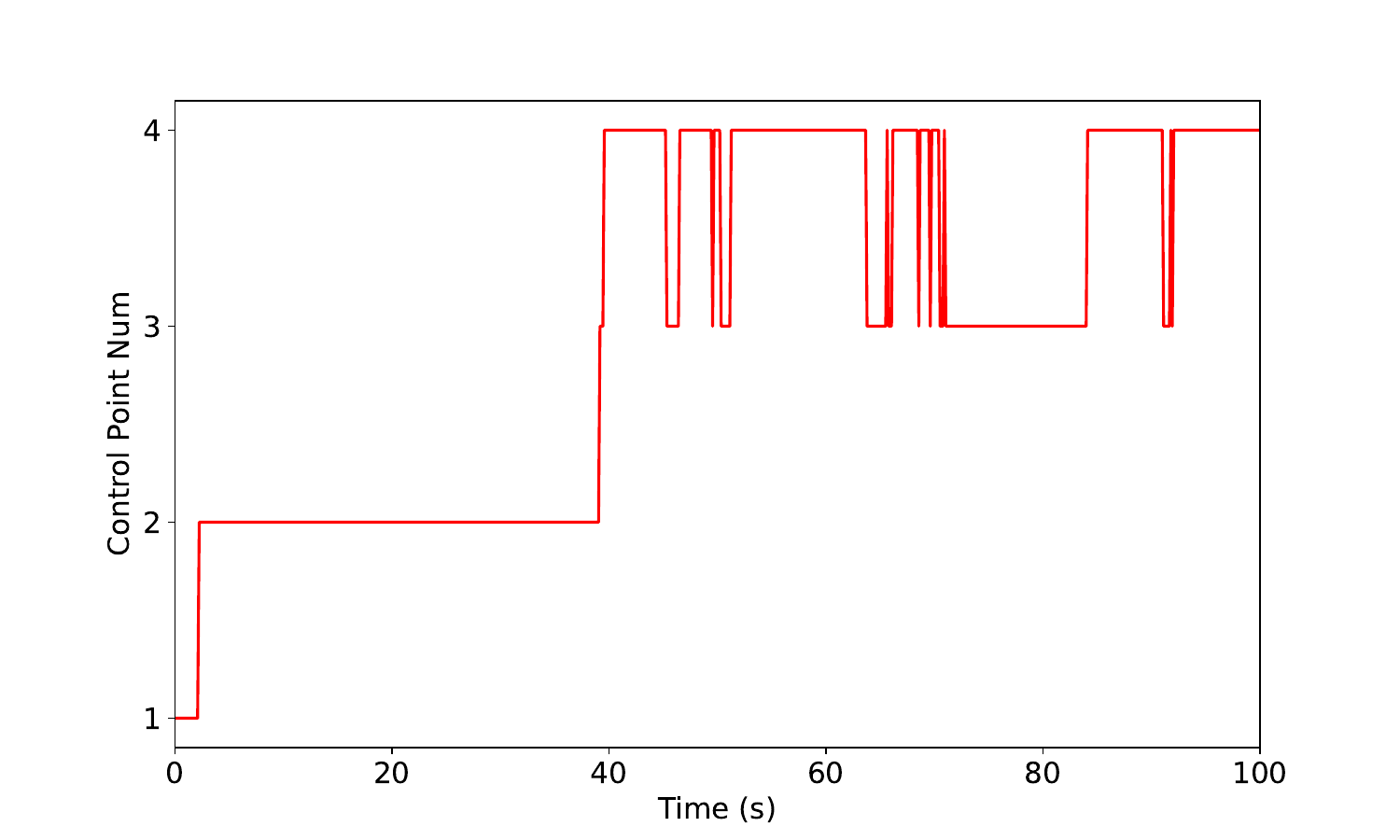} }
	\caption{Illustrations of the proposed adaptive knot span adjustment strategy. The top shows the robot's velocity, and the bottom shows that the distribution of control points is adaptive to speed variations.}
	\label{cp} 
\end{figure}

Namely, the current pose is registered as a Key Pose (KP), if the tracked trajectory segment has a local translational velocity or rotational velocity change above a certain threshold. 
Specifically, we obtain the translational velocity and rotational velocity respectively in a sliding window:
\begin{equation}
\begin{aligned}
\Delta v(T_{i-1},T_i) &= \frac{1}{n(T_{i-1},T_i)} \sum_{k \in n(T_{i-1},T_i)} &  \\ \Big(v_k 
& - \frac{1}{n(T_{i-1},T_i)} \sum_{k \in n(T_{i-1},T_i)} v_k \Big)^2,
\\
\Delta \omega(T_{i-1},T_i) &= \frac{1}{n(T_{i-1},T_i)} \sum_{k \in n(T_{i-1},T_i)} &  \\ \Big( \omega_k\
& - \frac{1}{n(T_{i-1},T_i)} \sum_{k \in n(T_{i-1},T_i)} \omega_k \Big)^2,
\end{aligned}
\end{equation}
where $n(T_{i-1},T_{i})$ represents the number of samples within a sliding window.
We associate each KP in $[T_{i-1},T_i]$ with the number of control points 
\begin{math} n_{cp}\end{math}
\begin{equation}
n_{cp} =
\begin{cases}
1, & \text{if } \Delta v \in [0, \lambda_{v1})  \text{ or } \Delta \omega \in [0, \theta_{\omega_1} ),\\
2, & \text{if } \Delta v \in [\lambda_{v1} , \lambda_{v2}) \text{ or } \Delta \omega \in [\theta_{\omega_1} , \theta_{\omega_2} ),\\
3, & \text{if } \Delta v \in [ \lambda_{v2} , \lambda_{v3})  \text{ or } \Delta \omega \in [\theta_{\omega_2} , \theta_{\omega_3} ),\\
4, & \text{if } \Delta v \in [\lambda_{v3}, \lambda_{v4} )  \text{ or } \Delta \omega \in [\theta_{\omega_3} ,\theta_{\omega_4}).
\end{cases}
\end{equation}
{where $\lambda_1$, $\lambda_2$, $\lambda_3$, $\lambda_4$  define the translational-velocity-change bins, and  $\theta_1 $, $\theta_2 $, $\theta_3 $, $\theta_4 $ define  the rotational-velocity-change bins.}
We know that the knot span is actually influenced by the number of control points at a specific time interval. Each of the knot spans is independent of the others, and the knot spans are divided by control points:
\begin{equation}\bigtriangleup t_i=\frac{1}{n_{cp}}(T_i-T_{i-1}).\end{equation} 
The non-uniform nature of the B-spline, characterized by its dynamic knot span adjustment strategy, contributes to the generation of smooth and adaptable trajectories that can better accommodate the dynamics of motion, thereby, improving the rationality of knot assignment.

Then, the trajectory segment over the sliding window \begin{math}[T_{i-1},T_i]\end{math} consists of \begin{math}n_{cp}\end{math} piecewise
functions. We can query the pose with an arbitrary timestamp \begin{math}t_{i}\end{math} in this segment. 
The control points of the active trajectory segment are further optimized and updated within the sliding window by a factor graph optimization. 
Meanwhile, control points outside the active segment remain static in the optimization but still retain marginalization priors for the next sliding window.

\subsection{{Backend}}
\label{sec:optimization}
\subsubsection{Virtual Anchor Ranging Factor}
Through constructing three non-collinear VAs, a global position $\textbf{x}_t$ of the robot at time $t$ can be obtained. Then, the residual of the virtual anchor ranging factor can be written as:
\begin{equation}
    \textbf{e}_{t}^{r} =  \sum_{m\in M}r_t^m - \left \| \textbf{x}_t - \hat{\textbf{p}}_{n}^{m} \right \|_2,
\end{equation}
where $M$ is the  set of location information $\hat{\textbf{p}}_{n}^{m}$ of the VAs detected at time $t$.

\subsubsection{IMU Factor}
\label{imu_factor}
If raw IMU measurements are directly used to formulate the IMU factor, the optimization processes incur high computational costs. 
To avoid this, we use the IMU pre-integration between two sequential UWB ranging measurements. Given IMU angular velocity and linear acceleration \begin{math}\left \{ (\boldsymbol{\omega}_I,\textbf{a}_I)_t \right \}\end{math} and the start robot state \begin{math}\textbf{x}_{t-1}\end{math}, we have
\begin{equation}
\textbf{e}_{t}^\text{IMU}=\left \| \textbf{x}_t-\xi (\textbf{x}_{t-1},\boldsymbol{\omega}_I(t),\textbf{a}_I(t),\varepsilon_g,n_g,n_a ) \right \|  _2,
\end{equation}
where \begin{math}\xi(\cdot)\end{math} represents the pre-integration of IMU measurements.
\begin{math}\boldsymbol{\omega}_I(t)\end{math} and \begin{math}
\textbf{a}_I(t)\end{math} can be obtained from the derivatives of the continuous-time B-splines in Eq.~\eqref{liegroup}, respectively. 
\begin{equation}
\begin{aligned}
\mathbf{a}_{I}(t) & ={ }_{I}^{W} \mathbf{R}^\text{T}(t)\left(\ddot{\mathbf{p}}_{I}(t)+g^{W}\right)+\varepsilon_{a}+n_{a}, \\
\boldsymbol{\omega}_{I}(t) & =\left({ }_{I}^{W} \mathbf{R}^\text{T}(t){ }_{I}^{W} \dot{\mathbf{R}}(t)\right)^{v}+\varepsilon_{g}+n_{g},
\end{aligned}
\end{equation}
where \begin{math}(\cdot)^v\end{math} maps the skew-symmetric matrix to the corresponding 3D vector, \begin{math}\ddot{\mathbf{p}}_{I}(t) \end{math} represents the translational component submatrices of the second order derivatives, 
\begin{math}_{I}^{W} \dot{\mathbf{R}}(t)\end{math} is the rotational component submatrices of the first order derivatives. The continuous-time trajectory of the IMU in the world frame is denoted as:
\begin{equation}
    { }_{I}^{W} \mathbf{T}(t)  =\left[{ }_{I}^{W} \mathbf{R}(t),{ }_{I}^{W} \mathbf{p}(t)\right].
\end{equation}

\subsubsection{Odometer Factor}
We assume all intrinsic and extrinsic parameters in UWB/IMU/odometer are pre-calibrated~\cite{bai2022graph,LI2024115186}, so we can handily get odometer trajectory  \begin{math}_{O}^{W}\mathbf{T}(t)\end{math}. 
The odometer factor can be defined as
\begin{equation}
\mathbf{e}_{t}^\text{odom}=\left\|\mathbf{x}_{t}-\psi\left(\mathbf{x}_{t-1}, \boldsymbol{\omega}_{o}(t), \mathbf{v}_{o}(t), \eta^{o}\right)\right\|_{2},
\end{equation}
where \begin{math}\psi(\cdot)\end{math} denotes the pre-integration of odometer measurements, 
\begin{math}\boldsymbol{\omega}_o(t)\end{math} and \begin{math}\boldsymbol{v}_o(t)\end{math} can be computed from the derivatives of the continuous-time trajectory, respectively.
\begin{equation}
\omega_{o}(t)=\left({ }_{o}^{W} \mathbf{R}^\text{T}(t){ }_{o}^{W} \mathbf{R}(t)\right)^{v}+\eta^{o}, 
\end{equation}
\begin{equation}
\mathbf{v}_{o}(t)={ }_{o}^{W} \mathbf{R}^\text{T}(t) \dot{\mathbf{p}}_{o}(t)+\eta^{o}.
\end{equation}
\begin{math}\dot{\mathbf{p}}_{o}(t)\end{math} represents the translational component submatrices of the first-order derivatives. 

\subsubsection{{Adaptive Sliding Window Factor Graph Optimization}
}
\label{sec:Sliding-Window}

We design a continuous-time factor graph optimization for robot trajectory estimation based on the association between UWB/IMU/odometer data streams. 
Non-uniform B-splines are employed to obtain the optimal trajectory estimation continuously over time. 
To keep the computation tractable for online performance, an adaptive sliding window is designed to maintain a limited number of control points. 
The window length is adaptively updated based on the control point density as follows: 
\begin{equation}
\Delta L=\left\{\begin{array}{lc}
L_{\min } & \alpha_{cp} \geq \lambda_{1}, \\
\left \lfloor \beta \frac{\left(L_{\max }-L_{\min }\right)\left(\lambda_{1}-\alpha_{cp}\right)}{\lambda_{1}-\lambda_{0}} \right \rfloor  & \lambda_{0}<\alpha_{cp}<\lambda_{1}, \\
L_{\max } & \alpha_{cp} \leq \lambda_{0}.
\end{array}\right. \\
\end{equation}
Here, \begin{math}\alpha_{cp}\end{math} is the control point density
\begin{equation}
\alpha_{cp}=\frac{n}{\triangle T}.
\end{equation}
\begin{math}\triangle T\end{math} is the given time interval, $n$ is the number of control points in the trajectory segment in \begin{math}[t,t+\triangle T]\end{math}. {  $\left \lfloor \cdot  \right \rfloor$ is the floor function, 
\begin{math}\lambda_0\end{math} and \begin{math}\lambda_1\end{math} denote the upper and lower thresholds of \begin{math}\alpha_{cp}\end{math}, \begin{math}L_\text{min}\end{math} and \begin{math}L_\text{max}\end{math} denote the minimum and maximum length values of the sliding window, respectively. }
The settings of \begin{math}L_\text{min}\end{math} and \begin{math}L_\text{max}\end{math} are adjusted based on the intrinsic correlations of historical data and the system’s available resources. \begin{math}\beta\end{math} is a constant related to the IMU measurement frequency.

The factor graph optimization is solved, consisting of UWB ranging factor, IMU factor, odometer factor, and the optimal state \begin{math}\hat{\mathbf{X}}\end{math} over an adaptive window $[t-\tau,t]$ of recent $\Delta L$ control points can be solved as follows:
\begin{equation}
\begin{split}
\hat{\mathbf{X}} = \arg \min \sum_{k=t-\tau }^{t} \Bigg( &\left\|\mathbf{e}_{k}^r\right\|_{\Sigma_{k}^r}^{2} + \left\|\mathbf{e}_{k}^{\text{IMU}}\right\|_{\Sigma_{k}^{\text{IMU}}}^{2}, \\
&+ \left\|\mathbf{e}_{k}^{\text{odom}}\right\|_{\Sigma_{k}^{\text{odom}}}^{2} + \left\|\mathbf{e}_{k}^{\text{Prior}}\right\|_{\Sigma_{k}^{\text{prior}}}^{2} \Bigg),
\end{split}
\end{equation}
\begin{equation}
\mathbf{X}^{\text {prior }}=\left\{\mathbf{x}_{t-\tau}, \ldots, \mathbf{x}_{t-\tau+L_w}\right\},
\end{equation}
where \begin{math}\sum_{k}^r\end{math},
\begin{math}\sum_{k}^\text{IMU}\end{math},
\begin{math}\sum_{k}^\text{odom}\end{math},
\begin{math}\sum_{k}^\text{Prior}\end{math} is the covariance matrix associated with UWB ranging, IMU, odometer measurements, and prior factor from marginalization, respectively. 
\begin{math}\textbf{X}^\text{prior}\end{math} represents the prior factor in $[t-\tau,t-\tau+L_w]$ shared by adjacent adaptive sliding windows. We maintain a temporal sliding window within an adaptive length, where the active control points within the current window are optimized and updated. As the adaptive window slides forward, old control points are removed from the sliding window.  However, they are still involved in constructing residuals within the current window {to ensure continuity} in motion estimation.

\begin{figure}[!t]
\centering
\includegraphics[width=\hsize,trim=0.1cm 1.2cm 0.1cm 0.1cm,clip]{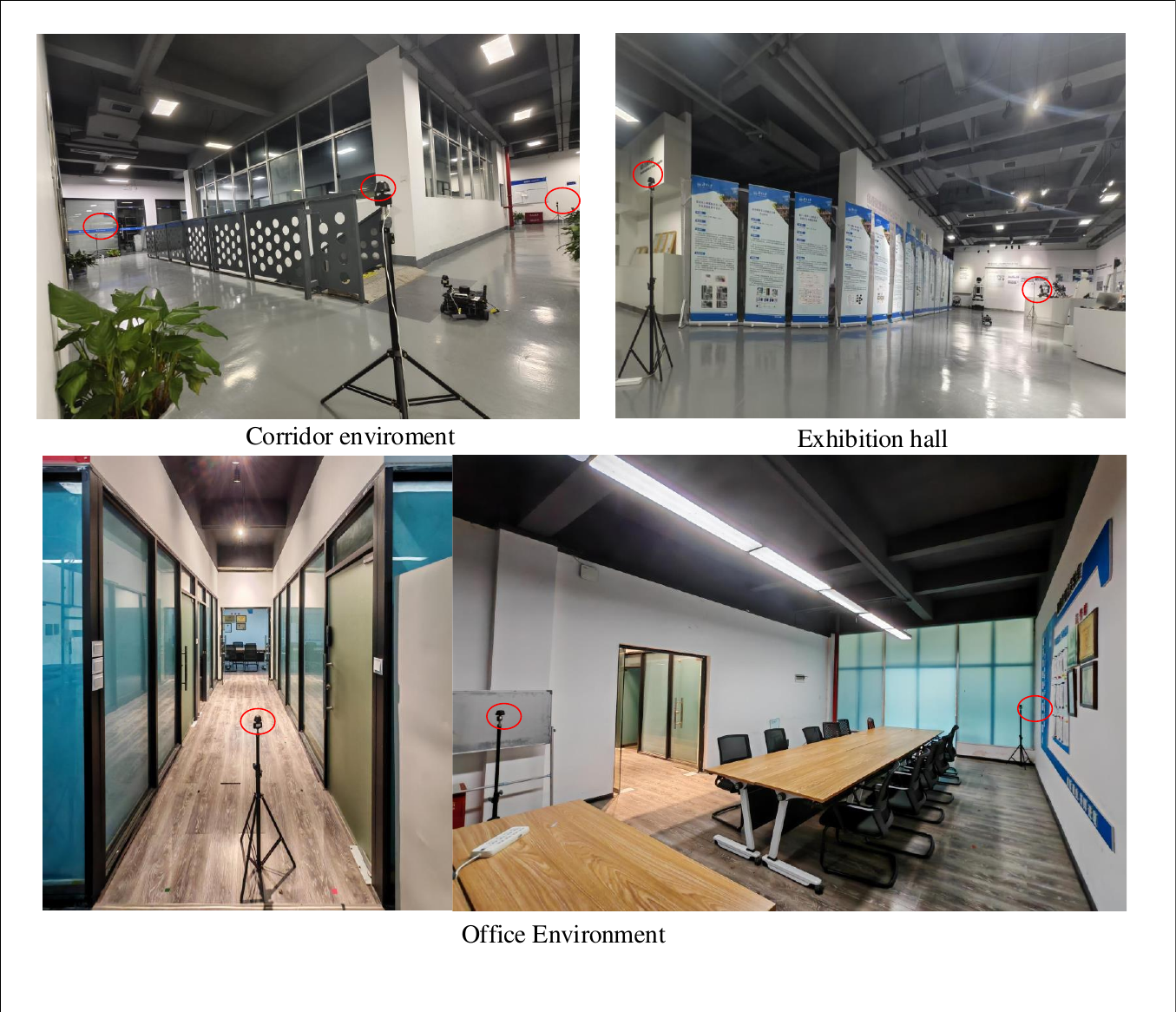}
\caption{{The experimental configurations in the corridor, exhibition hall, and office environments. The red circles represent the positions of the UWB anchors.}}
\label{fig_1}
\end{figure}

\section{Experiments}
\label{experiments}
\subsection{Experimental Setup}

Considering that there are no publicly available datasets for the UWB-Inertial-odometer system, we validated the effectiveness of the proposed CT-UIO on self-collected datasets.
{Experimental campaigns are conducted in the corridor, exhibition hall, and office scenes with different numbers of UWB anchors}, as shown in Fig.~\ref{fig_1}.
The UWB tag, namely, Nooploop LinkTrack, IMU, and odometer are integrated into the Turtlebot3 Waffle Pi, as shown in Fig.~\ref{fig_0}.   
{Before localization, the parameters of the UWB/IMU/odometer system have been calibrated. In LOS case, the overall UWB ranging error follows a Gaussian distribution, which can be fitted using $\mathcal{N}( -0.0552, 0.0514^2)$(unit: m).  The odometer scale factor error is set to be 0.5\%. The specification and noise characteristics of the IMU are  summarized in Table ~\ref{tab:bias}.}

\begin{table}[htbp]
  \centering
  \caption{Noise components in IMU}
  	\resizebox{0.48\textwidth}{!}{
    \begin{tabular}{cccccc}
    \toprule
     \makecell[c]{Noise \\Description}     & \makecell[c]{Velocity \\Random walk}  & Acc. Bias & \makecell[c]{Angular Random\\ walk}  &Gyro Bias   \\
    \midrule
    IMU & { 0.035  $m/s/\sqrt{h} $} & 0.2 $mg$ & 1.03 $deg/h$ & 8.23 $deg/h$ \\
  
    \bottomrule
    \end{tabular}%
    }
  \label{tab:bias}%
\end{table}%
}

The measurement rate for each UWB anchor is set to $32Hz$, while odometer and IMU measurements are recorded at $28Hz$ and $127Hz$, respectively. 
UWB anchors are mounted on the tripods at a preset height, and the position of the UWB anchors deployed in the environment has been calibrated.

{The proposed CT-UIO is compared with five state-of-the-art UIO methods in the same experimental environments. These include discrete-time methods (Optimization+Filtering~\cite{yang2023novel}, Optimization+Trust Region~\cite{zhou2024optimization}, GPDA~\cite{WOS:001173317800046}, and VMHE~\cite{9829196}) and uniform B-spline-based continuous-time method (SFUISE~\cite{li2023continuous}).} 
The robot trajectories, generated by the LaMa framework\footnote{https://github.com/iris-ua/iris\_lama} through the fusion of  LIDAR and IMU data,  are regarded as the ground truth. 
{All methods are designed with C++ and operated on the robot with the CPU of ARMv8 (NVIDIA JetsonTX2 @ 32 GHz, with 8 GB memory) in Melodic ROS (Robot Operating System)}. 
{In all experiments, the specific parameters of CT-UIO are listed in Table~\ref{tab_notations}. These parameter values are determined based on the noise characteristics of the onboard sensors and the computational constraints of the robot’s embedded processor. The selection aims to balance the trade-offs between localization accuracy and computational efficiency.} 
{To further investigate the influence of dataset characteristics on the selection of $\tau_F$, we performed the parameter analysis separately for the three datasets with varying numbers and layouts of anchors. 
For each dataset, a parameter sweep of $\tau_F$ was conducted on the MATLAB platform using a series of simulations with various VAs' positions. For each configuration, we evaluated the initialization success rate, the number of UWB observations, the time-to-initialization, and the waypoints obtained from the IMU/odometer fusion model. The optimal  threshold  $\tau_F$ is then determined as the mid-point of the margin of minimum eigenvalues of well-conditioned waypoints and ill-conditioned ones.}

\begin{table}
\caption{{Values of parameters used in experiments.}}%
\label{tab_notations}
\centering%
\begin{tabular}{p{5cm}c}
   \toprule
   Parameter & Value  \\
   \midrule
 {A constant related to the IMU measurement noise frequency $\alpha$}  & \makecell{$\alpha_\text{acc}=1.22\times 10^{-4}$\\ $\alpha_\text{gyro}=3.17\times 10^{-3}$}\\
 {Bounds for the $e(k)$} & \makecell{ $\lambda_\text{min}=0.56$\\ $\lambda_\text{max}=6.52$}\\
 {The maximum size of IMU observation statistics} & $\xi =50$\\
 {Threshold for the rejection domain of  \begin{math}E(k)\end{math}} & $thr = 10.85$\\
 { Outlier rejection threshold for the individual ranging measurement} &$\gamma = 1.25$ \\
 {The lower bound of the VA's position estimation error} & \makecell{Corridor :$\tau_F=25$\\Exhibition Hall :$\tau_F=44$\\Office:$\tau_F=16$}\\
 {Bounds for the sliding window} & \makecell{$L_\text{min}=5$\\$L_\text{max}=30$}\\
 {Bounds for the control point density $\alpha_{cp}$ }& \makecell{$\lambda_0=0.02$\\$\lambda_1=0.1$}\\
 The translational-velocity-change bins & \makecell{$\lambda_{v_1}=0.06,\lambda_{v_2}=0.12$ \\$\lambda_{v_3}=0.20,\lambda_{v_4}=0.26$}\\
 {The rotational-velocity-change bins} & \makecell{$\theta_{\omega_1}=0.20,\theta_{\omega_2}=0.50$ \\$\theta_{\omega_3}=0.13,\theta_{\omega_4}=1.82$}\\
   \bottomrule
\end{tabular}
\end{table}

The basic performance metric for CT-UIO evaluation by evo\footnote{https://github.com/MichaelGrupp/evo} is Absolute Positioning Error (APE). This metric is used to calculate the position deviation between the estimated robot trajectory and the ground truth trajectory after aligning to the same coordinate system. 

\subsection{{Evaluation  on the Corridor Dataset}}

\begin{figure*}[!t]
  \centering
  \subfloat[A-S1\label{fig:A-S1}]{
    \includegraphics[width=6cm]{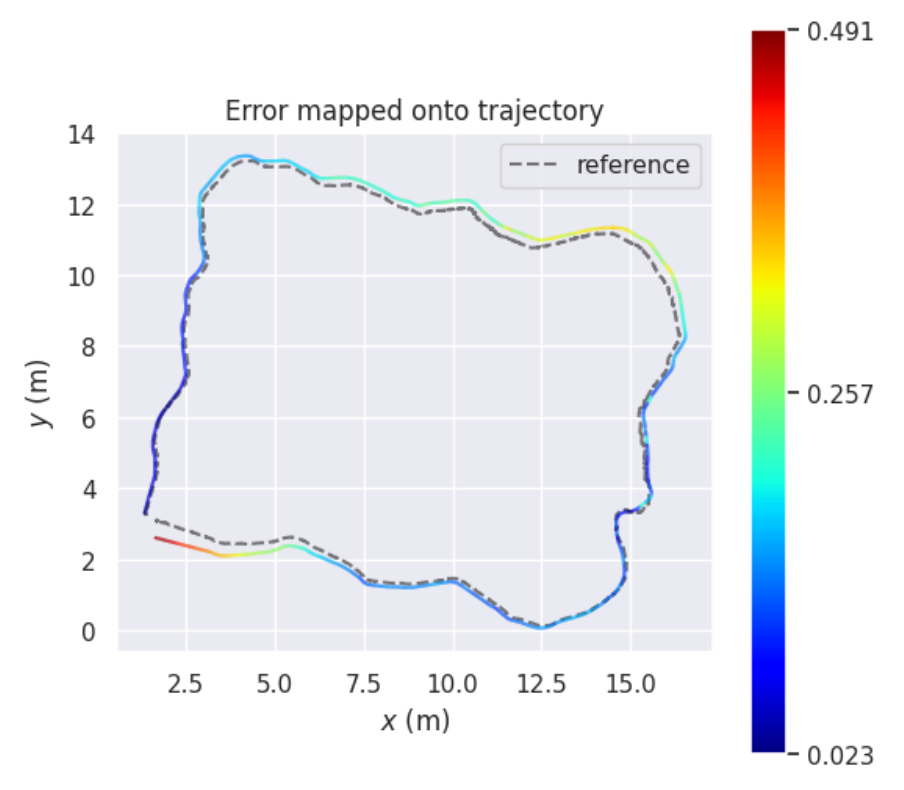}
  }\hspace{-8mm}
  \subfloat[A-S2\label{fig:A-S2}]
  {
    \includegraphics[width=6cm]{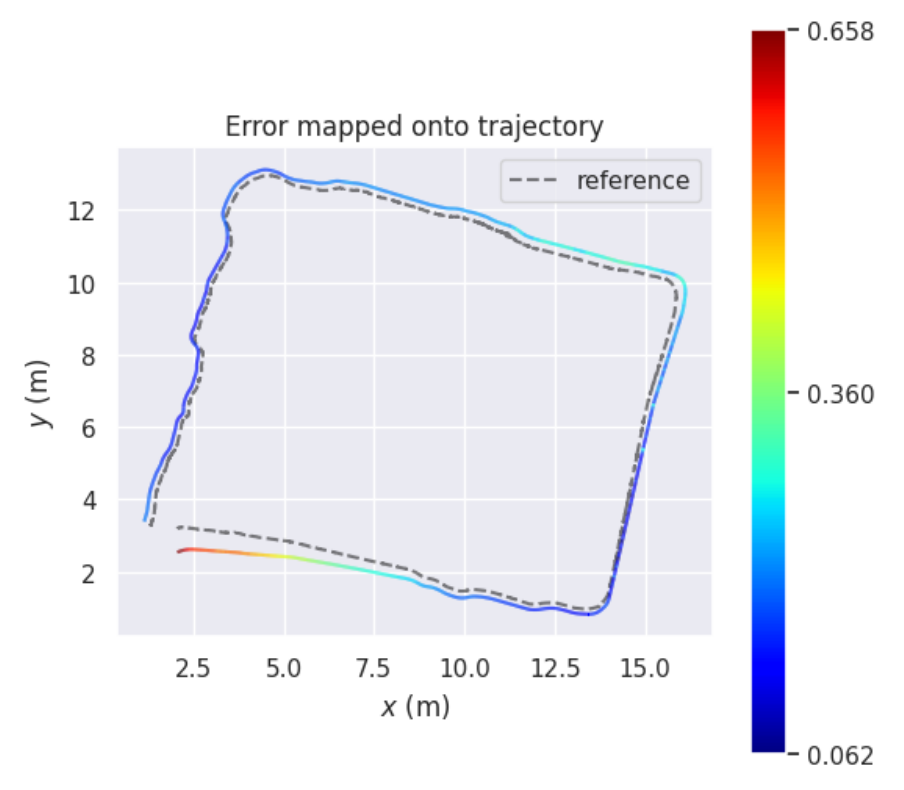}
  }\hspace{-5mm}
  \subfloat[A-J1\label{fig:A-J1}]
  {
    \includegraphics[width=6cm]{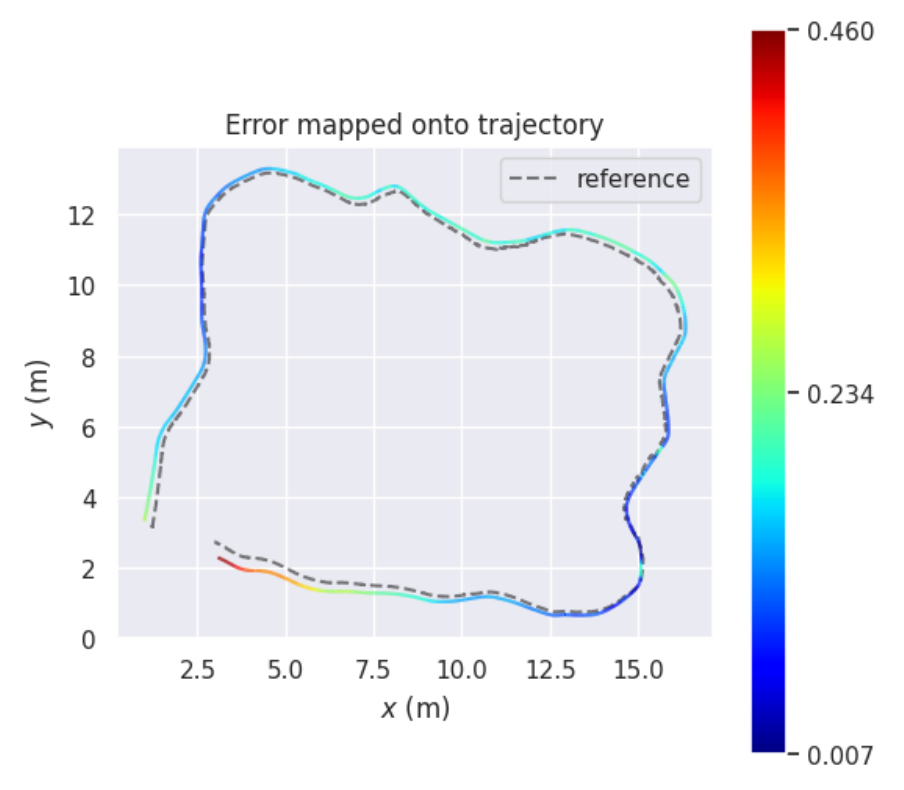}
}

  \vspace{-4mm}
  \subfloat[A-J2\label{fig:mad_a}]{
    \includegraphics[width=6cm]{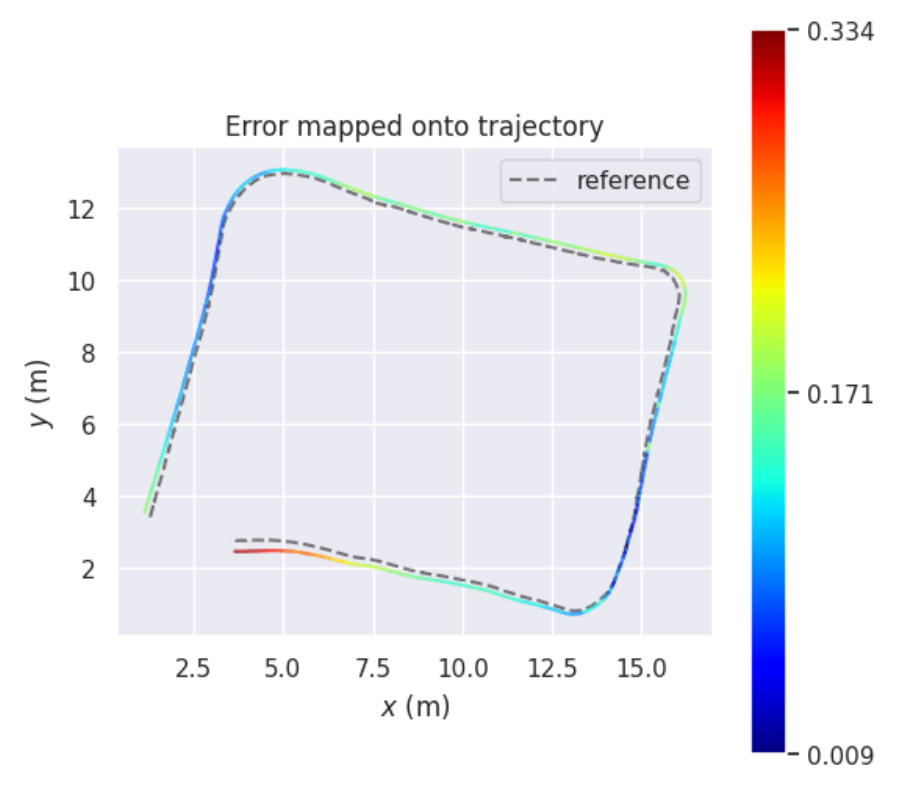}
  }\hspace{-5mm}
  \subfloat[A-H1\label{fig:irmad_a_h1}]
  {
    \includegraphics[width=6cm]{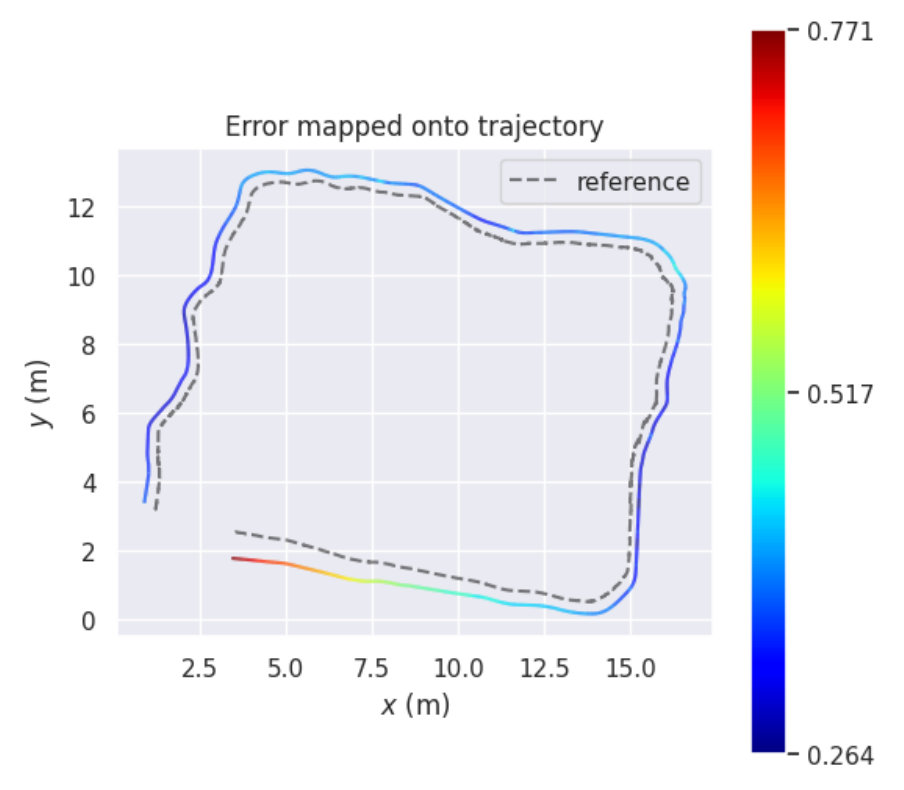}
  }\hspace{-5mm}
  \subfloat[A-H2\label{fig:isfa_a_h2}]
  {
    \includegraphics[width=6cm]{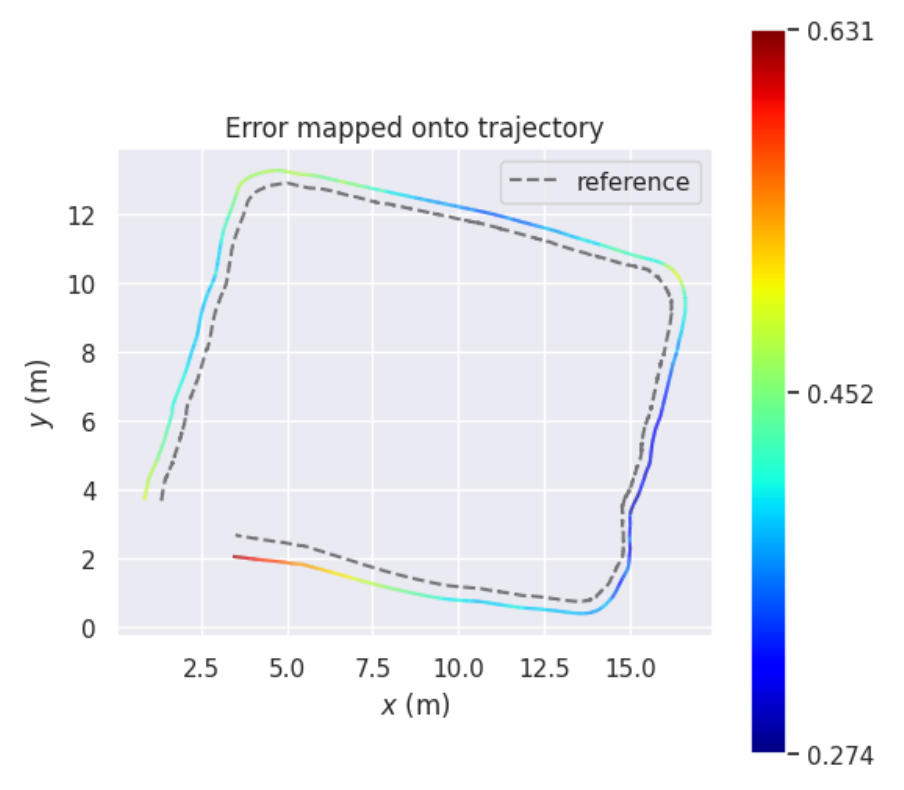}
  }
  \caption{We query the estimated trajectory from the obtained continuous-time trajectory and evaluate it against the ground-truth trajectory.
  {The ground-truth trajectory is shown as a black dashed line, and the estimated trajectory is colored by absolute translation error when compared against the ground truth in the corridor dataset}. 
  Our proposed CT-UIO not only achieves high accuracy in trajectory estimation but also effectively fits translational velocity variations.}
  \label{fig:A}
\end{figure*}
The proposed algorithm is first evaluated in the corridor environment, where four UWB anchors are set within a $16.2m{\times}15.6m$ area. 
These anchors are placed at the corners of the area, with coordinates $(0.88,2.99)m$, $(3.36,13.64)m$, $(15.01,{-}0.19)m$, and $(17.03,10.65)m$, respectively. 
Due to physical barriers, such as walls or metal structures, between  UWB anchors, only one or two UWB anchors are available under this non-fully observable condition. The robot's motion mode is varied by adjusting its linear acceleration. A total of six sequences are collected, consisting of three types of motion modes, that is slow, fast, and hybrid, as shown in Table~\ref{table0}.

\begin{table}[!htb]
\centering
\caption{Description of the sequences of the corridor dataset.}
\label{table0}
\setlength{\tabcolsep}{0.5mm}
\begin{tabular}{ccccc}
\toprule
\makecell{\textbf{Seq}} & 
\makecell{\textbf{Duration} \\ (second)} & 
\makecell{\textbf{Length} \\ {(m)}} & 
\makecell{\textbf{Translational velocity} \\ {max/min (m/s)}} & 
\makecell{\textbf{Description}} \\
\midrule
A-S1 & 380.1 & 65.03 & 0.18/0.13 & Slow sequences \\
A-S2 & 376.0 & 62.99 & 0.16/0.12 & Slow sequences \\
A-J1 & 214.9 & 53.66 & 0.26/0.22 & Fast sequences \\
A-J2 & 199.2 & 51.49 & 0.26/0.22 & Fast sequences \\
A-H1 & 251.0 & 54.64 & 0.26/0.18 & Hybrid sequences \\
A-H2 & 232.7 & 52.55 & 0.26/0.16 & Hybrid sequences \\
\bottomrule
\end{tabular}
\end{table}
\begin{itemize}
\item Slow sequences: The robot moves gently at a slow translational velocity with gentle speed variations, following a square path.
\item Fast sequences: The robot moves at a large translational velocity with intense speed variations, following a square path. 
\item Hybrid sequences: 
The robot follows a square trajectory that combines both slow and fast translational motion, incorporating a mix of low and high-speed phases.
\end{itemize}

\begin{figure*}[htbp]
  \centering
  \subfloat[A-S1\label{fig:cdf:A-S1}]{
    \includegraphics[width=6cm,]{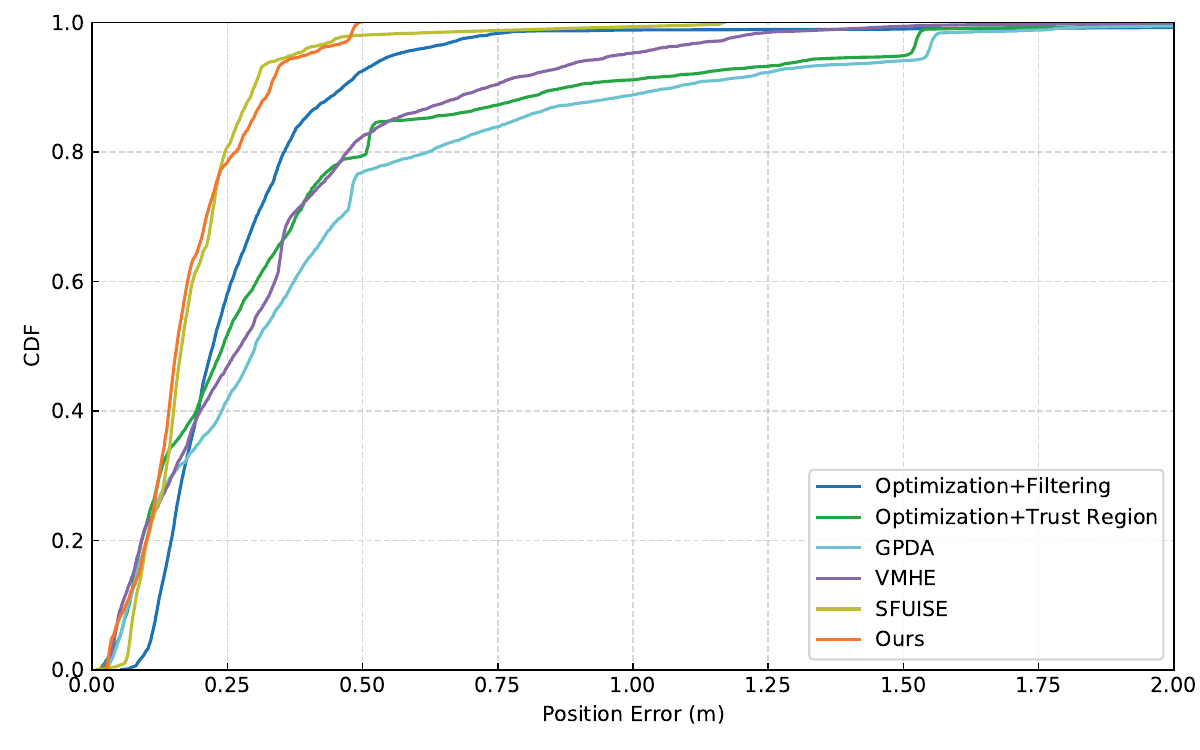}
  }\hspace{-5mm}
  \subfloat[A-S2\label{fig:cdf:A-S2}]
  {
    \includegraphics[width=6cm]{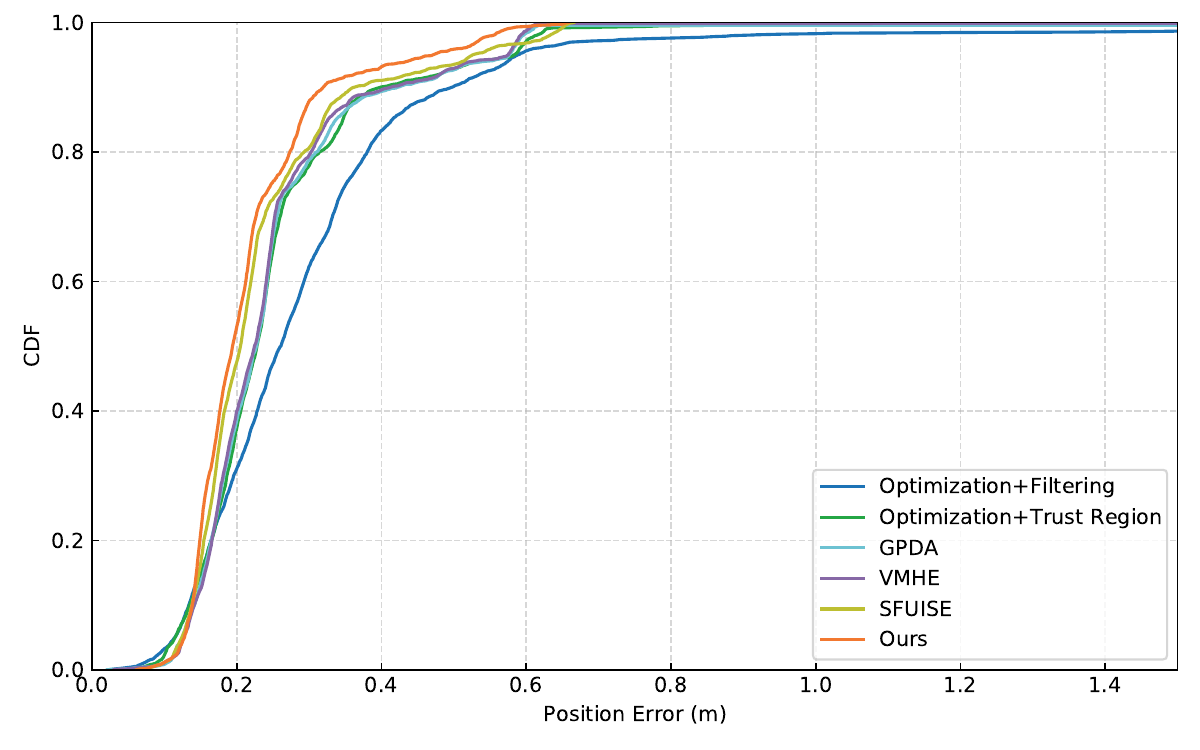}
  }\hspace{-5mm}
  \subfloat[A-J1\label{fig:cdf:A-J1}]
  {
    \includegraphics[width=6cm]{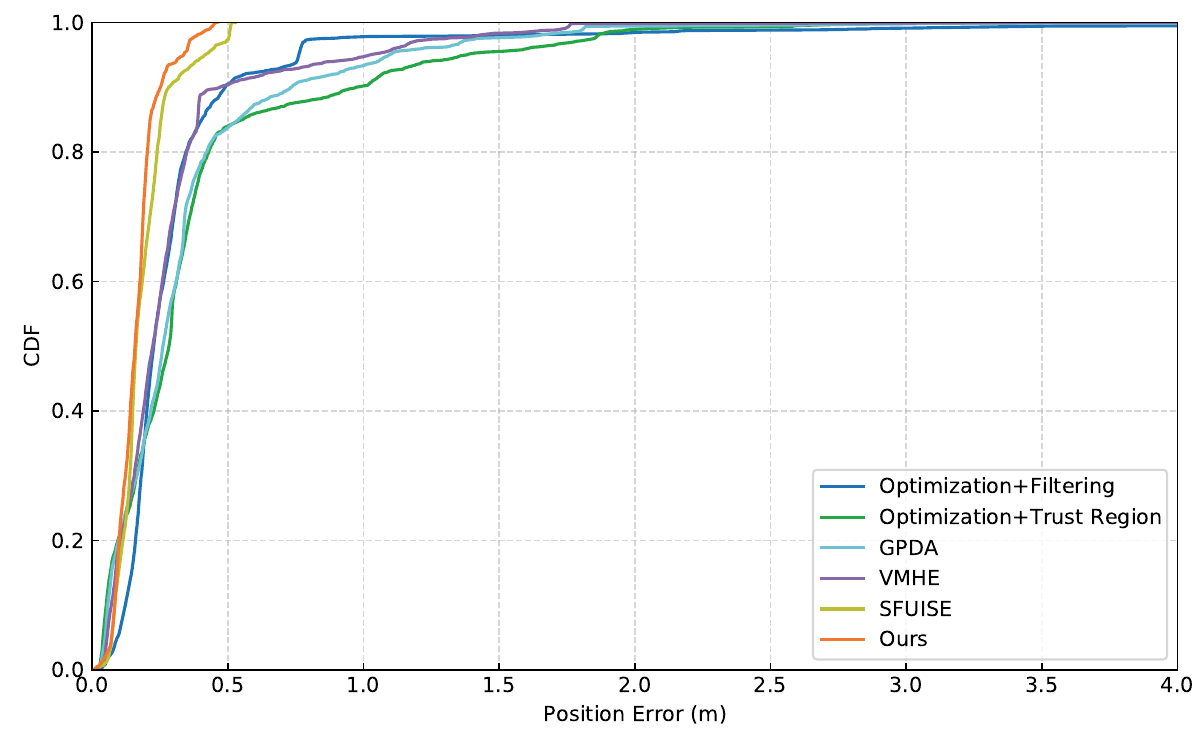}
}
  \vspace{-4mm}
  \subfloat[A-J2\label{fig:cdf:A-J2}]{
    \includegraphics[width=6cm]{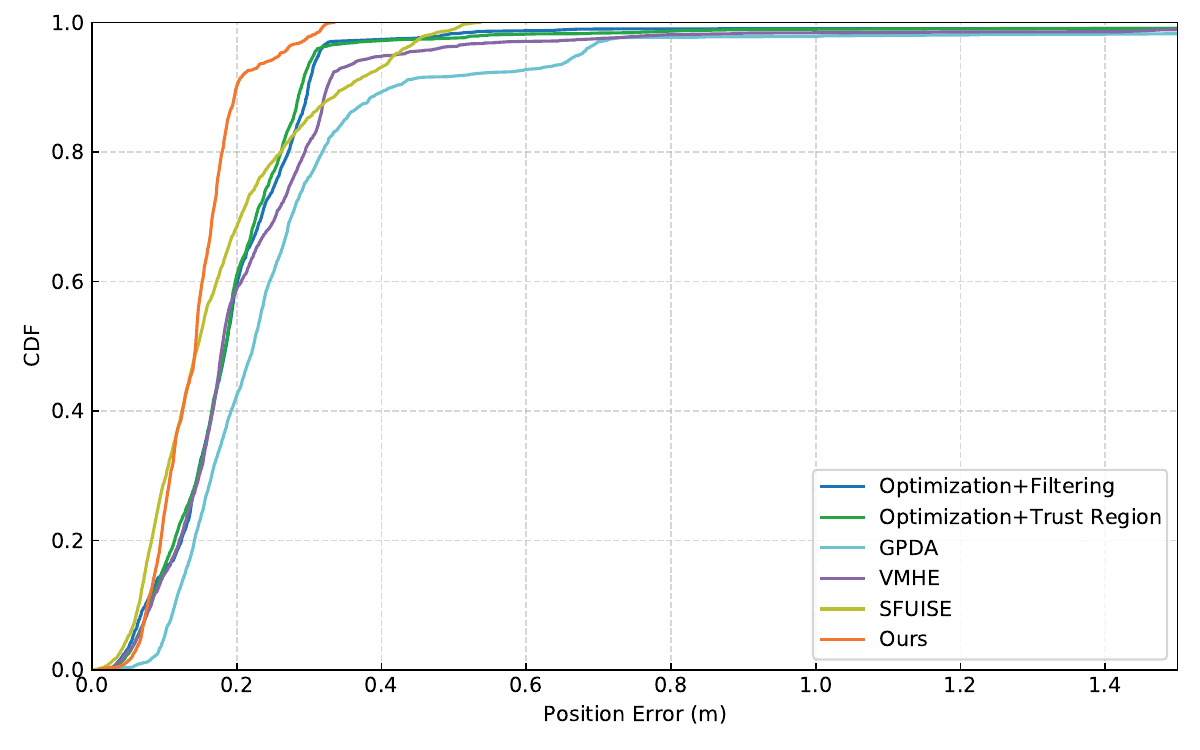}
  }\hspace{-5mm}
  \subfloat[A-H1\label{fig:cdf:A-H1}]
  {
    \includegraphics[width=6cm]{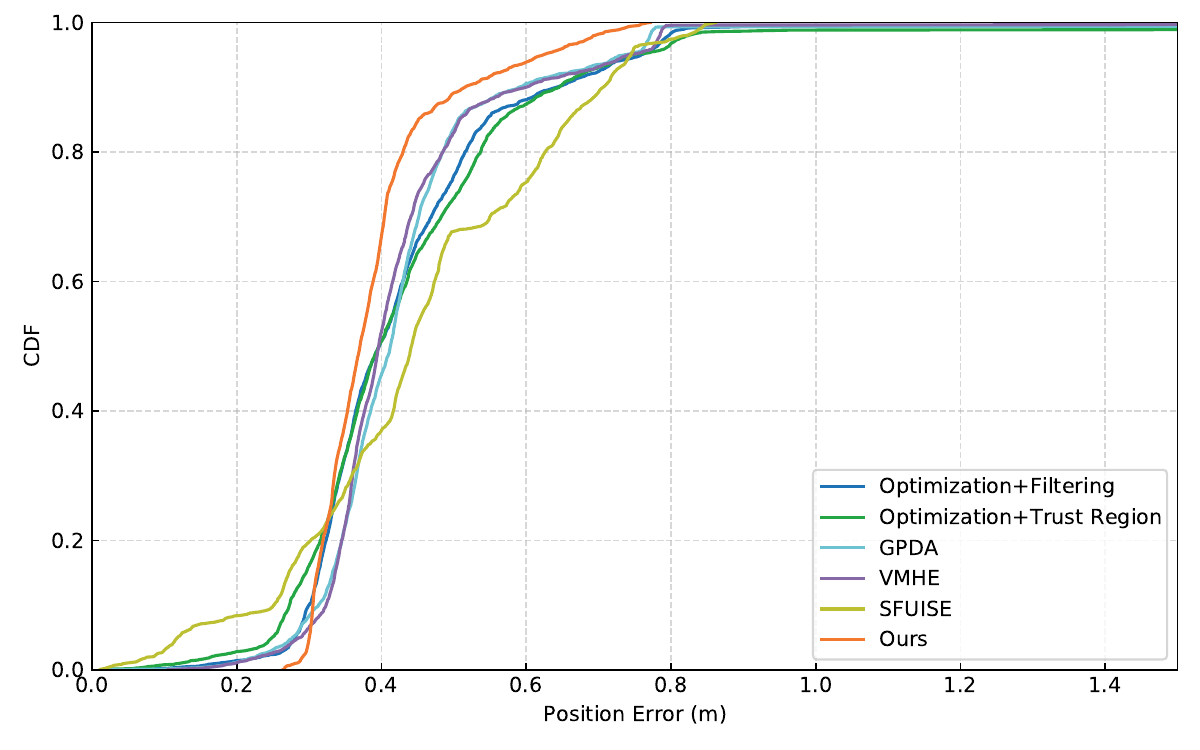}
  }\hspace{-5mm}
  \subfloat[A-H2\label{fig:cdf:A-H2}]
  {
    \includegraphics[width=6cm]{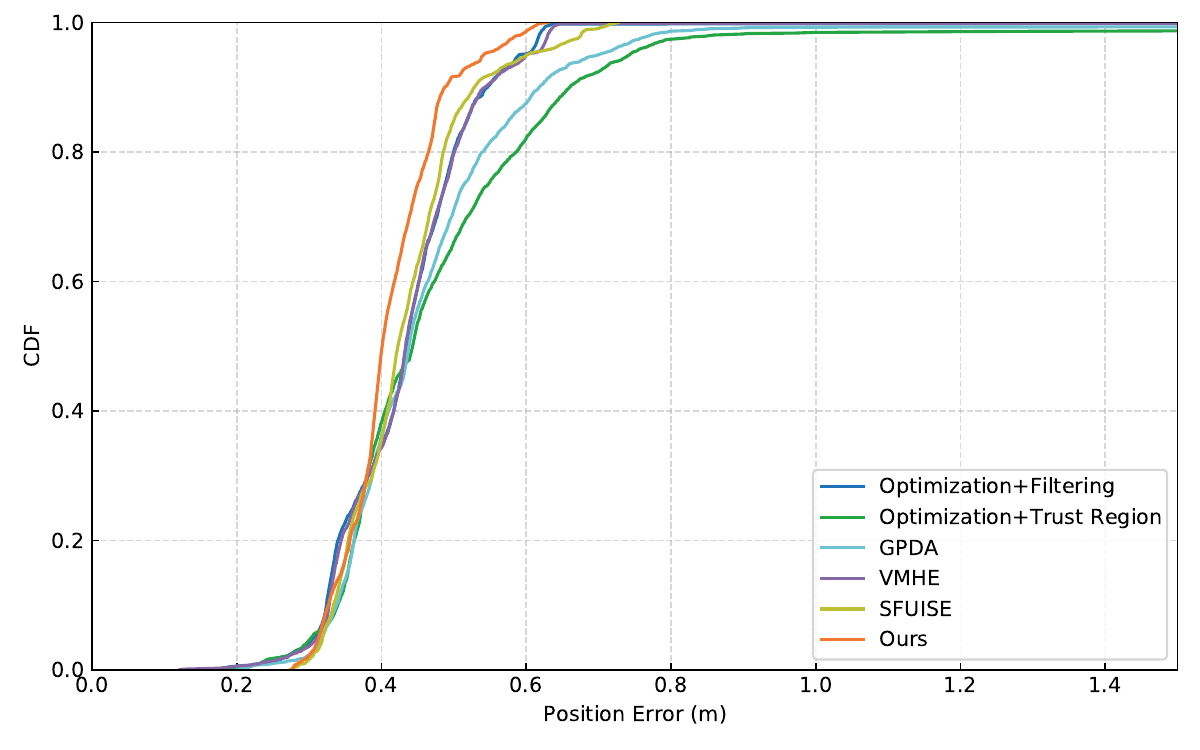}
  }
  \caption{{CDFs of APE for different methods on the corridor dataset. The APE values of our method are more concentrated below $0.5m$  compared to the other five methods. This result further demonstrates the effectiveness of the proposed method in achieving a better localization accuracy due to its adaptive knot span adjustment strategy and VA generation mechanism.}}
  \label{fig:A:CDF}
\end{figure*}
We compute the RMSE of APE error for each sequence across different methods. 
\begin{table}[!htb]
\centering
\caption{Performance comparison in APE (RMSE, Meter) of different methods on the corridor dataset. The best results are marked in bold.}
\label{table1}
\setlength{\tabcolsep}{1.7mm}
\begin{tabular}{ccccccc}
\toprule
\makecell{\textbf{Seq}} & 
\makecell{A-S1 } & 
 A-S2& 
A-J1& 
A-J2&
A-H1 &
A-H2\\
\midrule
\makecell[c]{{Optimization+}\\{Filtering}\cite{yang2023novel}}& 0.756 & 0.605 & 0.644 & 0.623 & 0.608 & 0.548 \\
\makecell[c]{{Optimization+}\\{Trust Region}\cite{zhou2024optimization}} & 0.550 & 0.520 & 0.575 & 0.593 & 0.658 & 0.671 \\
 GPDA\cite{WOS:001173317800046}& 0.617   & 0.338   & 0.523  & 0.605 &0.537 & 0.590 \\
VMHE\cite{9829196} & 0.467  & 0.313  & 0.433   & 0.539 &0.501 &0.499\\
{{SFUISE}\cite{li2023continuous}} & 0.210 & 0.262 & 0.229 & 0.208 & 0.487 & 0.448 \\
Ours & \textbf{0.208} & \textbf{0.242} & \textbf{0.181} & \textbf{0.153} & \textbf{0.403} & \textbf{0.416}\\
\bottomrule
\end{tabular}
\end{table}

The results are summarized in Table~\ref{table1}, showing that our CT-UIO achieves the best accuracy when compared with the state-of-the-art methods. 
Due to the lack of sufficient available UWB anchors in the corridor environment, leading to fewer constraints in the pose graph optimization process and reduced localization accuracy for Optimization+Filtering~\cite{yang2023novel}, Optimization+Trust Region~\cite{zhou2024optimization} and GPDA\cite{WOS:001173317800046} methods. {Despite the additional radial velocity constraint introduced in VMHE~\cite{9829196}, the formulation still suffers from ambiguity and drift when the trajectory is dominated by tangential motion.} 
In contrast, our CT-UIO introduces VAs to ensure full observability in the few-anchor localization system. This allows for the incorporation of multiple constraints into the factor graph optimization, resulting in more accurate localization estimates. 

In sequences such as A-S1 and A-S2, the improvement in APE is relatively minor when compared with the uniform B-spline-based method SFUISE.
In these sequences, where the robot’s movement is relatively slow, the proposed adaptive knot span adjustment does not necessarily lead to a significant improvement in performance. However, in sequences involving fast linear motion, such as A-J1 and A-J2, the errors in SFUISE increase significantly.  In contrast, CT-UIO outperforms state-of-the-art methods in terms of the overall performance of pose interpolation.  
Across hybrid sequences, CT-UIO can improve localization accuracy significantly by $38.75\%$ in APE. 
This improvement is attributed to the adaptive knot span adjustment strategy, which allows dynamic adjustment of the number of control points by either lowering or increasing the knot span to accommodate varying motion speeds. 
Fig.~\ref{fig:A} shows the overview of trajectories in the corridor dataset, where our method achieves consistent moving trajectories in most sequences. 

\begin{figure}[!htb]
\centering
\includegraphics[width=\linewidth,trim=2cm 0.6cm 2.5cm 1.8cm,clip]{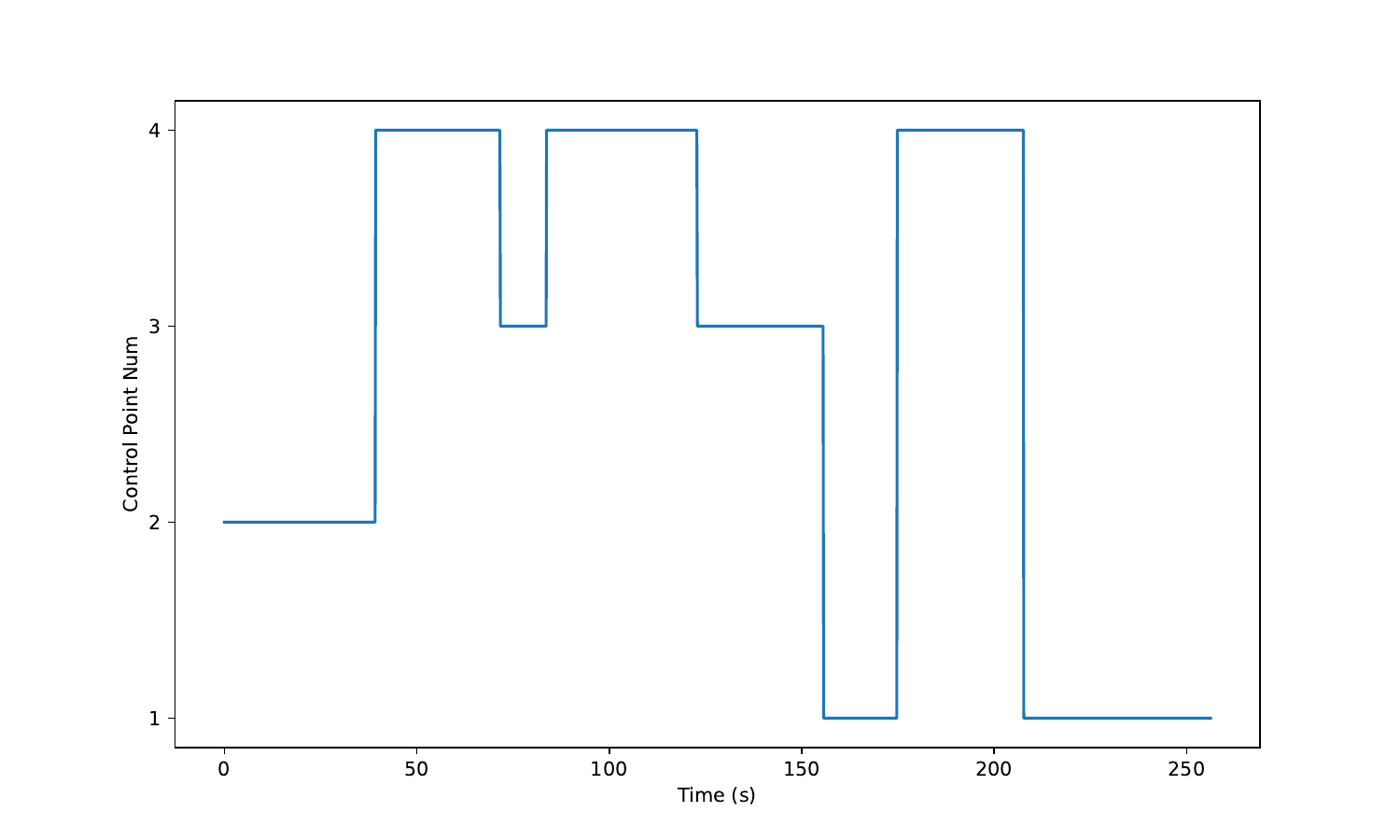}
\caption{The distribution of control points in sequence A-H1.
}
\label{A-h2-cp}
\end{figure}

{The positioning errors are further evaluated using cumulative distribution functions (CDFs), as shown in Fig.~\ref{fig:A:CDF}. Fig.~\ref{A-h2-cp} shows the illustration of the proposed adaptive knot span adjustment strategy in sequence A-H1.
For sequence A-H1, $90\%$ of the APE in our method are less than $0.515m$, while for {Optimization+Filtering}, {Optimization+Trust Region}, GPDA, VMHE, and {SFUISE}, the values are $0.643m$,  $0.645m$, $0.590 m$,  $0.642m$, and $0.706m$, respectively. The results demonstrate that our method achieves high levels of positioning accuracy.}
\begin{figure*}[!htb]
  \centering
  \subfloat[B-S1\label{fig:B-S1}]{
    \includegraphics[width=0.22\linewidth]{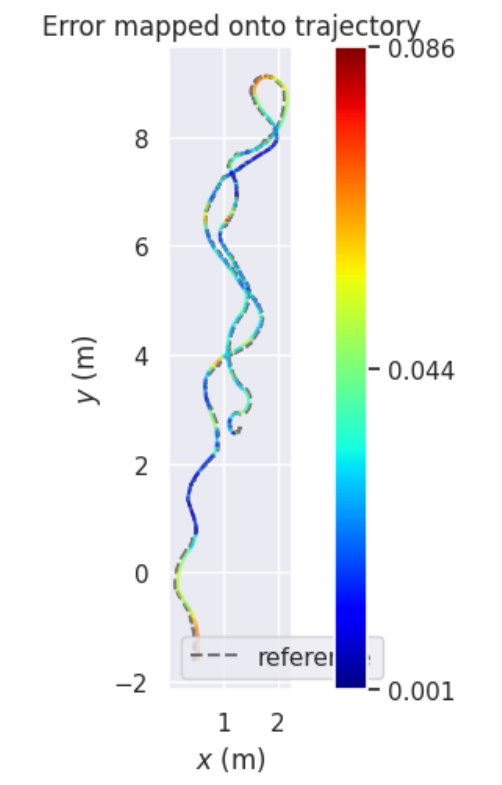}
  }
  \subfloat[B-S2\label{fig:B-S2}]
  {
    \includegraphics[width=0.22\linewidth]{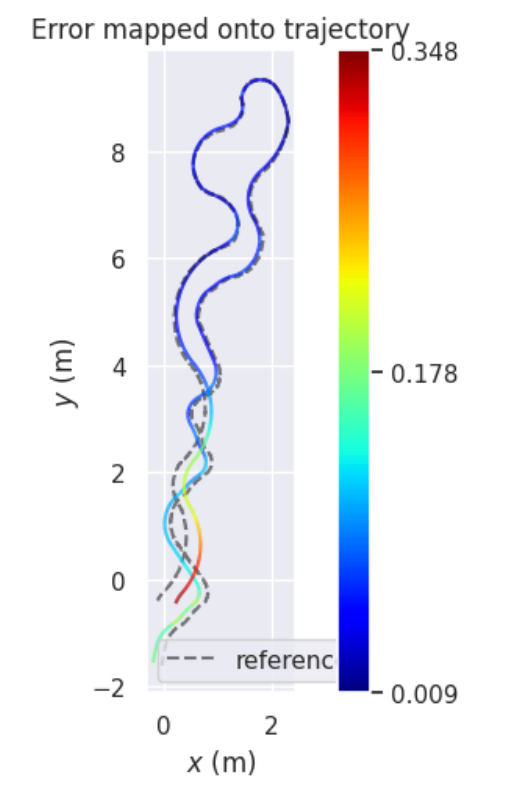}
  }
  \subfloat[B-J1\label{fig:B-J1}]
  {
    \includegraphics[width=0.22\linewidth]{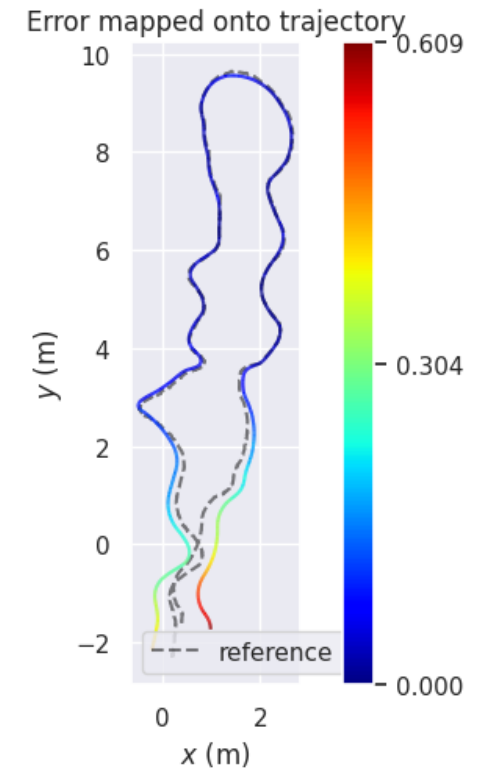}
}

  \subfloat[B-J2\label{fig:B-J2}]{
    \includegraphics[width=0.22\linewidth]{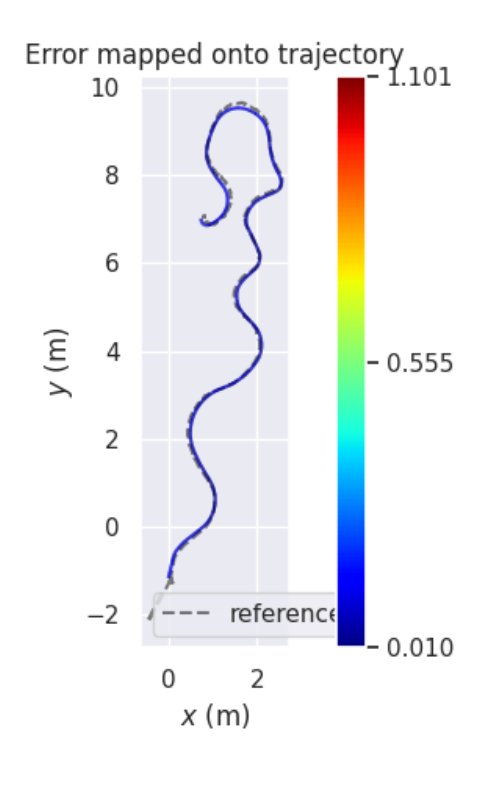}
  }
  \subfloat[B-h1\label{fig:B-H1}]
  {
    \includegraphics[width=0.22\linewidth]{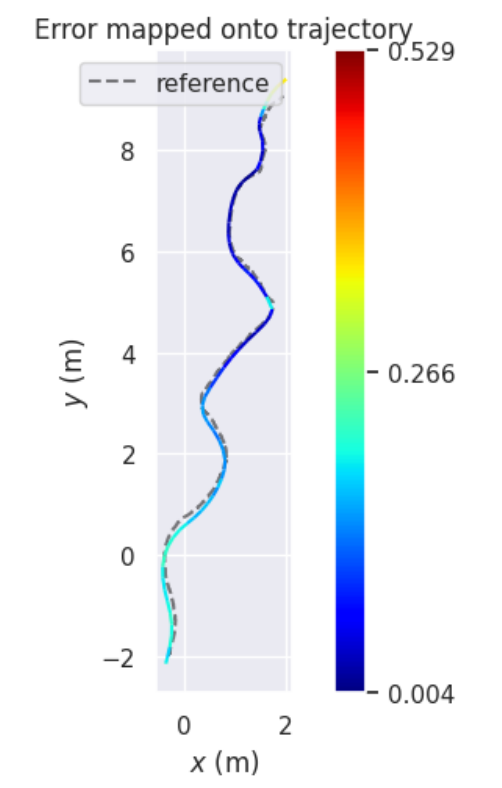}
  }
  \subfloat[B-h2\label{fig:B-H2}]
  {
    \includegraphics[width=0.18\linewidth]{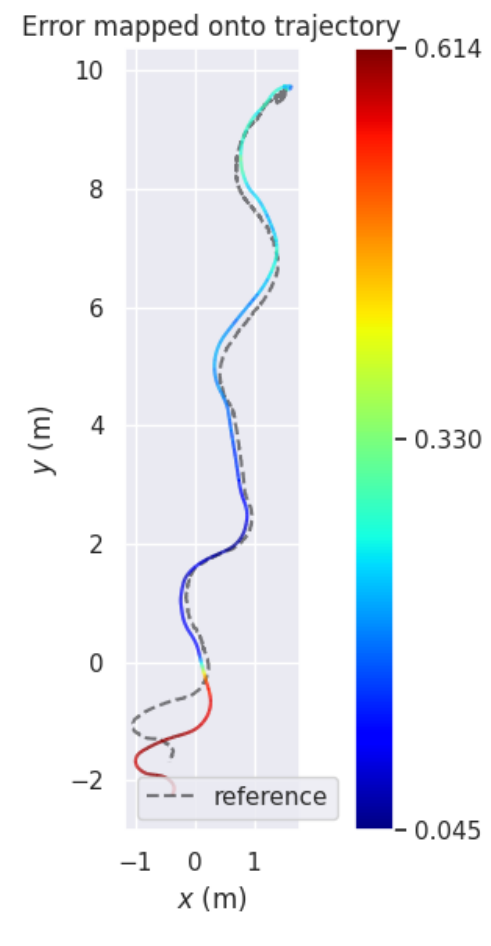}
  }
  \caption{We align the estimated trajectory for all sequences in the exhibition hall dataset with the ground-truth trajectory at the same frequency. {The ground-truth trajectory is shown as a black dashed line, and the estimated trajectory is colored by absolute translation error when compared against the ground truth}. 
  It can be observed that our CT-UIO achieves high accuracy in trajectory estimation, closely matching the ground-truth trajectories across all sequences, even in those involving fast turns.}
  \label{fig:B}
\end{figure*}

\subsection{{Evaluation  on the Exhibition Hall Dataset}}
We also evaluate CT-UIO and competing state-of-the-art methods on the exhibition hall dataset. 
In this case, two UWB anchors are placed at positions $(4.52,10.68)m$, $({-}1.71,{-}2.29)m$ to create the non-fully observable condition within an area of $20.0m{\times}11.6 m$, and there is no occlusion between UWB anchors. 
To evaluate the effectiveness of the proposed method under aggressive turns, we provide the corresponding range of rotational velocity readings, which indicate instances of aggressive motion. A total of six sequences are detailed in Table~\ref{table2}.

\begin{table}[!htb]
\centering
\caption{Description of the sequences of the exhibition hall dataset.}
\label{table2}
\setlength{\tabcolsep}{0.5mm}
\begin{tabular}{ccccc}
\toprule
\makecell{\textbf{Seq}} & 
\makecell{\textbf{Duration} \\ {(second)}} & 
\makecell{\textbf{Length} \\ {(m)}} & 
\makecell{\textbf{Rotational velocity} \\ {max/min (rad/s)}} & 
\makecell{\textbf{Description}} \\
\midrule
B-S1 & 224.2 & 30.37 & 0.20/0.14 & Slow sequences \\
B-S2 & 185.3 & 30.05 & 0.22/0.16 & Slow sequences \\
B-J1 & 135.7 & 31.09 & 1.76/0.22 & Fast sequences \\
B-J2 & 98.3 & 22.99 & 1.72/0.20 & Fast sequences \\
B-H1 & 84.1 & 18.02 & 1.70/0.15 & Hybrid sequences \\
B-H2 & 102.0 & 19.41 & 1.75/0.16 & Hybrid sequences \\
\bottomrule
\end{tabular}
\end{table}
\begin{itemize}
\item Slow sequences: the robot moves gently with low rotational velocity and gentle speed variations, following a wave path.
\item Fast sequences: the robot moves at a large rotational velocity with intense speed variations, following a wave path.  
\item Hybrid sequences: The robot follows a wave trajectory that combines both slow and fast rotational motion, incorporating a mix of low and high-speed phases.
\end{itemize}

\begin{table}[!htb]
\centering
\caption{Performance comparison in APE (RMSE, Meter) of different methods on the exhibition hall dataset. The best results are marked in bold.}
\label{table3}

\setlength{\tabcolsep}{1.7mm}
\begin{tabular}{ccccccc}
\toprule
\makecell{\textbf{Seq}} & 
\makecell{B-S1 } & 
 B-S2& 
B-J1& 
B-J2&
B-H1 &
B-H2\\
\midrule
\makecell[c]{{Optimization+}\\{Filtering}\cite{yang2023novel}}& 0.432 & 0.372& 0.760 & 0.391 & 0.885 & 0.728 \\
\makecell[c]{{Optimization+}\\{Trust Region}\cite{zhou2024optimization}} & 0.393 & 0.405 & 0.588 & 0.562 & 0.543 & 0.413 \\
 GPDA\cite{WOS:001173317800046}& 0.319   & 0.372   & 0.523  & 0.460 &0.537 & 0.409\\
VMHE\cite{9829196} & 0.255  & 0.347  & 0.433   & 0.380 &0.501 &0.373\\
{{SFUISE}\cite{li2023continuous}} & 0.042 & 0.127 & 0.243 & 0.168 & 0.203 & 0.358 \\
Ours & \textbf{0.040} & \textbf{0.122} & \textbf{0.219} & \textbf{0.098} & \textbf{0.150} & \textbf{0.311}\\
\bottomrule
\end{tabular}
\end{table}

\begin{figure*}[!htb]
  \centering
  \subfloat[B-S1\label{fig:cdf:B-S1}]{
    \includegraphics[width=6cm]{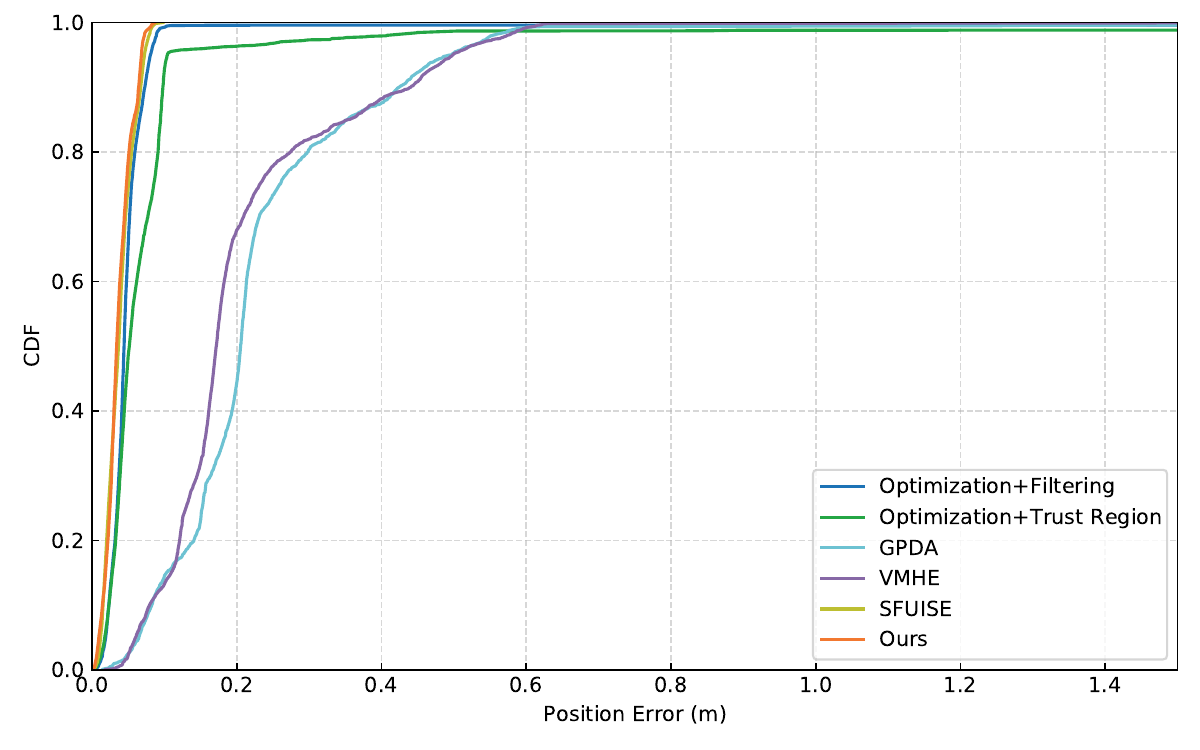}
  }\hspace{-5mm}
  \subfloat[B-S2\label{fig:cdf:B-S2}]
  {
    \includegraphics[width=6cm]{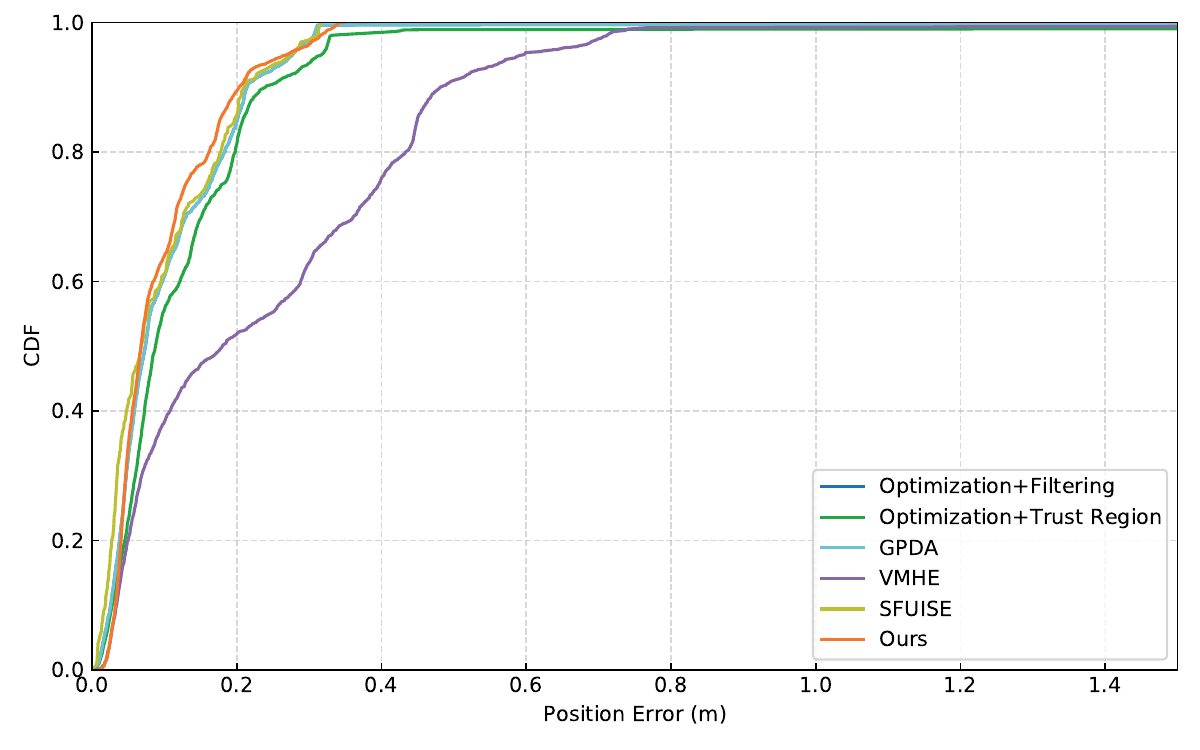}
  }\hspace{-5mm}
  \subfloat[B-J1\label{fig:cdf:B-J1}]
  {
    \includegraphics[width=6cm]{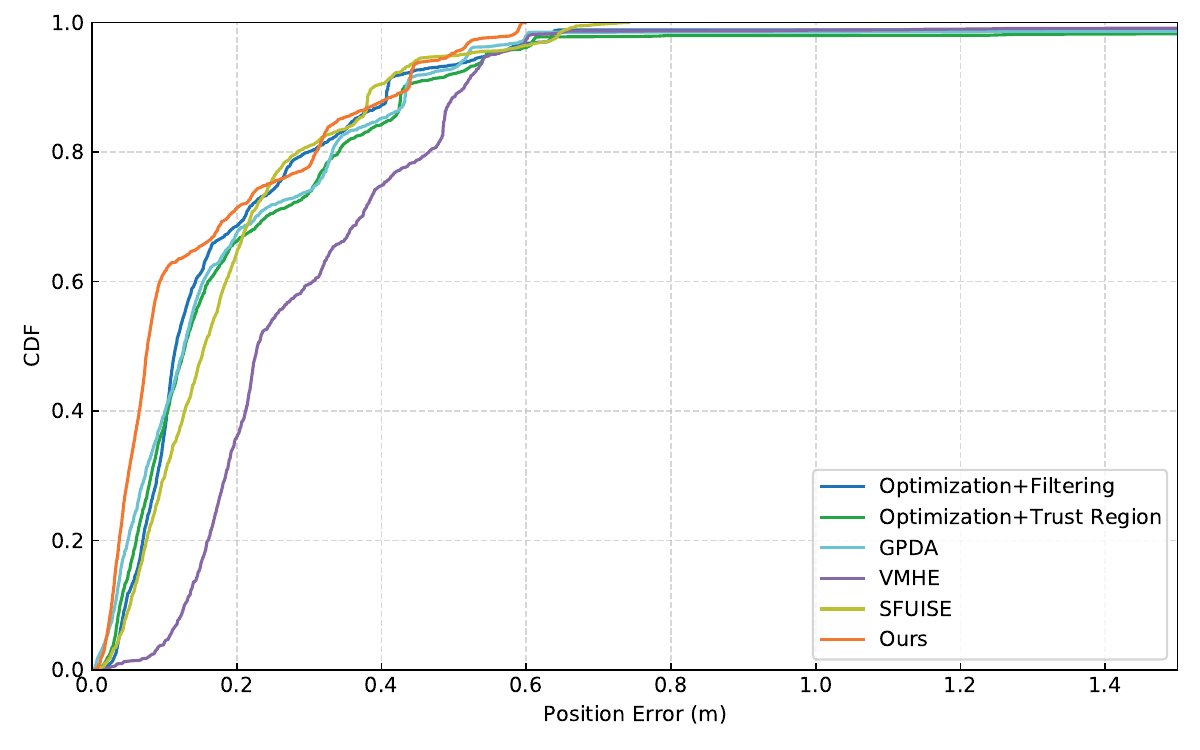}
}
  \vspace{-4mm}
  \subfloat[B-J2\label{fig:cdf:B-J2}]{
    \includegraphics[width=6cm]{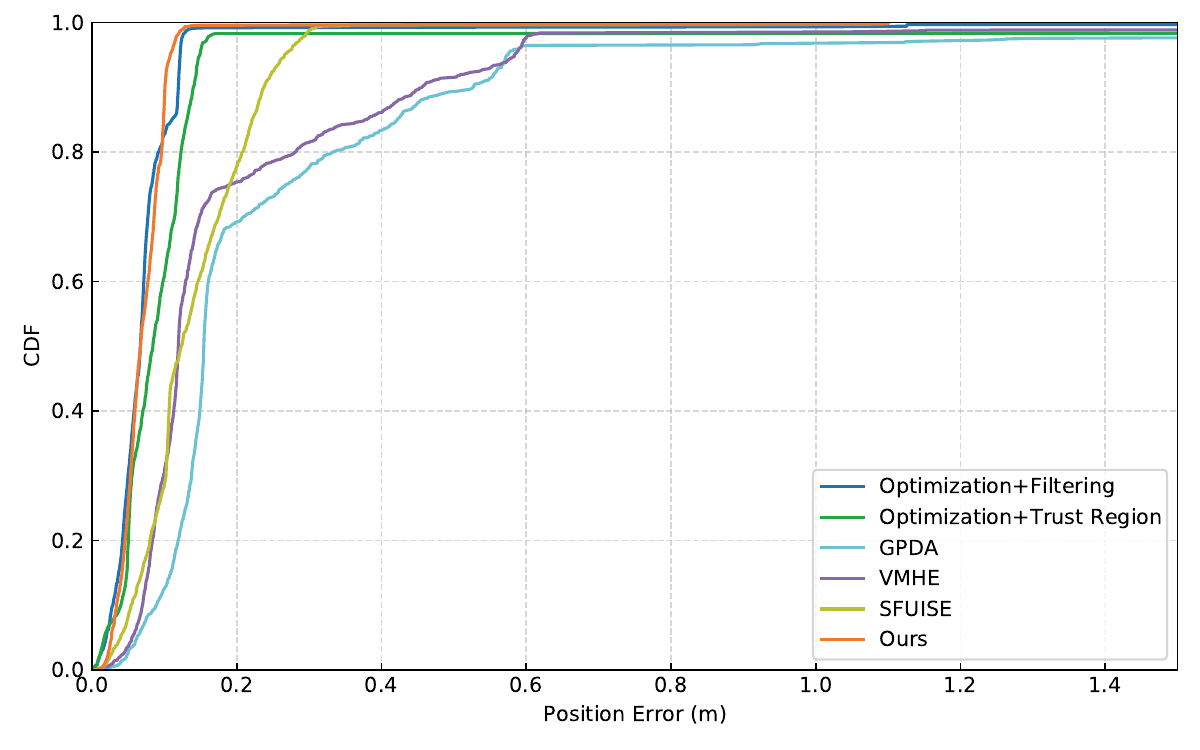}
  }\hspace{-5mm}
  \subfloat[B-H1\label{fig:cdf:B-H1}]
  {
    \includegraphics[width=6cm]{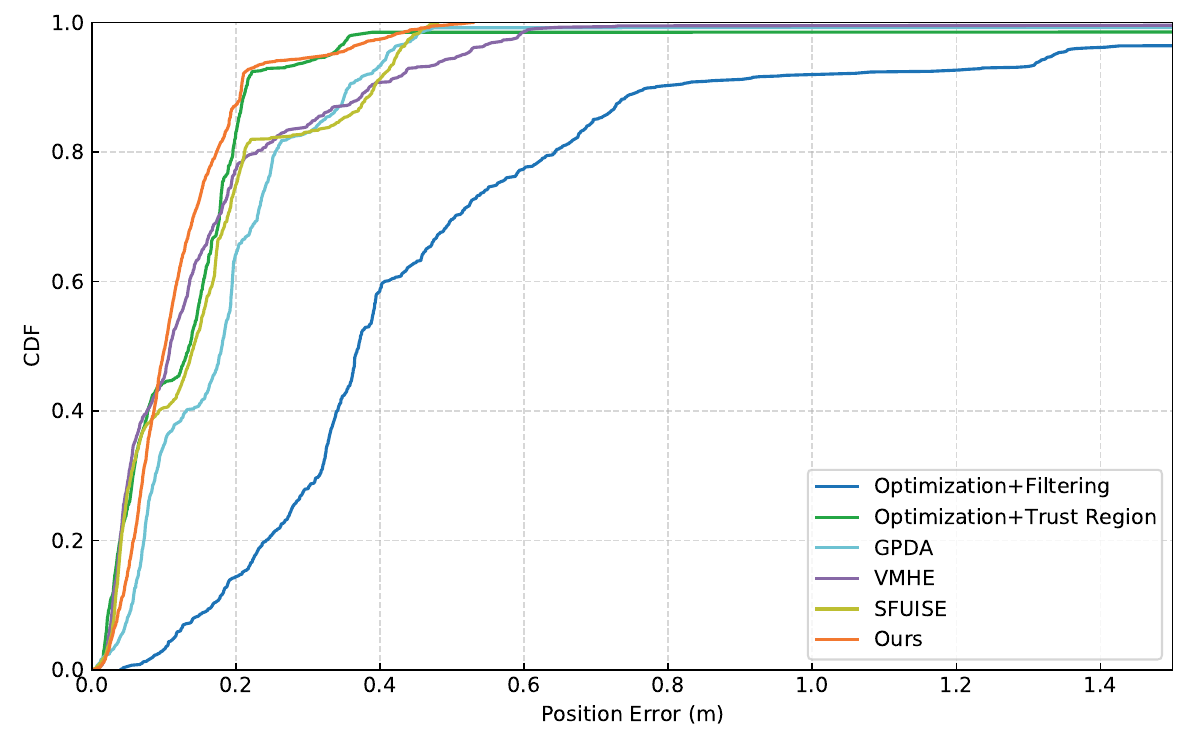}
  }\hspace{-5mm}
  \subfloat[B-H2\label{fig:cdf:B-H2}]
  {
    \includegraphics[width=6cm]{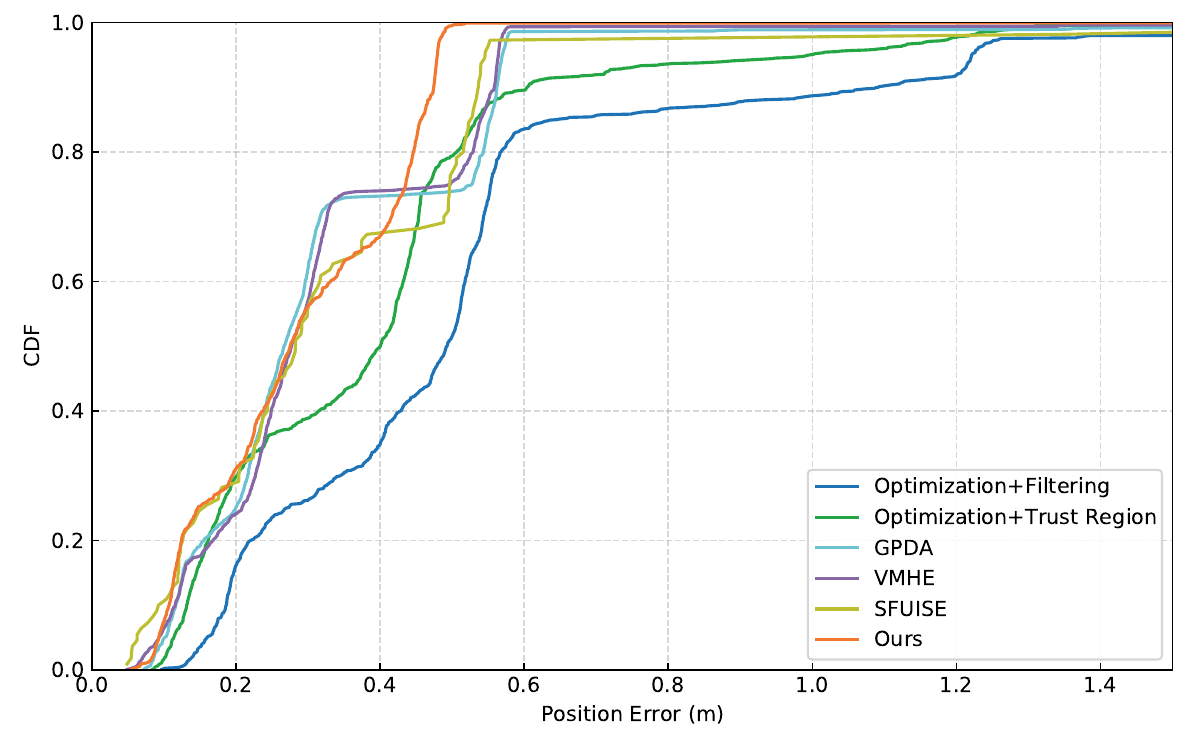}
  }
  \caption{{CDFs of APE for different methods on the exhibition hall dataset. The results of SFUISE with uniform B-splines are closer to those of our method, but its performance holds only for slow sequences, and cannot maintain consistent in fast and hybrid sequences. 
  In contrast, our method shows a more concentrated APE distribution across all sequences compared to the other methods.}}
  \label{fig:B:CDF}
\end{figure*}
\begin{figure}[!htb]
\centering
\includegraphics[width=\linewidth,trim=0.8cm 0.6cm 2.5cm 1.8cm,clip]{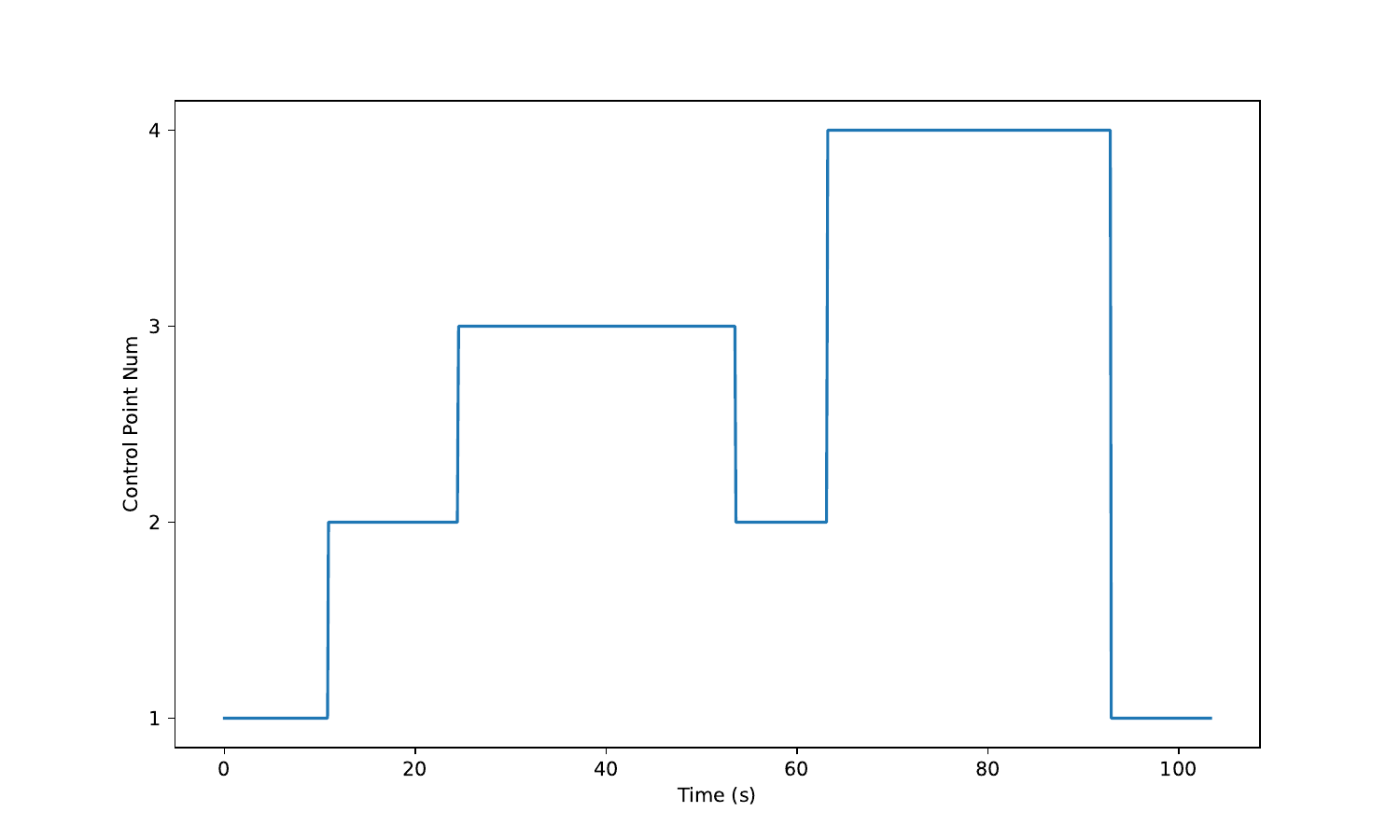}
\caption{The distribution of control points in sequence B-H2.
}
\label{B-h2-cp}
\end{figure}
Fig.~\ref{fig:B} shows the overview of trajectories in the exhibition hall dataset. 
The APE results of the compared methods on the exhibition hall dataset are summarized in Table~\ref{table3}. 
The results notably indicate that CT-UIO achieves an overall improvement in localization performance, even in sequences involving fast turning, thereby verifying the effectiveness of its dynamic knot span adjustment strategy. 

{We also obtain the CDF plots for each sequence, as shown in Fig.~\ref{fig:B:CDF}. It can be seen that the CDF curve of our method is located at the top of the remaining five methods. For sequence B-H2, the proposed method achieves a $90\%$ error of $0.474m$, outperforming Optimization+Filtering ($1.092m$), Optimization+Trust Region ($0.604m$),
GPDA ($0.562m$), VMHE ($0.558m$), and SFUISE ($0.536m$).} Fig.~\ref{B-h2-cp} shows the distribution of control points in sequence B-H2.  

{Furthermore, it is noteworthy that the discrete-time methods (Optimization+Filtering~\cite{yang2023novel}, Optimization+Trust Region~\cite{zhou2024optimization}, GPDA\cite{WOS:001173317800046}, and VMHE\cite{9829196} ) lead to an increase of APE or even exhibit obvious location deviations when the robot undergoes rapid rotational motion. }
Compared with the uniform B-splines method (SFUISE), CT-UIO ensures a better capture of motion characteristics by adjusting the control point density. Specifically, CT-UIO  reduces the density of control points in sections with less speed variation and increases it during significant speed changes. SFUISE cannot adapt to trajectory complexity, as it sets the knot span as a constant value throughout the trajectory.  Besides, in the case of only two UWB anchors, UWB positioning suffers a notable accuracy degradation. 
To address the observability issue in a few-anchor system, CT-UIO introduces VAs by coupling UWB-ranging data and motion priors from the IMU/odometer fusion model to provide more comprehensive constraints. Based on this, the proposed CT-UIO exhibits robust performance through multiple VAs' ranging constraints even with limited anchor availability.
\begin{figure*}[!htb]
  \centering
  \subfloat[C-S1\label{fig:C-S1}]{
    \includegraphics[width=0.32\linewidth,trim=1.5cm 0.4cm 1cm 0.35cm,clip]{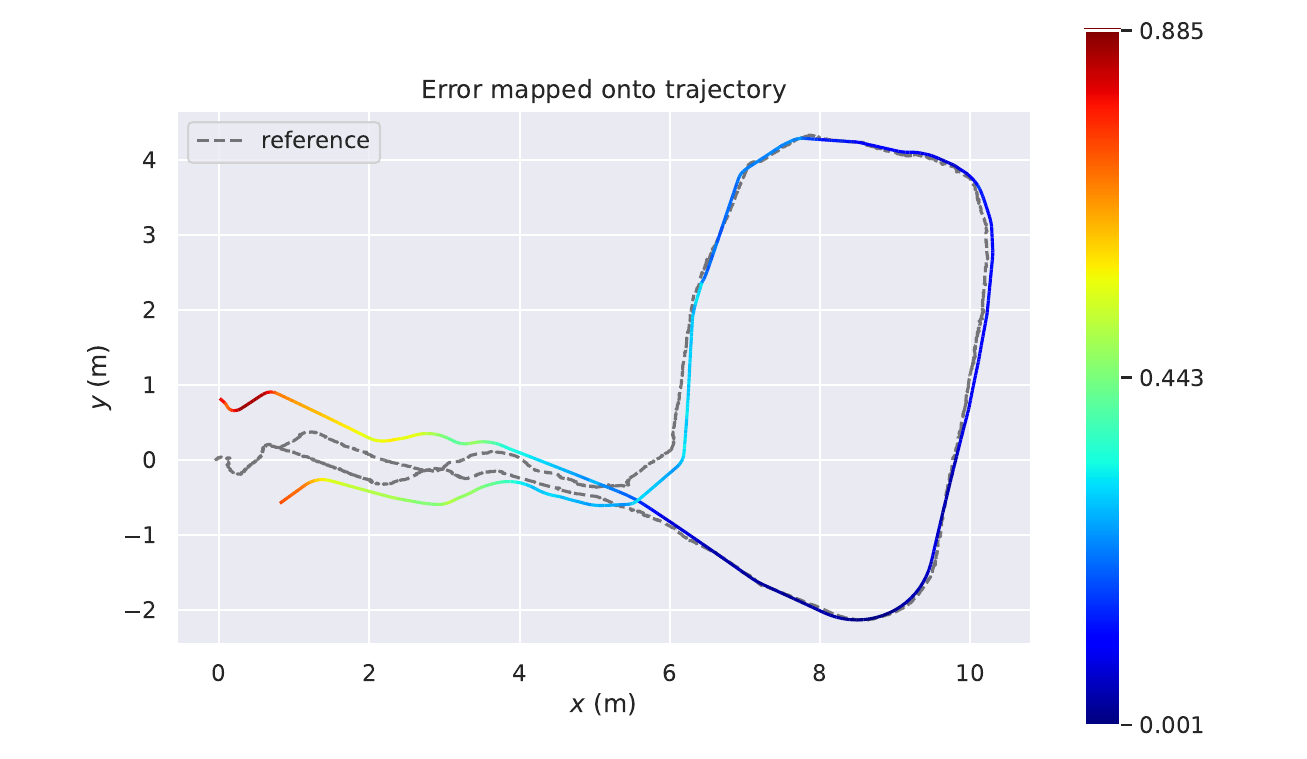}
  }\hspace{-5mm}
  \subfloat[C-S2\label{fig:C-S2}]
  {
    \includegraphics[width=0.32\linewidth,trim=1.5cm 0.4cm 1.6cm 0.4cm,clip]{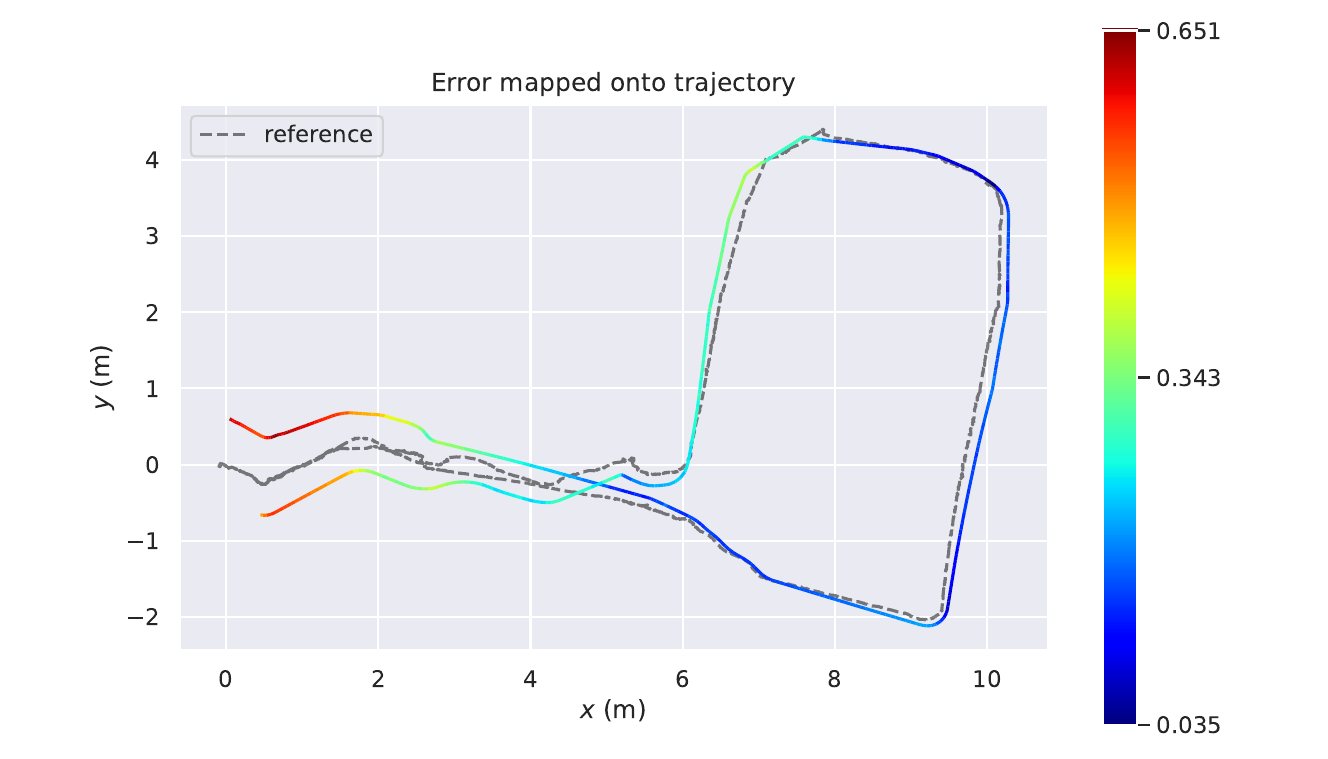}
  }\hspace{-5mm}
  \subfloat[C-J1\label{fig:C-J1}]
  {
    \includegraphics[width=0.32\linewidth,trim=1.5cm 0.4cm 1.6cm 0.4cm,clip]{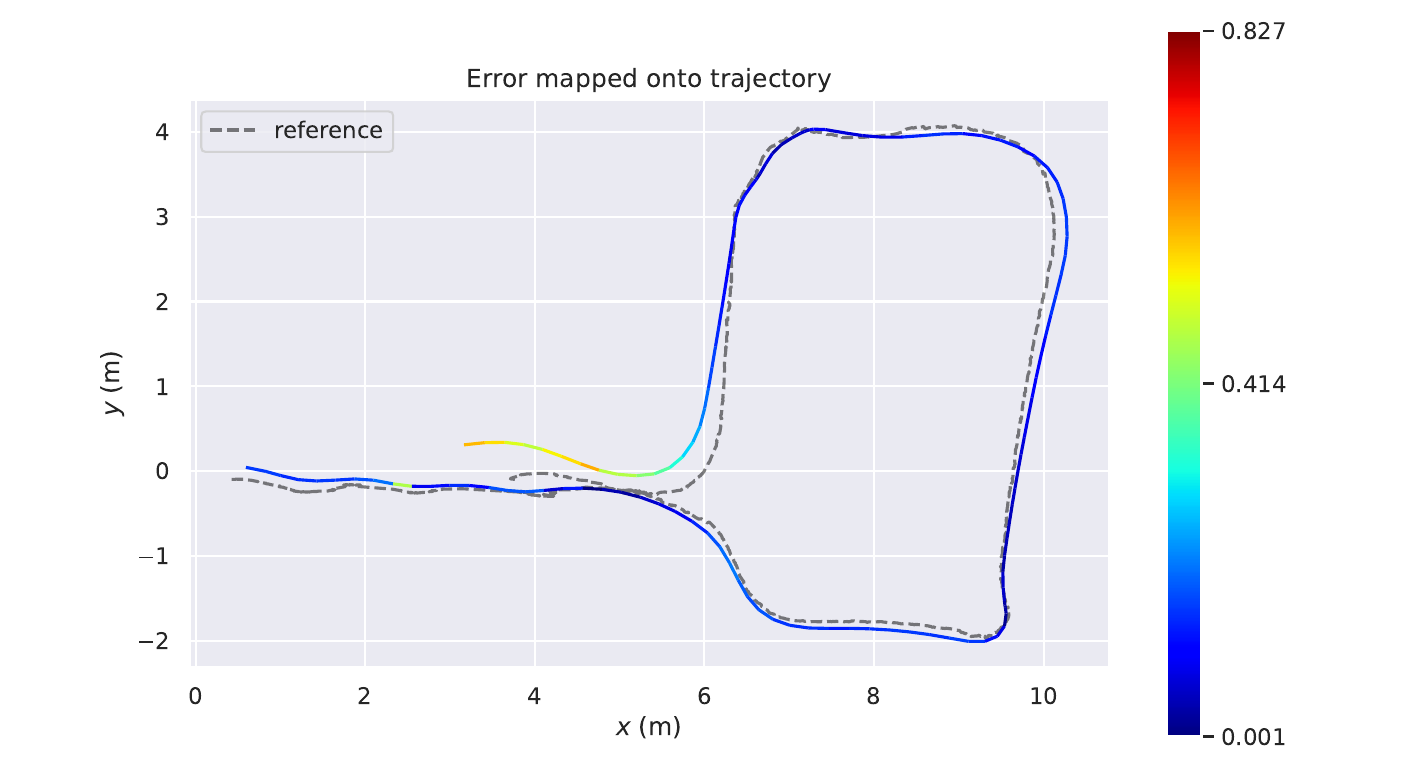}
}
  \vspace{-4mm}
  \subfloat[C-J2\label{fig:mad_c}]{
    \includegraphics[width=0.34\linewidth,trim=1.6cm 0.2cm 1.6cm 0.4cm,clip]{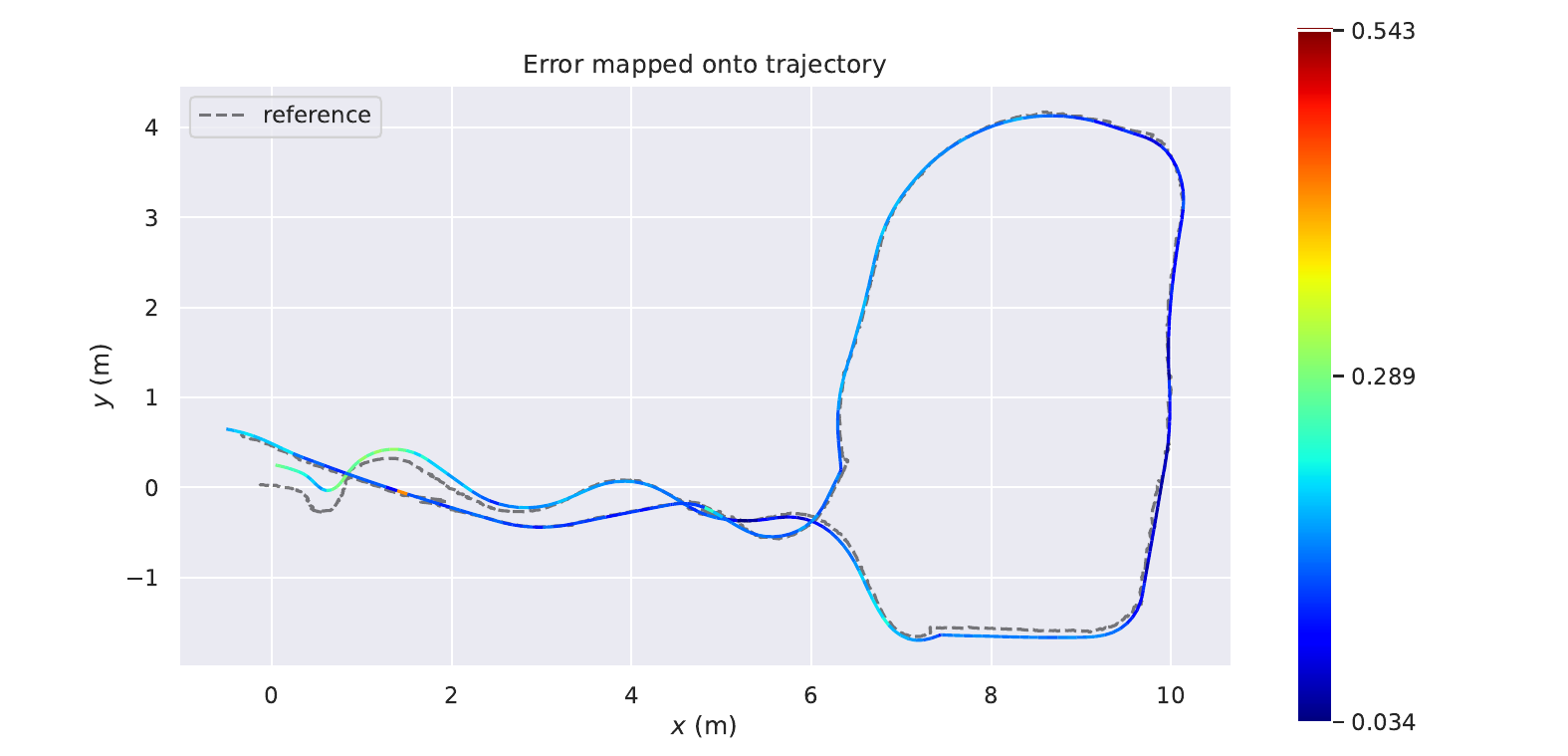}
  }\hspace{-6mm}
  \subfloat[C-H1\label{fig:irmad_c_j2}]
  {
    \includegraphics[width=0.30\linewidth,trim=1.5cm 0.2cm 1.6cm 0.4cm,clip]{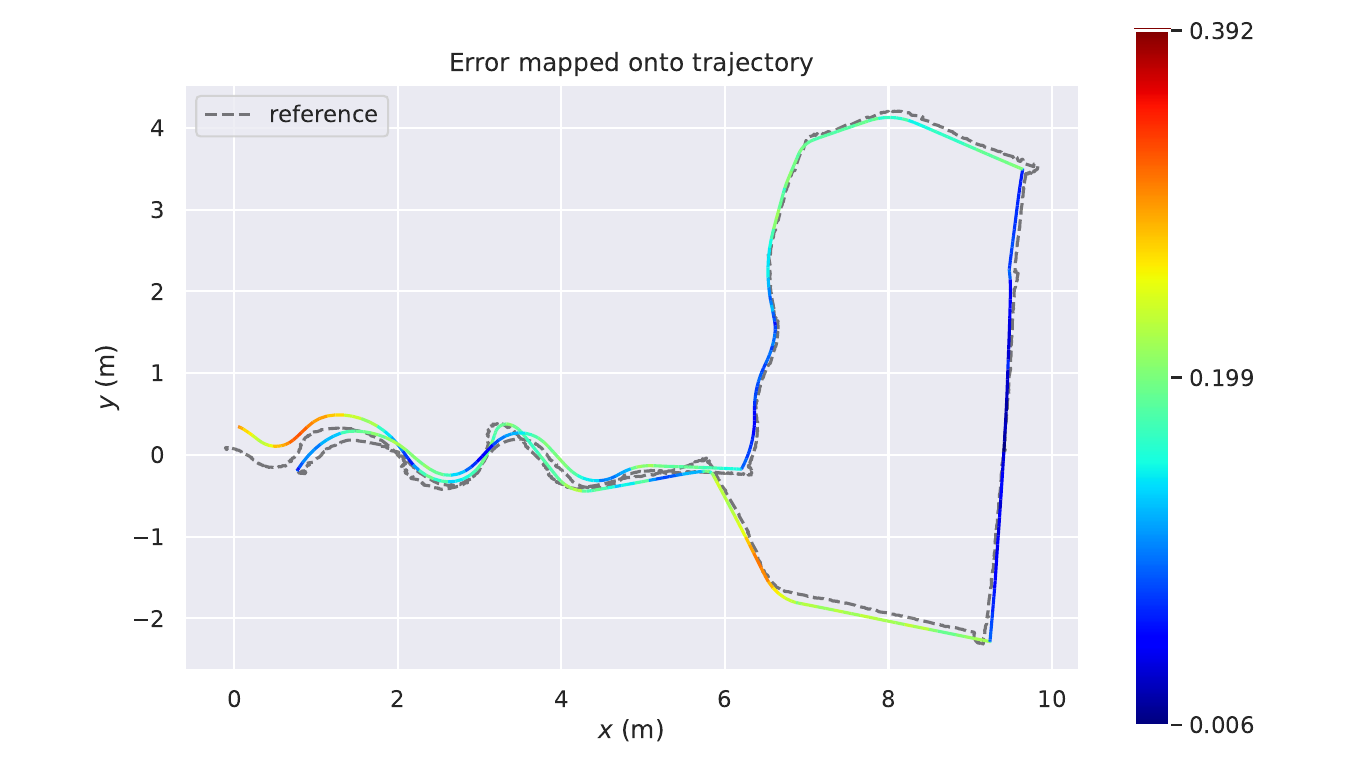}
  }\hspace{-4mm}
  \subfloat[C-H2\label{fig:isfa_c_h2}]
  {
    \includegraphics[width=0.33\linewidth,trim=1.cm 0.4cm 1.6cm 0.4cm,clip]{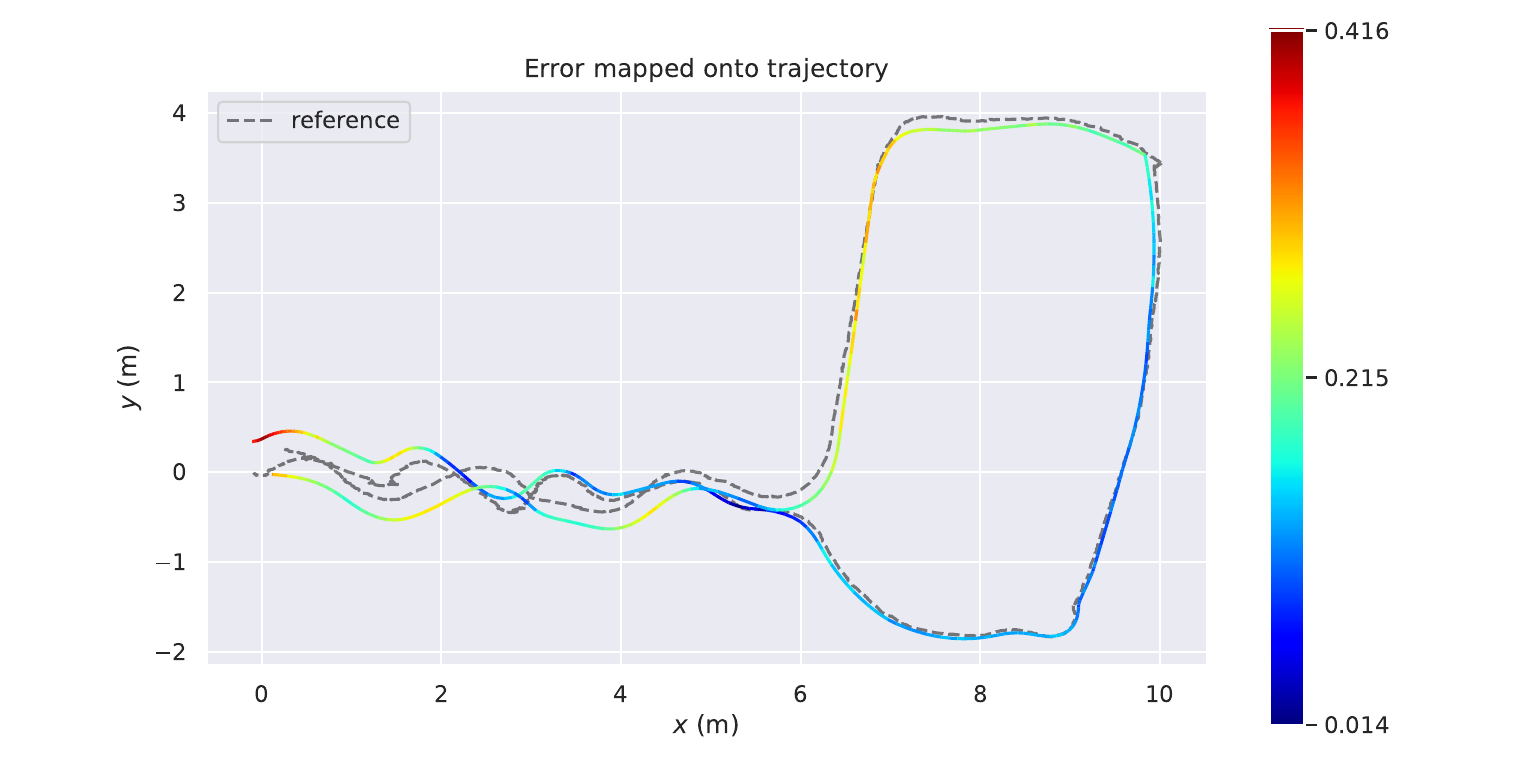}
  }
  \caption{{We query the estimated trajectory from the obtained continuous-time trajectory and evaluate it against the ground-truth trajectory. The absolute translation error is mapped onto the estimated trajectory for each sequence in the corridor dataset. 
  Our proposed CT-UIO not only achieves high accuracy in trajectory estimation but also effectively fits translational velocity variations.}}
  \label{fig:C}
\end{figure*}

\begin{figure}[!htb]
\centering
\includegraphics[width=\linewidth,trim=0.5cm 0.6cm 2.5cm 1.8cm,clip]{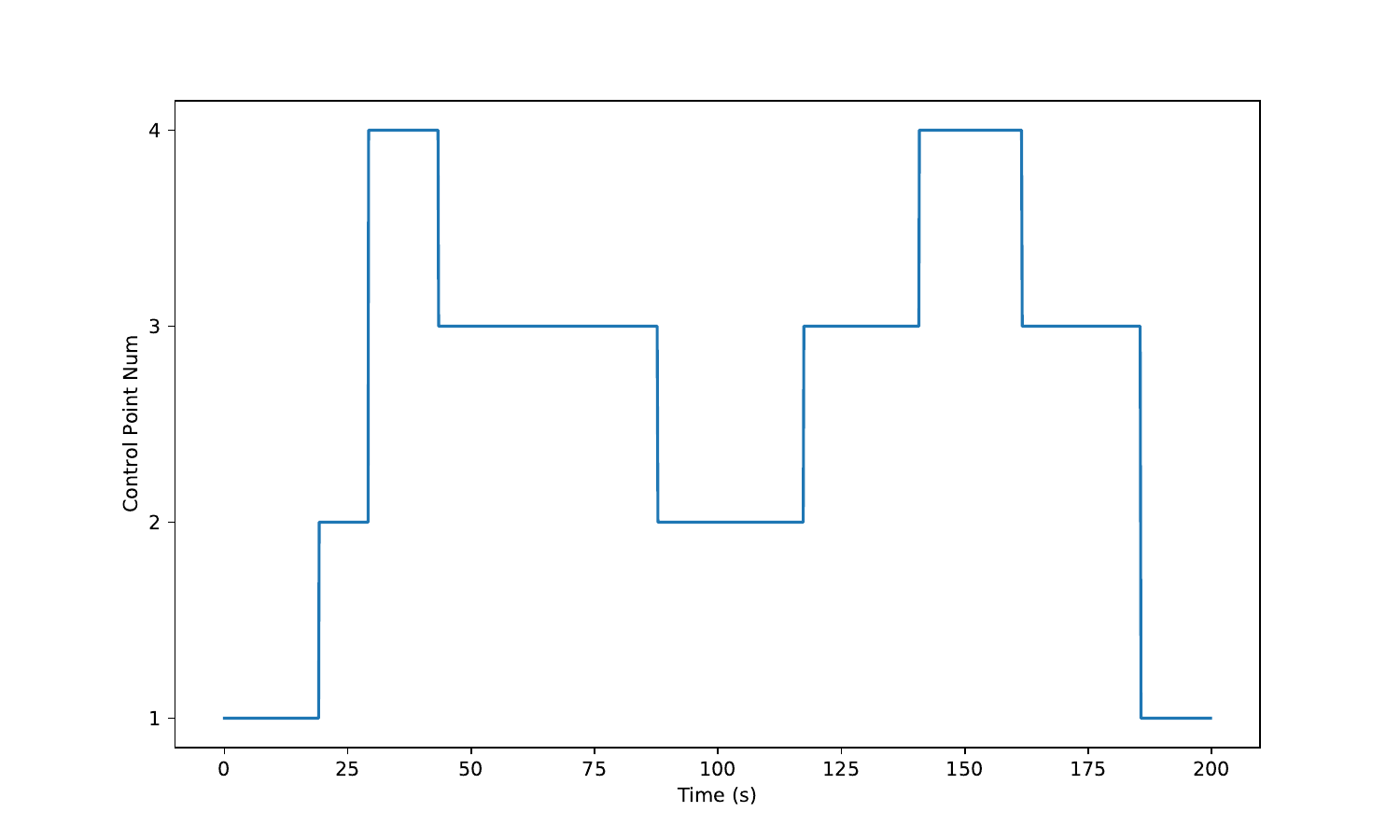}
\caption{The distribution of control points in sequence C-H1.
}
\label{C-h2-cp}
\end{figure}
\begin{figure*}[!htb]
  \centering
  \subfloat[C-S1\label{fig:cdf:C-S1}]{
    \includegraphics[width=6cm]{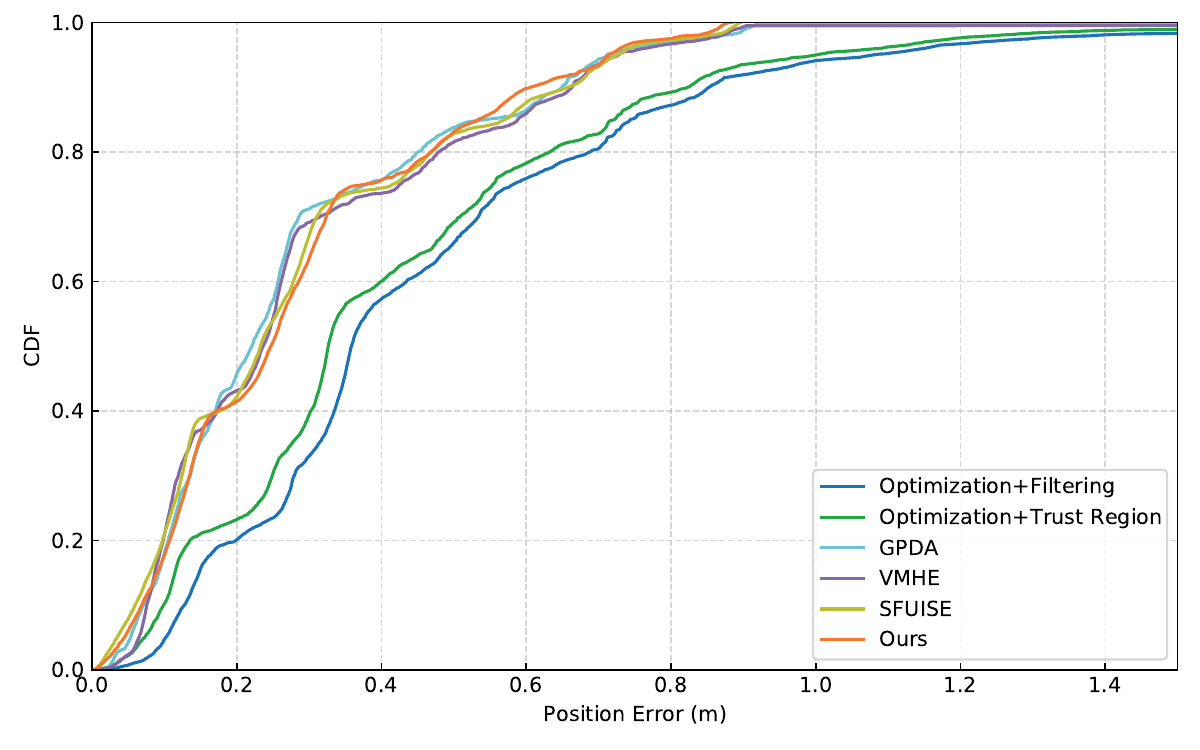}
  }\hspace{-5mm}
  \subfloat[C-S2\label{fig:cdf:C-S2}]
  {
    \includegraphics[width=6cm]{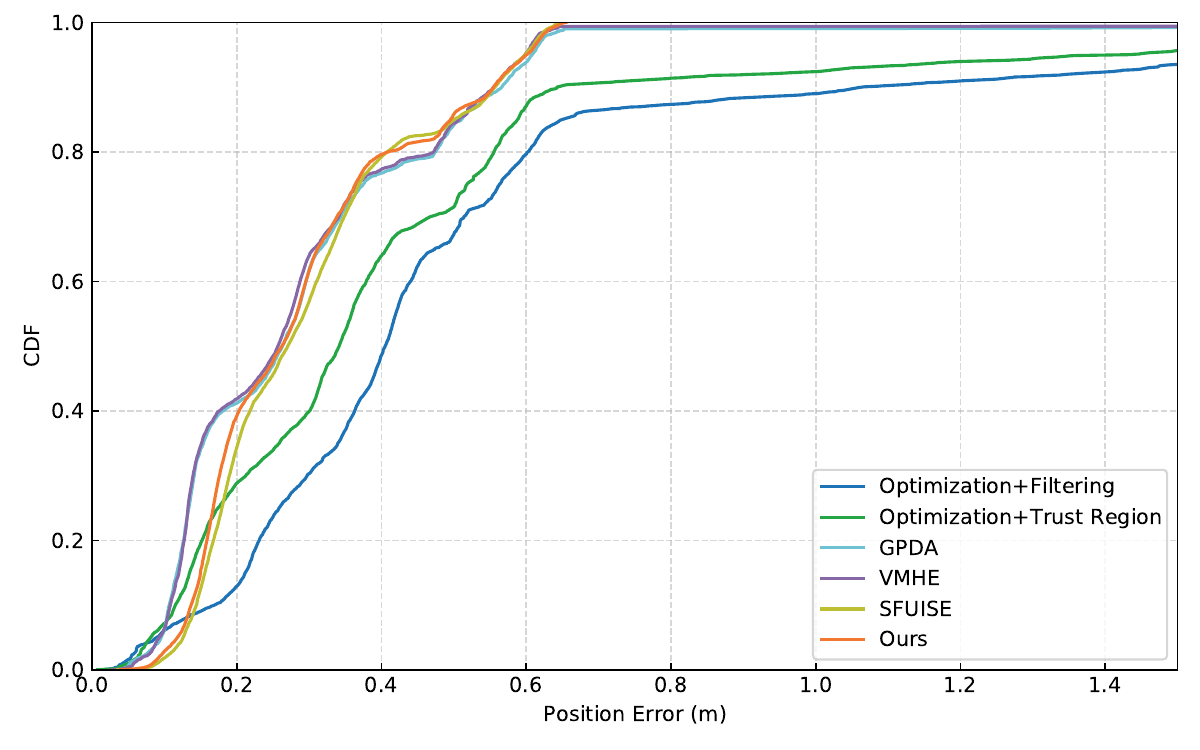}
  }\hspace{-5mm}
  \subfloat[C-J1\label{fig:cdf:C-J1}]
  {
    \includegraphics[width=6cm]{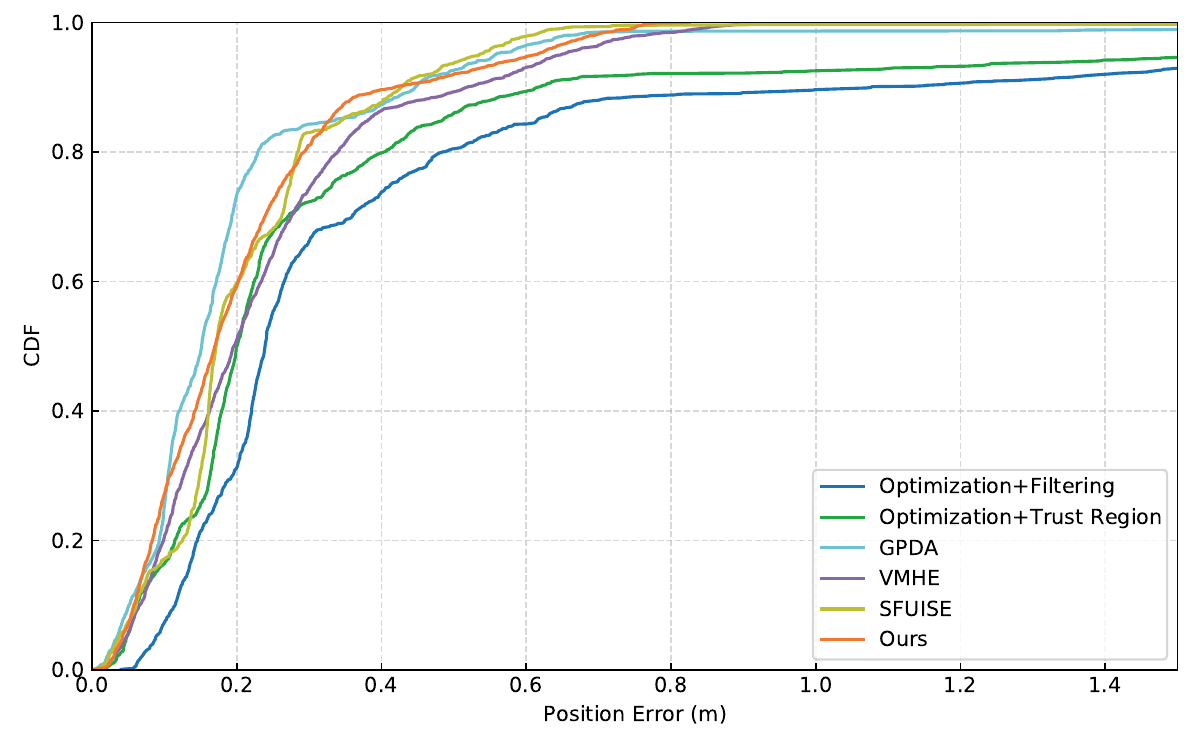}
}
  \vspace{-4mm}
  \subfloat[C-J2\label{fig:cdf:C-J2}]{
    \includegraphics[width=6cm]{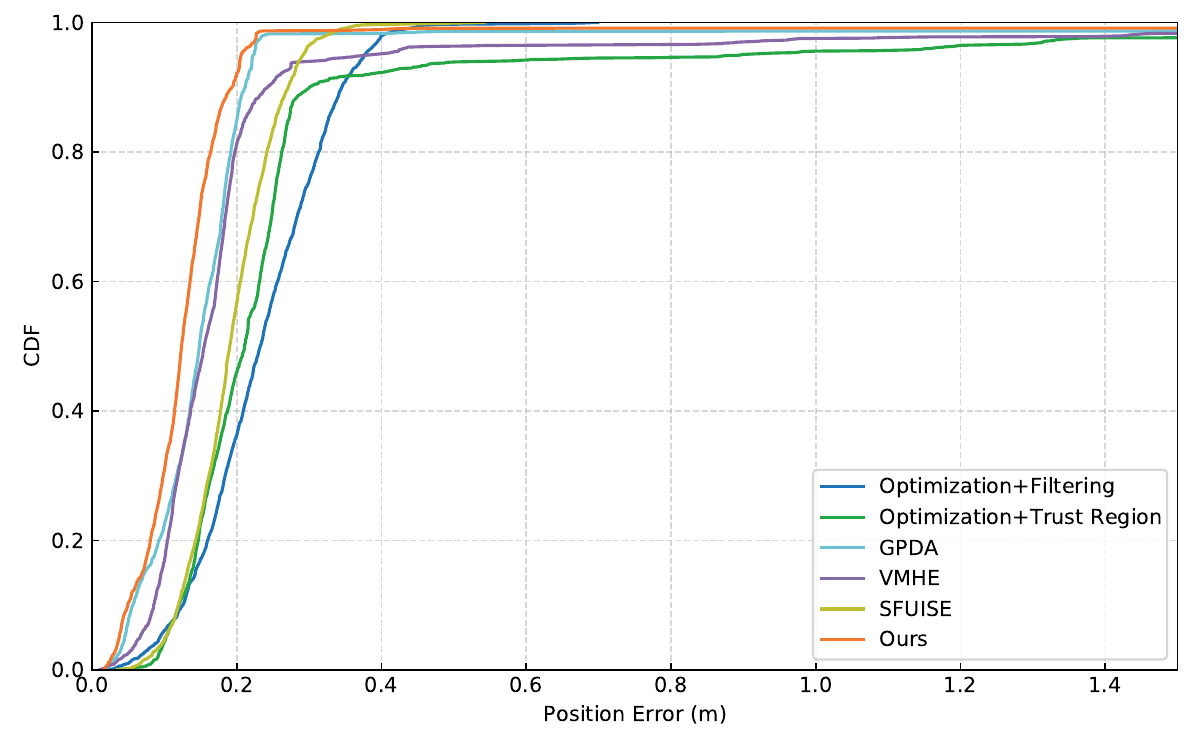}
  }\hspace{-5mm}
  \subfloat[C-H1\label{fig:cdf:C-H1}]
  {
    \includegraphics[width=6cm]{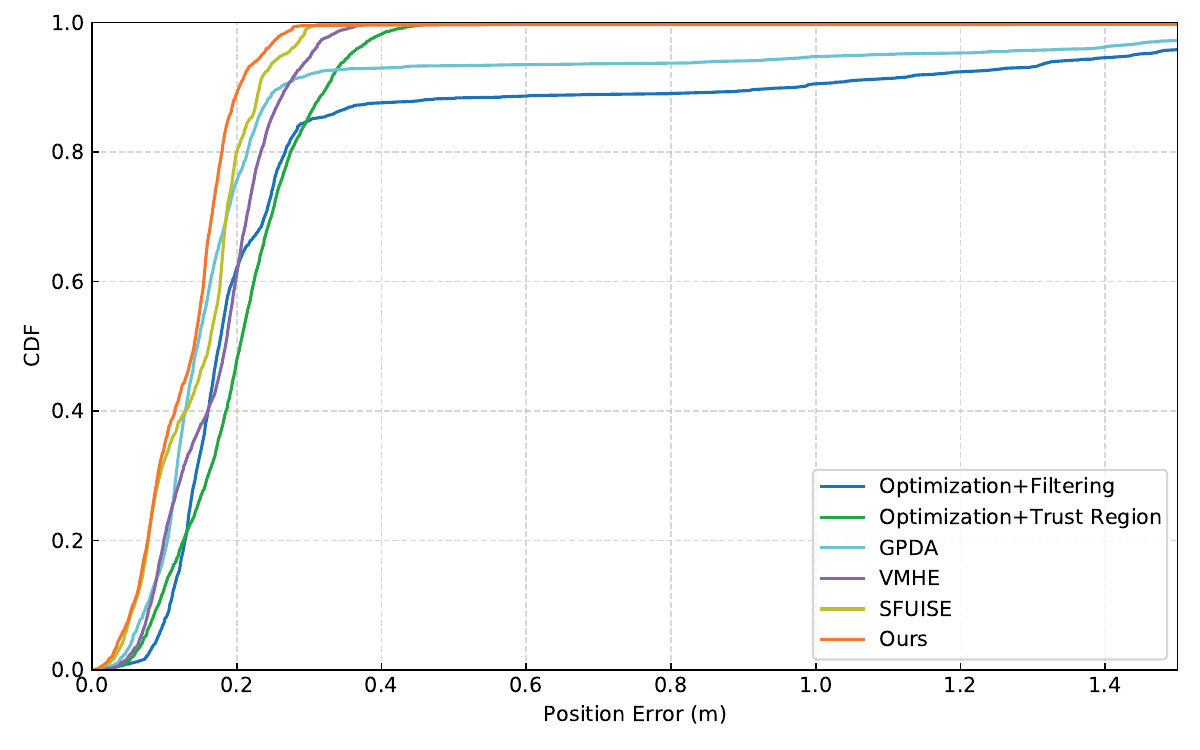}
  }\hspace{-5mm}
  \subfloat[C-H2\label{fig:cdf:C-H2}]
  {
    \includegraphics[width=6cm]{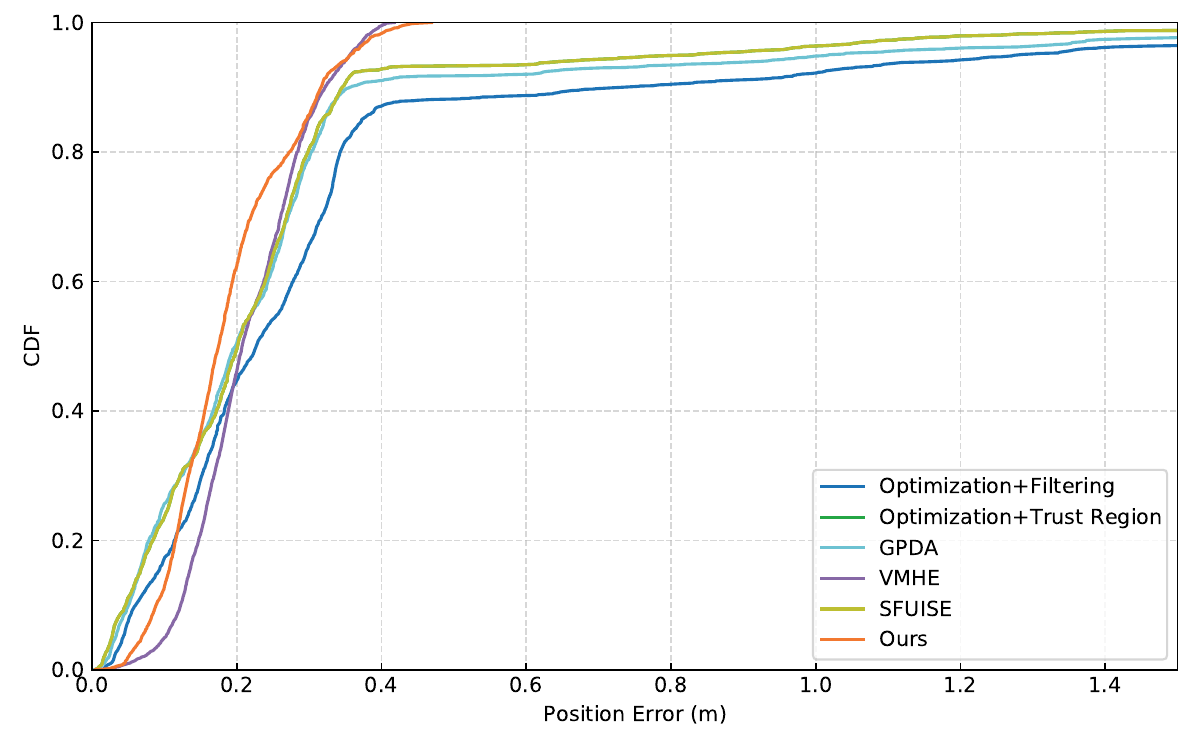}
  }
  \caption{{CDFs of APE for different methods on the office dataset. The proposed method exhibits the closest performance to the SFUISE in slow sequences while outperforming all compared methods in fast and hybrid sequences, indicating superior accuracy and stability.}}
  \label{fig:C:CDF}
\end{figure*}
\subsection{{Evaluation on the Office Dataset}}
   {Lastly, the proposed CT-UIO is evaluated in the office environment of meeting rooms and corridors, using fewer UWB anchors. The robot starts in the corridor, passes through the meeting room, and then returns to the corridor. While two UWB anchors are available at positions $(10.20,3.94)m$, $(6.23,{-}2.33)m$ in the meeting room $(6.9m{\times}10.5m)$, only a single UWB anchor is available at the position $(-0.51, 0.18)m$ in the corridor $(2.0m{\times}9.8m)$. 
   To evaluate the effectiveness of the proposed method under rapid translational speed changes and sharp turns, we present the corresponding range of translational and rotational velocities, which indicate instances of aggressive motion. A total of six sequences are detailed in Table~\ref{table4}.}

\begin{table}[!htb]
\centering
\caption{{Description of the sequences of the office dataset.}}
\label{table4}
\fontsize{8}{11}\selectfont
\setlength{\tabcolsep}{0.5mm}
\begin{tabular}{cccccc}
\toprule
\makecell{\textbf{Seq}} & 
\makecell{\textbf{Duration} \\ {(second)}} & 
\makecell{\textbf{Length} \\ {(m)}} & 
\makecell{\textbf{Translational}\\\textbf{velocity} \\ {max/min (m/s)}}&
\makecell{\textbf{Rotational}\\\textbf{ velocity} \\ {max/min (rad/s)}} & 
\makecell{\textbf{Description}} \\
\midrule
C-S1 & 526.4 & 51.01 & 0.15/0.10 &0.20/0.12 & Slow sequences \\
C-S2 & 474.2 & 53.24 & 0.17/0.12 &0.20/0.15 & Slow sequences \\
C-J1 & 138.6 & 37.42 & 0.26/0.22 & 1.74/0.20 &Fast sequences \\
C-J2 & 159.8 & 37.30 & 0.26/0.22 & 1.76/0.22 &Fast sequences \\
C-H1 & 242.5 & 39.95 & 0.26/0.13 & 1.74/0.12 & Hybrid sequences \\
C-H2 & 198.6 & 36.36 & 0.26/0.12 & 1.75/0.14 & Hybrid sequences \\
\bottomrule
\end{tabular}
\end{table}
{\begin{itemize}
\item Slow sequences: the robot moves smoothly with low translational and rotational velocities, following straight and rectangular trajectories.
\item Fast sequences: the robot exhibits high translational and rotational velocities with rapid speed variations, following both a rectangular path and a wave path. 
\item Hybrid sequences: the robot follows a combination of wave and rectangular trajectories, featuring alternating phases of slow and fast translational and rotational motion.
\end{itemize}
}

{Fig.~\ref{fig:C} shows the overview of trajectories in the office dataset.  Fig.~\ref{C-h2-cp} shows the distribution of control points in sequence C-H1. The APE results of the compared methods on the office dataset are summarized in Table~\ref{table5}.} 

{According to the APE results, compared with the existing state-of-the-art discrete-time and continuous-time methods, our CT-UIO outperforms the existing methods and achieves the best results in most sequences. Furthermore, the estimated trajectories of our method are in good agreement with the ground truth in all the sequences. As for SFUISE, its localization results on C-S1 and C-S2 are close to CT-UIO. 
However, large errors are generated due to the fast translational and rotational motion in fast and hybrid sequences. Conversely, the proposed CT-UIO achieves consistent localization estimation. This can be attributed to the proposed adaptive knot span adjustment strategy that adjusts the density of control points based on the robot speed ranges during rapid and hybrid motion.}

{CDF plots are also an additional visual representation of the positioning error in Fig.~\ref{fig:C:CDF}. 
All the CDFs of our method are higher than those of the five state-of-the-art methods. For sequence C-H1, $90\%$ of the estimates from our method achieve APE less than $0.203m$, better than Optimization+Filtering ($0.974m$), Optimization+Trust Region ($0.324m$), GPDA ($0.261m$), VMHE ($0.269m$), and SFUISE ($0.231m$).}

\begin{table}[!htb]
\centering
\caption{Performance comparison in APE (RMSE, Meter) of different methods on the office dataset. The best results are marked in bold.}
\label{table5}
\resizebox{0.5\textwidth}{!}{
\setlength{\tabcolsep}{1.7mm}
\begin{tabular}{ccccccc}
\toprule
\makecell{\textbf{Seq}} & 
\makecell{C-S1 } & 
 C-S2& 
C-J1& 
C-J2&
C-H1 &
C-H2\\
\midrule
\makecell[c]{{Optimization+}\\{Filtering}\cite{yang2023novel}}& 0.655 &  0.764 & 0.804  &  0.760 &  0.661 & 0.691  \\
\makecell[c]{{Optimization+}\\{Trust Region}\cite{zhou2024optimization}} & 0.587 & 0.652 & 0.737  & 0.638 & 0.537 &  0.629 \\
 GPDA\cite{WOS:001173317800046}& 0.499  & 0.515    &0.412    & 0.542 & 0.384 & 0.591 \\
VMHE\cite{9829196} & 0.447  &  0.486   &  0.329  & 0.535  & 0.272 & 0.349 \\
{{SFUISE}\cite{li2023continuous}} & 0.357 & 0.333  & 0.300   &  0.250 &   0.223& 0.261  \\
Ours & \textbf{0.352} & \textbf{0.320} & \textbf{0.263} & \textbf{ 0.202} & \textbf{0.189} & \textbf{0.208}\\
\bottomrule
\end{tabular}
}
\end{table}

{The results of these experiments in the corridor, exhibition hall, and office datasets prove that the proposed CT-UIO is highly effective in situations involving fast translational and rotational motion. 
When the UWB/IMU/odometer sensors are not time-synchronized, the discrete-time formulation (Optimization+Filtering, Optimization+Trust Region, GPDA, and VMHE) suffers from a loss in accuracy, particularly under fast motion, due to its reliance on a constant velocity assumption in the period of time between consecutive UWB ranging measurements. 
The hypothesis of time synchronization may not always hold for multiple sensors and fast motions. 
In contrast, the continuous-time formulation (SFUISE and our proposed CT-UIO) effectively handles asynchronous measurements and enables accurate trajectory estimates to be sampled at any time, thereby maintaining high positioning accuracy.} Across all sequences, the CT-UIO maintains more stable and consistent accuracy by utilizing a non-uniform continuous-time trajectory representation. This strategy effectively adjusts the control point interval and ensures better capture of motion pattern characteristics.
  
\subsection{Ablation Study}
{To further evaluate the effectiveness of each novelty in the proposed method, ablation experiments are conducted on the corridor dataset, exhibition hall dataset, and office dataset.  The APE results of robot localization under different ablation conditions are presented in Table \ref{tab:3_dataset}. \textbf{Ours (None)} refers to our proposed method with the adaptive knot span adjustment strategy, improved EKF module, and VA generation disabled. The uniform B-splines and conventional EKF are utilized. 
\textbf{Ours (AKSA)}, \textbf{Ours (IEKF)}, and \textbf{Ours (VA}) denote variants of Ours (None) with the adaptive knot span adjustment, improved EKF module, and VA generation enabled individually, respectively. \textbf{Ours} represents the full implementation of our proposed method. }

{The effectiveness of the adaptive knot span adjustment strategy in handling dynamic motion is highlighted in the fast and hybrid sequences. By adjusting knot spans in response to sudden motion changes, the strategy enhances the localization robustness. 
As shown in Table~\ref{tab:3_dataset}, both Ours (IEKF) and Ours (VA) consistently outperform Ours (None) across all sequences.  While IMU are prone to noise, particularly during aggressive motions,  odometry provides more reliable velocity data, enabling effective sensor fusion.  Ours (IEKF) constructs an IMU/odometer fusion model using an improved Extended Kalman Filter (EKF) with innovation-based adaptive estimation, successfully integrating IMU and odometry data. 
Ours (VA) further leverages the IMU/odometer fusion model to generate VAs, providing additional observation constraints for the UWB location system and enhancing overall system stability. When all modules are enabled,  the full framework \textbf{Ours} achieves the highest overall accuracy and robustness.}

\begin{table*}[!htb]
\centering
\caption{{Performance comparison in APE (RMSE, Meter)
under different conditions in ablation.}}
\resizebox{0.98\textwidth}{!}{
	\begin{threeparttable}
		\begin{tabular}{c c c c c c c c c c c c c c c c c c c c c c c}   
			\toprule
			\midrule
			\multirow{2}{*}{Method}  & \multicolumn{6}{c}{Corridor Dataset}  & \multicolumn{6}{c}{Exhibition Hall Dataset} & \multicolumn{6}{c}{Office Dataset}\\
			\cmidrule(r){2-7} 
			\cmidrule(r){8-13}
			\cmidrule(r){14-19}
			 & A-S1 & A-S2 & A-J1 & A-J2 & A-H1 & A-H2 & B-S1 & B-S2 & B-J1 & B-J2 & B-H1 & B-H2 & C-S1 & C-S2 & C-J1 & C-J2 & C-H1 & C-H2\\
			\midrule
			Ours (None)   & 0.270 & 0.266 & 0.482 & 0.503 & 0.512 & 0.488 & 0.052 & 0.138 & 0.344 & 0.320 & 0.405 & 0.396 & 0.381 & 0.357 & 0.341 & 0.342 & 0.267 & 0.305\\
			Ours (AKSA)  & 0.243 & 0.255 & 0.190 & 0.182 & 0.455 & 0.428 & 0.045 & 0.135 & 0.220 & 0.138 & 0.167 & 0.330 & 0.360 & 0.345 & 0.305 & 0.241& 0.207 & 0.249 \\
			Ours (IEKF) & 0.250 & 0.235 & 0.406 & 0.431 & 0.492 & 0.457 &  0.043 & 0.132 & 0.268 & 0.161 & 0.264 & 0.351 & {0.344} & {0.352} & {0.331} & 0.249& 0.243 & 0.252  \\
			Ours (VA)  & 0.262 & 0.240 & 0.451 & 0.482 & 0.456 & 0.467 & 0.046 & 0.134 & 0.306 & 0.291 & 0.347 & 0.329 & 0.362 & 0.348 & 0.306 & 0.247 & 0.230 & 0.235 \\
			\textbf{Ours}  & \textbf{0.208} & \textbf{0.242} & \textbf{0.181} & \textbf{0.153} & \textbf{0.403} & \textbf{0.416} & \textbf{0.040} & \textbf{0.122} & \textbf{0.219} & \textbf{0.098} & \textbf{0.150} & \textbf{0.311} & \textbf{0.352} & \textbf{0.320} & \textbf{0.263} & \textbf{0.202} & \textbf{0.189} & \textbf{0.208} \\

			\bottomrule
		\end{tabular}
			
	\end{threeparttable}
}
\label{tab:3_dataset}
\vspace{-0.2in}
\end{table*}

\subsection{Computational Costs}
{In this part, we investigate the time efficiency of continuous-time methods on the corridor dataset, and Table~\ref{tab:pose_hz} shows the results. 
The time consumptions for optimization of CT-UIO are $12.64\mathrm{ms}$, $26.07\mathrm{ms}$, and $21.42\mathrm{ms}$ for the slow, fast, and hybrid sequences, respectively. 
In slow sequences, the motion is typically gentle, resulting in a larger knot span and fewer control points to be optimized, thereby reducing the optimization time. In contrast,  in fast or hybrid sequences, larger variations in translational or rotational velocity require an increased number of control points to improve accuracy,  which in turn leads to higher computational costs. 
For comparison, SFUISE employs a uniform control point insertion strategy, configured with $10$ control points at $1Hz$ as suggested in~\cite{li2023continuous}, regardless of motion characteristics, leading to a more consistent but potentially less efficient time performance.}

\begin{table}[!htb]
	\centering
	\caption{The average time consumption for optimization ($ms$).}
	\resizebox{0.48\textwidth}{!}{
			\begin{tabular}{c c c c c c c}
				\toprule
				\midrule
				Sequences &  \textit{Slow sequences} & \textit{Fast sequences} & \textit{Hybrid sequences} \\
				\midrule
				SFUISE\cite{li2023continuous  } & 24.50 & 27.61 & 25.83 \\
                Ours& 12.64 & 26.07 & 21.42 \\
				\bottomrule
			\end{tabular}
		}
		\label{tab:pose_hz}
	\end{table}

\section{Conclusion}
\label{conclusion}
In this article, we present CT-UIO, a continuous-time UWB-Inertial-odometer localization system designed to achieve accurate and robust localization under few-anchor conditions. CT-UIO tightly integrates UWB ranging, IMU, and odometer data using a non-uniform B-spline trajectory representation. Specifically, the proposed system enhances state estimation accuracy in fast-moving scenarios by adaptively placing control points according to varying motion speeds. 
To ensure the UWB-based localization system is fully observable, CT-UIO combines UWB ranging with motion prior from the IMU/odometer fusion model to generate VAs. 
Additionally, CT-UIO’s backend conducts adaptive sliding window factor pose graph optimization, incorporating constraints from VA's ranging, IMU, and odometer measurements.  The proposed CT-UIO system is evaluated in several real-world few-anchor situations using self-collected datasets. 
Compared to state-of-the-art discrete-time and continuous-time estimation methods, CT-UIO consistently produces more consistent and robust trajectory estimation, particularly in fast-motion scenes.

{In this work, the anchor positions are manually pre-calibrated using laser meters, which limits the applicability of the proposed CT-UIO method in UWB networks with unknown anchor locations or scenarios that require fully automated deployment without manual intervention. 
In the future, it is worthwhile exploring the implementation of the proposed CT-UIO method with unknown anchor positions, as well as incorporating complementary sensors, such as cameras, to enhance the accuracy of short-term pose estimates.}

\bibliography{references}
\bibliographystyle{IEEEtran}

\vfill

\end{document}